%% file: main.tex
\documentclass[sigconf,screen]{acmart} 
\acmConference[ISSTA 2023]{ACM SIGSOFT International Symposium on Software Testing and Analysis}{17-21 July, 2023}{Seattle, USA}

\setcopyright{acmlicensed}
\acmPrice{15.00}
\acmDOI{10.1145/3597926.3598127}
\acmYear{2023}
\copyrightyear{2023}
\acmSubmissionID{issta23main-p346-p}
\acmISBN{979-8-4007-0221-1/23/07}
\acmConference[ISSTA '23]{Proceedings of the 32nd ACM SIGSOFT International Symposium on Software Testing and Analysis}{July 17--21, 2023}{Seattle, WA, United States}
\acmBooktitle{Proceedings of the 32nd ACM SIGSOFT International Symposium on Software Testing and Analysis (ISSTA '23), July 17--21, 2023, Seattle, WA, United States}
\received{2023-02-16}
\received[accepted]{2023-05-03}

\usepackage{amsthm}
\newtheorem{definition}{Definition}
\newtheorem{example}{Example}
\usepackage{graphicx}
\usepackage{subcaption}
\usepackage[ruled,vlined,linesnumbered]{algorithm2e}

\SetCommentSty{mycommfont}
\usepackage{tabularx}
\usepackage{hhline}
\usepackage{makecell}
\usepackage{multirow}
\usepackage{tikz}
\usepackage{enumerate}
\usepackage{microtype}

\newtoggle{conf-ver}
\settoggle{conf-ver}{false}

\begin{document}

\title[Tightening Over-Approximation for DNN Robustness Verification via Under-Approximation]{A Tale of Two Approximations: 
Tightening Over-Approximation for DNN Robustness Verification via Under-Approximation}

\author{Zhiyi Xue}
\email{51255902046@stu.ecnu.edu.cn}
\affiliation{
    \institution{East China Normal University}
    \city{Shanghai}
    \country{China}
}

\author{Si Liu}
\email{si.liu@inf.ethz.ch}
\affiliation{
	\institution{ETH Z{\"u}rich}
	\city{Z{\"u}rich}
	\country{Switzerland}
}

\author{Zhaodi Zhang}
\email{zdzhang@stu.ecnu.edu.cn}
\affiliation{
    \institution{Chengdu Education Research Institute}
    \city{Chengdu}
    \country{China}
}

\author{Yiting Wu}
\email{51205902026@stu.ecnu.edu.cn}
\affiliation{
    \institution{East China Normal University}
    \city{Shanghai}
    \country{China}
}

\author{Min Zhang}
\email{zhangmin@sei.ecnu.edu.cn}
\affiliation{
    \institution{East China Normal University}
    \city{Shanghai}
    \country{China}
}

\begin{abstract}
The robustness of deep neural networks (DNNs) is crucial to the hosting system's reliability and security. Formal verification has been demonstrated to be effective in providing  provable robustness guarantees. To improve  its scalability, over-approximating the non-linear activation functions in DNNs by linear constraints has been widely adopted, which transforms the verification problem into an efficiently solvable linear programming problem. 
Many efforts have been dedicated to defining the so-called tightest approximations to reduce overestimation imposed by over-approximation.

In this paper, we study existing approaches and identify a dominant factor in defining tight approximation, namely the \emph{approximation domain} of the activation function. 
We find out that tight approximations defined on approximation domains may not be as tight as the ones on their actual domains, yet existing approaches all rely only on  approximation domains.
Based on this observation, we propose a novel 
dual-approximation approach to tighten over-approximations,  
leveraging an activation function's underestimated domain 
to define tight approximation bounds.
We implement our approach with two complementary algorithms based respectively on Monte Carlo simulation and gradient descent into a tool called DualApp. 
We assess it on a comprehensive benchmark of DNNs with different architectures. Our experimental results show that DualApp significantly outperforms the state-of-the-art approaches with  $100\%-1000\%$ improvement on the verified robustness ratio and $10.64\%$ on average (up to $66.53\%$) on the certified lower bound. 
\end{abstract}



\maketitle

\input{1introduction.tex}

\input{2preliminaries.tex}

\input{3motivation.tex}

\input{4method.tex}

\input{5experiment.tex}

\input{6relatedwork.tex}

\input{7conclusion.tex}


\bibliographystyle{ACM-Reference-Format}
\bibliography{ref}


\iftoggle{conf-ver}{}{
\appendix
\input{8appendix.tex}

}

\end{document}

%% file: 1introduction.tex
\section{Introduction}
Deep neural networks (DNNs) are the most crucial components in AI-empowered software systems. They must be guaranteed reliable and secure when the hosting system is safety-critical. Robustness is central to their safety and reliability, ensuring that neural networks can function correctly even under environmental perturbations and adversarial attacks~\cite{szegedy2013intriguing,goodfellow2014explaining,wu2020robustness}. Studying the robustness of DNNs from both training and engineering perspectives attracts researchers from both AI and SE communities~\cite{szegedy2013intriguing,goodfellow2014explaining,ilyas2019adversarial,zhang2023boosting,LiuFY022,Pan20}. More recently, the emerging formal verification efforts on the robustness of neural networks aim at providing certifiable robustness guarantees for the neural networks~\cite{huang2020survey,liu2021algorithms,wing2021trustworthy}. Certified robustness of neural networks is necessary for guaranteeing that the hosting software system is both safe and secure. 
It is particularly crucial to those safety-critical applications such as autonomous drivings~\cite{bojarski2016end,apollo}, medical diagnoses~\cite{titano2018automated}, and face recognition ~\cite{sun2015deepid3}. 

Formally verifying the robustness of neural networks is computationally complex and expensive due to the high non-linearity and non-convexity of neural networks. The problem has been proved NP-complete even for the simplest fully-connected networks with the piece-wise linear activation function ReLU~\cite{katz2017reluplex}. It is  significantly more difficult for those networks that contain differentiable S-curve  activation functions such as Sigmoid, Tanh, and Arctan~\cite{zhang2018efficient}. To improve scalability, a practical solution is to over-approximate the nonlinear activation functions by using linear upper and lower bounds. The verification problem is then transformed into an efficiently solvable linear programming problem. The linear over-approximation is a prerequisite for other advanced verification approaches based on abstraction~\cite{singh2019abstract,pulina2010abstraction,elboher2020abstraction}, interval bound propagation (IBP)~\cite{huang2019achieving}, and convex optimization~\cite{wong2018provable,salman2019convex}. 

As over-approximations inevitably introduce overestimation, the corresponding verification approaches  sacrifice completeness and may fail to prove or disprove the robustness of a neural network~\cite{liu2021algorithms}. Consequently, we cannot conclude that a neural network is not robust when we fail to prove its robustness via over-approximation. An ideal approximation must be as tight as possible to resolve such uncertainties. Intuitively, an approximation is tighter if it introduces less overestimation to the robustness verification result.

Considerable efforts have been devoted to finding tight over-approximations for precise verification results \cite{zhang2018efficient,lee2020lipschitz,wu2021tightening,lin2019robustness,DBLP:conf/nips/TjandraatmadjaA20}. 
The definition of tightness can be classified into two categories: neuron-wise and network-wise. An approximation method based on network-wise tightness is dedicated to defining a linear approximation so that the output for each neuron in the neural network is tight. An approximation method based on neuron-wise tightness only guarantees that the approximation is tight on the current neuron, while it does not consider the tightness of networks widely. Lyu \textit{et al.} \cite{lyu2020fastened} and Zhang \textit{et al.}~\cite{DBLP:conf/kbse/ZhangWLLZ22} claim that computing the tightest approximation is essentially a network-wise non-convex optimization problem, and thus almost impractical to solve directly due to high computational complexity. Hence, approximating each individual activation function separately is still an effective and practical solution. 
Experimental results have shown that existing tightness characterizations of neuron-wise over-approximations do not always imply precise verification results \cite{DBLP:conf/kbse/ZhangWLLZ22,salman2019convex}. It is therefore desirable to explore missing factors in defining tighter neuron-wise approximations. 

In this paper we report a new, crucial factor for defining tight over-approximation, namely \emph{approximation domain} of an activation function, which is missed by all  existing approximation approaches. 
An approximation domain refers to an interval of $x$, on which an activation function $\sigma(x)$ is over-approximated by upper and lower linear bounds. Through both theoretical and experimental analyses, we identify that existing approaches  rely  only on the approximation domain of an activation function to define linear lower and upper bounds, 
yet the bound that is tight on the approximation domain may not be tight on the activation function's \emph{actual domain}. 
The actual domain of $\sigma(x)$ must be enclosed by the approximation domain to guarantee the soundness of the over-approximation. 
Unfortunately, computing the actual domain of an activation function on each neuron of a DNN is as difficult as the verification problem,  thus impractical. 

Towards estimating the actual domain,
we propose a novel dual-approximation approach which,
unlike existing approaches, 
leverages the \emph{underestimated domain}, i.e., an interval that is enclosed by the actual domain of an activation function, to define a tight linear approximation.
We first devise two under-approximation algorithms to compute the underestimated domain based on Monte Carlo simulation and gradient descent, respectively. 
In the Monte Carlo algorithm, we select a number of samples from the perturbed input region and feed them into a DNN, recording the maximum and minimum of each neuron as the underestimated domain. For the gradient-based algorithm, we feed the image into a DNN to obtain the gradient of each neuron relative to the input. Based on this, we fine-tune the input value and feed them into the DNN again to get the underestimated domain. 

We then use both underestimated and approximation domains to define tight linear bounds for the activation function. 
Specifically, we 
define a linear over-approximation bound on the underestimated domain and check if it is valid on the approximation domain. In a valid case, we approximate the activation function using the bound; otherwise, we define a bound on the original approximation domain. 
The underestimated domain is an inner approximation of the actual domain, which guarantees tightness, whereas the approximation domain guarantees soundness. 
Through an extensive analysis on a wide range of benchmarks and datasets, we demonstrate that our dual-approximation approach can produce tighter linear approximation than the state-of-the-art approaches that claim to provide the tightest approximation. In particular, our approach achieves $100\%-1000\%$ improvement on the verified robustness ratio and $10.64\%$ on average (up to 66.53\%) on the certified lower bound.

Overall, we make three main contributions:
\begin{enumerate}[(1)]

	\item We identify a crucial factor, called  \textit{approximation domain}, in defining tight over-approximation for the DNN robustness verification by a thorough study of the state-of-the-art over-approximation methods. 

	\item We propose two under-approximation algorithms for computing  underestimated domains, together with a dual-approximation approach to defining tight over-approximation for the DNN robustness verification. 
	
	\item We implement our approach into a tool called DualApp and demonstrate its outperformance over the state-of-the-art tools on a wide range of benchmarks. 
 We also experimentally explore the optimal parameter settings for computing more precise underestimated approximation domains. 

\end{enumerate}


%% file: 2preliminaries.tex
\section{Preliminaries}

\label{sec:preliminaries}


\subsection{Deep Neural Networks}

\begin{figure}[t]
	\centering
	\includegraphics[width=0.46\textwidth]{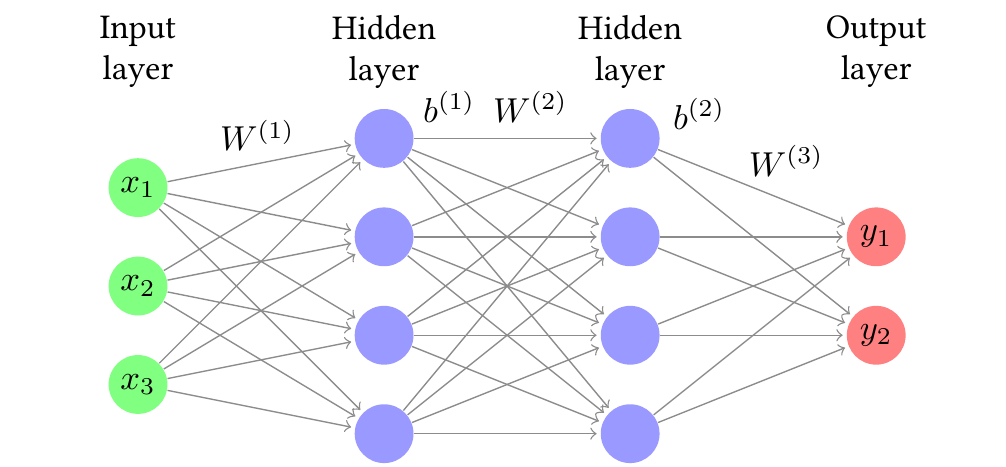}
	\caption{A 4-layer feedforward DNN with two hidden layers.}
	\label{fig:dnn}
\end{figure}

A deep neural network (DNN) is a network of nodes called neurons connected end to end as shown in Figure \ref{fig:dnn}, which implements a mathematical function $F:\mathbb{R}^n \rightarrow \mathbb{R}^{m}$, e.g., $n=3$ and $m=2$ for the 2-hidden-layer DNN in Figure \ref{fig:dnn}. Neurons except input ones are also functions $f:\mathbb{R}^k \rightarrow \mathbb{R}$ in the form of $f(x) = \sigma(Wx+b)$, where $k$ is the dimension of input vector $x$, $\sigma(\cdot)$ is called an \emph{activation function}, $W$ a matrix of weights and $b$ a bias. During calculation, a vector of $n$ numbers is fed into the neural network from the \emph{input layer} and propagated layer by layer through the internal \emph{hidden layers} after being multiplied by the weights on the edges, summed at the successor neurons with the bias and then computed by the neurons using the activation function. The neurons on the \emph{output layer} compute the output values, which are regarded as probabilities of classifying an input vector to every label. The input vector can be an image, a sentence, a voice, or a system state, depending on the application domains of the deep neural network.

Given an $l$-layer neural network, let $W^{(i)}$ be the matrix of weights between the $i$-th and $(i+1)$-th layers, and $b^{(i)}$ the biases on the corresponding neurons, where $i=1,\ldots,l-1$. The function $F:\mathbb{R}^n \rightarrow \mathbb{R}^{m}$ implemented by the neural network can be defined by:
\begin{align}
	&F(x)=W^{(l-1)}\sigma(z^{(l-1)}(x)),\tag{Network Function} \\
	\text{where} ~&z^{(i)}(x) = W^{(i)} \sigma(z^{(i-1)}(x)) + b^{(i)}\tag{Layer Function} \\
        \text{and} ~&z^{(0)}(x) = x \tag{Initialization}
\end{align}
for $i=1,\ldots,l-1$. For the sake of simplicity, we use $\hat{z}^{(i)}(x)$ to denote $\sigma(z^{(i)}(x))$ and $\Phi(x)= \arg \max_{\ell\in L} F(x)$ to denote the label $\ell$ such that the probability $F_{\ell}(x)$ of classifying $x$ to $\ell$ is larger than those to other labels, where $L$ represents the set of all labels.
The activation function $\sigma$ usually can be a Rectified Linear Unit (ReLU), $\sigma(x)=max(x,0)$), a Sigmoid function $\sigma(x)=\frac{1}{1+e^{-x}}$, a Tanh function $\sigma(x) = \frac{e^x - e^{-x}}{e^x + e^{-x}}$, or an Arctan function $\sigma(x) = tan^{-1}(x)$. As ReLU neural networks have been 
comprehensively studied \cite{liu2021algorithms}, we focus on the networks with only \emph{S-curved} activation functions, i.e., Sigmoid, Tanh, and Arctan.

Given a training dataset, the task of training a DNN is to fine-tune the weights and biases so that the trained DNN achieves desired precision on test sets. 
Although a DNN is a precise mathematical function, its correctness is very challenging to guarantee due to the lack of formal specifications and the inexplicability of itself. Unlike programmer-composed programs, the machine-trained models are almost impossible to assign semantics to the internal computations.

\subsection{Neural Network Robustness Verification}

Despite the challenge in verifying the correctness of DNNs, formal verification is still useful to verify their safety-critical properties. One of the most important properties is \emph{robustness}, stating that the prediction of a neural network is still unchanged even if the input is manipulated under a reasonable range:

\begin{definition}[Neural Network Robustness]
    \label{def:robustness}
    A neural network $F:\mathbb{R}^n \rightarrow \mathbb{R}^{m}$ is called \textit{robust} with respect to an input $x_0$ and an input region $\Omega$ around $x_0$ if $\forall x \in \Omega, \Phi(x) = \Phi(x_0)$ holds.
\end{definition}

Usually, input region $\Omega$ around input $x_0$ is defined by a $\ell_p$-norm ball around $x_0$ with a radius of $\epsilon$, i.e. $\mathbb{B}_p (x_0, \epsilon)=\{x | \| x-x_0\|_p \le \epsilon\}$. In this paper, we focus on the \emph{infinity norm} and verify the robustness of the neural network in $\mathbb{B}_\infty (x_0, \epsilon)$ on image classification tasks. A corresponding robust verification problem is to compute the largest $\epsilon_0$ s.t. neural network $F$ is robust in $\mathbb{B}_\infty (x_0, \epsilon_0)$. The largest $\epsilon$ is called a \textit{certified lower bound}, which is a metric for measuring both the robustness of neural networks and the precision of robustness verification approaches. Another problem is to compute the ratio of pictures that can be classified correctly when given a fixed $\epsilon$, and that is called a verified robustness ratio.

Assuming that the output label of $x_0$ is $c$, i.e. $\Phi(x_0) = c$, proving $F$'s robustness in Definition \ref{def:robustness} is equivalent to showing $\forall x \in \Omega, \forall \ell\in L/\{c\}, F_c(x) - F_\ell(x) > 0$ holds. Thus, the verification problem is equivalent to solving the following optimization problem:
\begin{equation}
    \label{eq:verification_problem_1}
    \min_{x \in \Omega} (F_c(x) - \max_{\ell \in L/\{c\}} (F_{\ell}(x)))
\end{equation}
We can conclude that $F$ is robust in $\Omega$ if the result is positive. Otherwise, there exists some input $x'$ in $\Omega$ and $\ell'$ in $L/\{c\}$ such that $F_{\ell'}(x')\geq F_c(x')$. Namely, the probability of classifying $x'$ by $F$ to $\ell'$ is greater than or equal to the one to $c$, and consequently, $x'$ may be classified as $\ell'$, meaning that $F$ is not robust in $\Omega$.

\subsection{Verification via Linear Over-Approximation}
\label{sec:linear_approximation}

The optimization problem in Equation \ref{eq:verification_problem_1} is computationally expensive, and it is typically impractical to compute the precise solution. The root reason for the high computational complexity of the problem is the non-linearity of the activation function $\sigma$. Even when $\sigma$ is piece-wise linear, e.g., the commonly used ReLU ($\sigma(x)=\max(x,0)$), the problem is  NP-complete~\cite{katz2017reluplex}. A pragmatic solution to simplify the verification problem is to over-approximate $\sigma(x)$ by linear constraints and symbolic propagation and transform it into an efficiently-solvable linear programming problem  \cite{wang2018formal,wang2022interval}.

\begin{definition}[Linear Over-Approximation]
    \label{def:linear_over_approximation}
    Let $\sigma(x)$ be a non-linear function on $[l,u]$ and $h_L(x)=\alpha_L x + \beta_L, h_U(x)=\alpha_U x + \beta_U$ be two linear functions for some $\alpha_L, \alpha_U, \beta_L, \beta_U \in \mathbb{R}$. 
	$h_L(x)$ and $h_U(x)$ are called the lower and upper linear bounds of $\sigma(x)$ on $[l,u]$ respectively if $h_L(x) \le \sigma(x) \le h_U(x)$ holds for all $x$ in $[l,u]$.
\end{definition}

By Definition \ref{def:linear_over_approximation}, we can simplify Equation \ref{eq:verification_problem_1} to the following efficiently solvable linear optimization problem. Note here that $z$ is  an interval, instead of  a number:
\begin{equation}
    \label{eq:verification_problem_2}
	\begin{split}
		& \min ( \min(z_c^{(m)}(x)) - \max(z_{\ell}^{(m)}(x))) \\
		s.t. \quad & z^{(i)}(x) = W^{(i)} \hat{z}^{(i-1)}(x) + b^{(i)}, i \in {1,...,m}\\
		& h_L^{(i)}(x) \le \hat{z}^{(i)}(x) \le h_U^{(i)}(x), i \in {1,...,m-1}\\
		& x \in \Omega, \ell \in L/c, \hat{z}^{(0)}(x) = [x, x]
	\end{split}
\end{equation}

\begin{figure}[t]
	\begin{center}
		\includegraphics[width=0.45\textwidth]{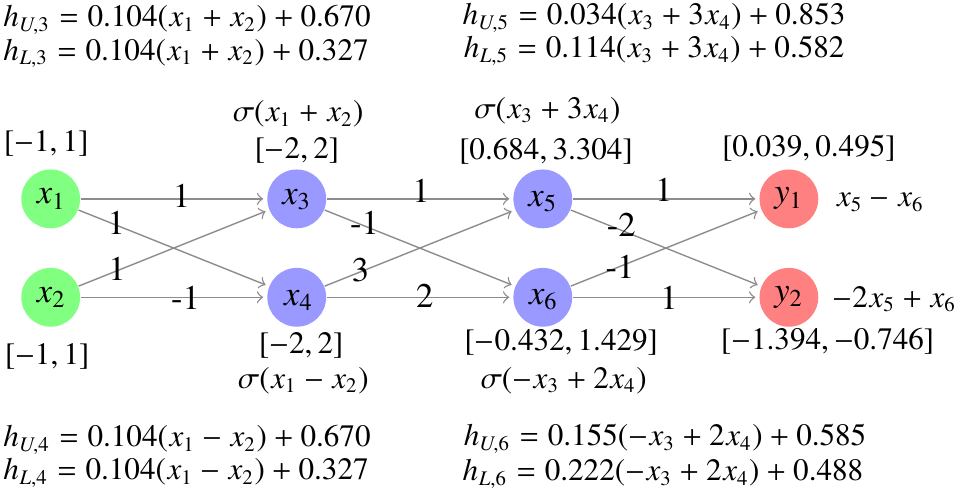}
		\caption{Verifying the robustness of a  4-layer Sigmoid network by the linear over-approximation.}
		\label{fig:example1}
	\end{center}
\end{figure}

\begin{example}
    \label{example1}
    Let us consider the example in Figure \ref{fig:example1}, which shows the verification process of a simple neural network based on linear approximation. It is a fully-connected neural network with two hidden layers, $x_1, x_2 $ are input neurons, and $y_1, y_2$ are output neurons. The intervals represent the range of neurons before the application of the activation function. We conduct linear bounds for each neuron with an activation function using the information of intervals. $h_{U,i}$ and $h_{L,i}$ are the upper and lower linear bounds of $\mathit{\sigma}(x_i)$ respectively. From the computed intervals of output neuron, we have $\min(y_1) - \max(y_2) > 0$ for all the possible $(x_1,x_2)$ in the input domain $[-1,1]\times [-1,1]$. Consequently, we can conclude that the network is robust in the input domain with respect to the class corresponding to $y_1$.
\end{example}

\begin{definition}[Approximation Soundness]
    Given a neural network $F$ and its input region $\Omega$, a linear over-approximation is called \textit{sound} if, for all $x$ in $\Omega$, we have $F_L(x) \leq F(x) \leq F_U(x)$, where $F_L$ and $F_U$ are the approximated lower and upper bounds of $F$. 
    \label{def:soundness}
\end{definition}
The essence of the soundness lies in the actual output range of a network on each output neuron being enclosed by the one after over-approximation. The guarantee of no input in $\Omega$ misclassified by an over-approximated network implies that there must be no input in $\Omega$ misclassified by the original network. 

The approximation inevitably introduces the overestimation of output ranges. 
In Example \ref{example1}, the real output range of $y_1$ is $[0.127, 0.392]$, which is computed by solving the optimization problems of minimizing and maximizing $y_1$, respectively. The one computed by over-approximation is $[0.039,0.495]$.
We use the increase rate of an output range with and without the over-approximation to measure the overestimation. Specifically, 
let $s_{over}$ and $s_{act}$ be the lengths of the overestimated interval and the actual one, respectively. The overestimation ratio $r$ is $\frac{s_{over} - s_{act}}{s_{act}}$, which can be up to 72.08\% for $y_1$, even for such a simple neural network in Example \ref{example1}.

The overestimation introduced by over-approximation usually 
renders verifying 
 an actual robust neural network infeasible (also known as \emph{incompleteness}).
For instance, when we have $\min(y_1) - \max(y_2) < 0$ by solving Problem \ref{eq:verification_problem_2}, there may be two  reasons. One is that there  exists some input such that the output on $y_1$ is indeed less than the one on $y_2$; the other  reason is that the overestimation of $y_1$ and $y_2$ causes inequality. The network is robust in the latter case while not in the former; however, the algorithms  simply report \emph{unknown} as they cannot distinguish the underlying causes.

\subsection{\hspace{-2.5mm}Variant Approximation Tightness Definitions}

Reducing the overestimation of approximation is the key to reducing failure cases. The precision of approximation is characterized by the notion of \emph{tightness} \cite{DBLP:conf/kbse/ZhangWLLZ22}. Many efforts have been made to define the tightest possible approximations. The tightness definitions can be classified into \textit{neuron-wise} and \textit{network-wise} categories.

\emph{1) Neuron-wise Tightness. }	
The tightness of activation functions' approximations can be measured independently. Given two upper bounds $h_{U}(x)$ and $h'_{U}(x)$ of activation function $\sigma(x)$ on the interval $[l,u]$, $h_{U}(x)$ is apparently tighter than $h'_{U}(x)$ if $h_{U}(x)<h'_{U}(x)$ for any $x$ in $[l,u]$ \cite{lyu2020fastened}. 
However, when $h_{U}(x)$ and $h'_{U}(x)$ intersect between $[l,u]$, their tightness becomes non-comparable. 
Another neuron-wise tightness metric is the area size of the gap between the bound and the activation function, i.e., $\int_{l}^{u}(h_{U}(x)-\sigma(x))dx$. 
A smaller area implies a tighter approximation \cite{HenriksenL20,muller2022prima}. 
Apparently, an over-approximation that is tighter than another by the definition of \cite{lyu2020fastened} is also tighter by the definition of \cite{HenriksenL20}, but not vice versa.  
What's more, another metric is the output range of the linear bounds. 
An approximation is considered to be \textit{the tightest} if it preserves the same output range as the activation function \cite{DBLP:conf/kbse/ZhangWLLZ22}. 

\emph{2) Network-wise Tightness. }
Recent studies have shown that neuron-wise tightness does not always guarantee that the compound of all the approximations of the activation functions in a network is tight too \cite{DBLP:conf/kbse/ZhangWLLZ22}. 
This finding explains why the so-called tightest approaches based on their neuron-wise tightness metrics achieve the best verification results only for certain networks. It inspires new approximation approaches that consider multiple and even all the activation functions in a network to approximate simultaneously. The compound of all the activation functions' approximations is called the network-wise tightest with respect to an output neuron if the neuron's output range is the precisest.

Unfortunately, finding the  network-wise tightest approximation has been proved a non-convex optimization problem, and thus computationally expensive \cite{lyu2020fastened,DBLP:conf/kbse/ZhangWLLZ22}. From a pragmatic viewpoint, a neuron-wise tight approximation is useful if all the neurons' composition is also network-wise tight.  The work \cite{DBLP:conf/kbse/ZhangWLLZ22} shows that there exists such a neuron-wise tight approach under certain constraints when the networks are monotonic. However, their approach does not guarantee to be optimal when the neural networks contain both positive and negative weights. 

Note that an activation function can be approximated more tightly using two more pieces of linear bounds. In this paper, we focus on one-piece linear approximation defined in Definition \ref{def:linear_over_approximation}. This method is the most efficient since the reduced problem is a linear programming problem that can be efficiently solved in polynomial time. For piece-wise approximations, it is challenging because as the number of linear bounds drastically blows up, and the corresponding reduced problem is proved to be  NP-complete~\cite{singh2019abstract}.

%% file: 3motivation.tex
\vspace{-1mm}
\section{Motivation}
\label{sec:motivation}

In this section we show that the over-approximation inevitably introduces overestimation. We observe that the overestimation of intervals is accumulatively  propagated as the intervals are propagated on a layer basis. We then identify an important factor called \emph{approximation domain} for explaining why existing, so-called tightest over-approximation approaches still introduce large overestimation. 
Finally, we illustrate that the precise approximation domain and the tight over-approximation are interdependent. 

\vspace{-1mm}
\subsection{Interval Overestimation Propagation}
\label{sec:problem}
Neural network verification methods based on over-approximation will inevitably introduce overestimation more or less, as shown in Section \ref{sec:linear_approximation}. For the approximation of an activation function, if the maximum value of its upper bound is larger than the maximum value of the actual value, or the minimum value of its lower bound is smaller than the minimum value of the actual value, the approximation is imprecise and introduces too much overestimation.

\begin{figure}
	\centering
	\begin{subfigure}{0.49\linewidth}
		\centering
		\includegraphics[width=\linewidth]{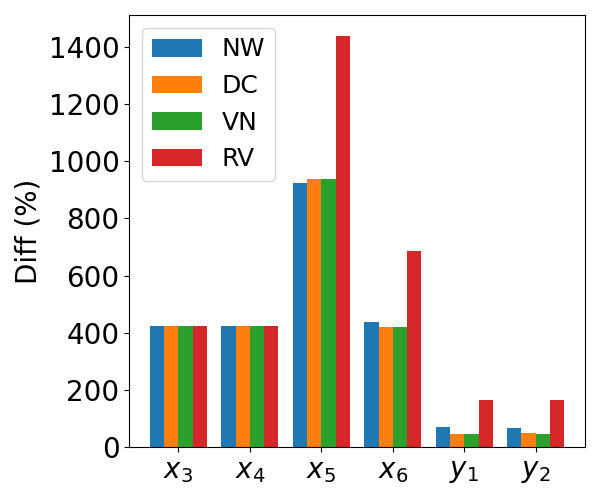}
		\caption{Overestimation on a network. }
		\label{fig:diff_case}
	\end{subfigure}
	\begin{subfigure}{0.49\linewidth}
		\centering
		\includegraphics[width=\linewidth]{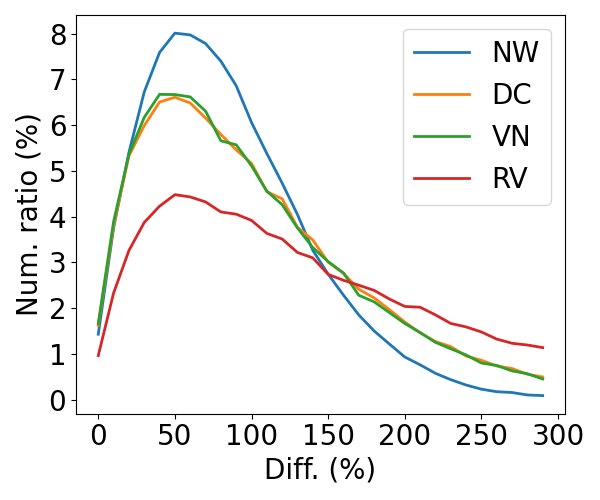}
		\caption{Overestimation distribution.}
		\label{fig:percentage}
	\end{subfigure}
	\vspace{-3mm}
	\caption{The overestimation of each neuron in the network in Figure \ref{fig:example1} (a) and the overestimation distribution on 50,000 variant networks with the same network architecture (b). }
	\vspace{-4mm}
	%
\label{fig:overestimate}
\end{figure}

In our experiment, we found that the overestimation can be accumulated and propagated layer by layer. We evaluate four state-of-the-art linear over-approximation approaches, including NeWise (NW)~\cite{DBLP:conf/kbse/ZhangWLLZ22}, DeepCert (DC)~\cite{wu2021tightening}, VeriNet (VN)~\cite{HenriksenL20}, and RobustVerifier (RV)~\cite{lin2019robustness}, to verify the neural network model defined in Figure \ref{fig:example1} with $50,000$ different weights and input intervals, and record the size of real intervals and overestimated intervals during the process. Figure \ref{fig:overestimate} shows the overestimation of each neuron with the network in Figure \ref{fig:example1}, together with the overestimation distribution of the $50,000$ cases for $y_1$ and $y_2$. In Figure \ref{fig:diff_case}, the overestimation is over $400\%$ for $x_3$, $x_4$, $x_5$, $x_6$, and around $50\%$ for $y_1$, $y_2$. Note that the overestimation of $y_1$ and $y_2$ is smaller than the ones in the hidden layers. This is due to the non-monotonicity of neural networks and the normalization of activation functions. We can find in Figure \ref{fig:percentage} that most of the overestimations are distributed between $50\%$ and $100\%$, while it is over $300\%$ in more than $10\%$ cases. 

There are two main reasons for the accumulative propagation. One apparent reason is the over-approximations of activation functions, which is inevitable but can be reduced by defining tight ones. The other reason is that over-approximations must be defined on overestimated domains of the activation function to guarantee the soundness of it. This further introduces overestimation to approximations as the domains'  overestimation increases. Due to the layer dependency in neural networks, such dual overestimation is accumulated and propagated to the output layer. 

\vspace{-1mm}
\subsection{Approximation Domain}

\begin{figure*}
    \centering
    \begin{subfigure}{0.24\linewidth}
        \centering
        \includegraphics[width=0.95\linewidth]{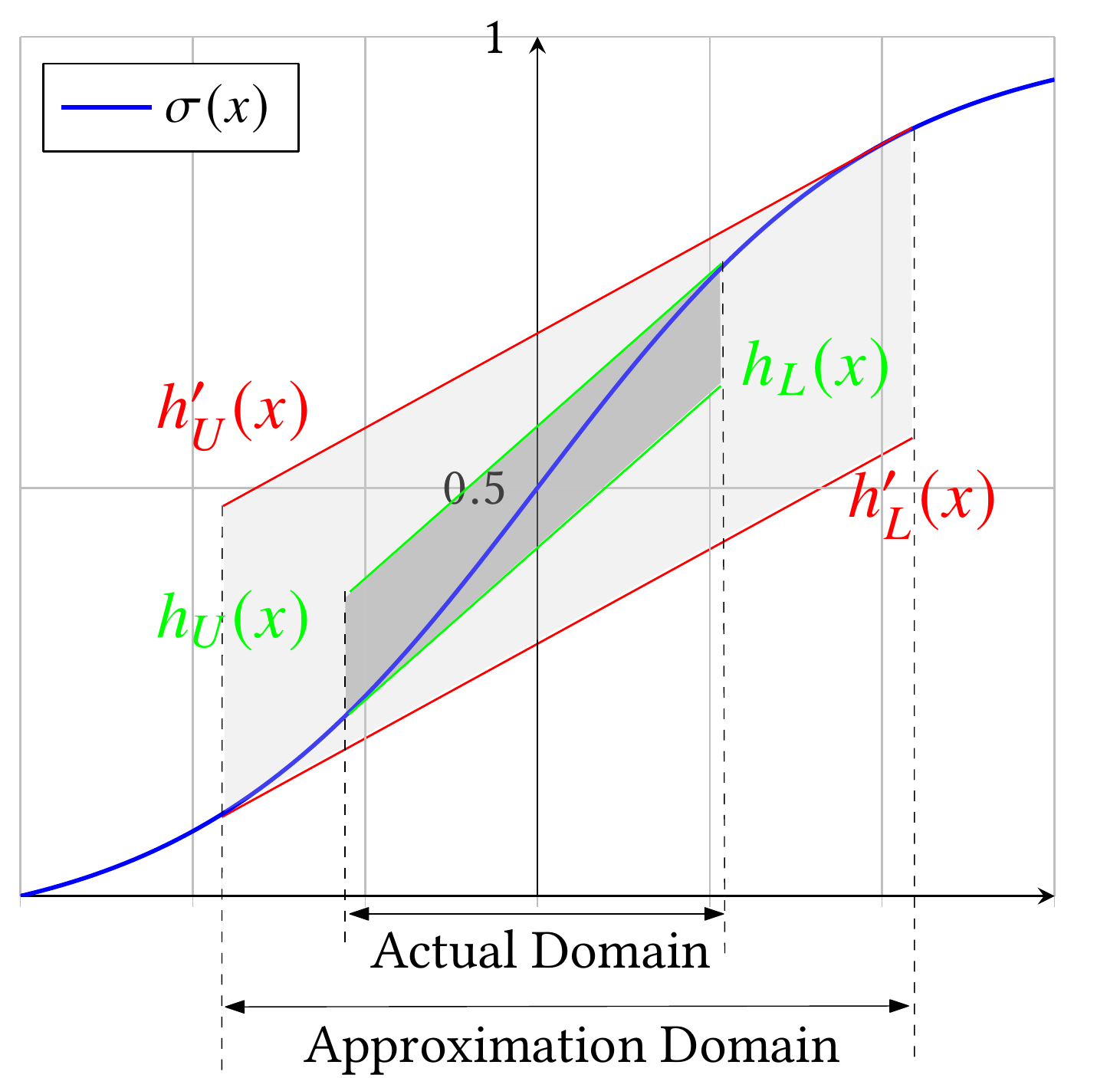}
        \caption{Tangent line at end points~\cite{DBLP:conf/kbse/ZhangWLLZ22}.}
    \end{subfigure}
    \begin{subfigure}{0.24\linewidth}
        \centering
        \includegraphics[width=0.95\linewidth]{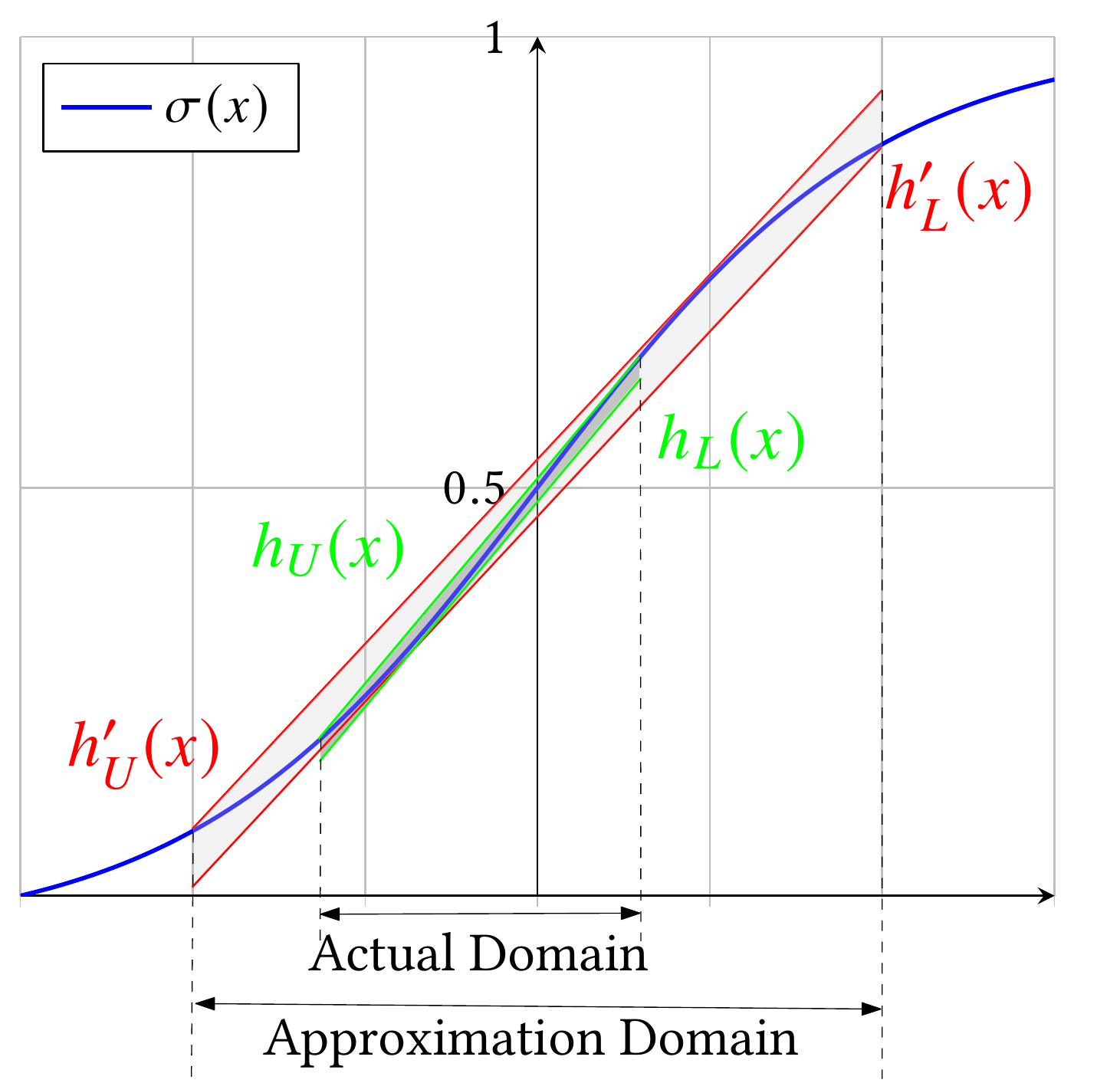}
        \caption{Minimal area~\cite{HenriksenL20}.}
    \end{subfigure}
    \begin{subfigure}{0.24\linewidth}
        \centering
        \includegraphics[width=0.95\linewidth]{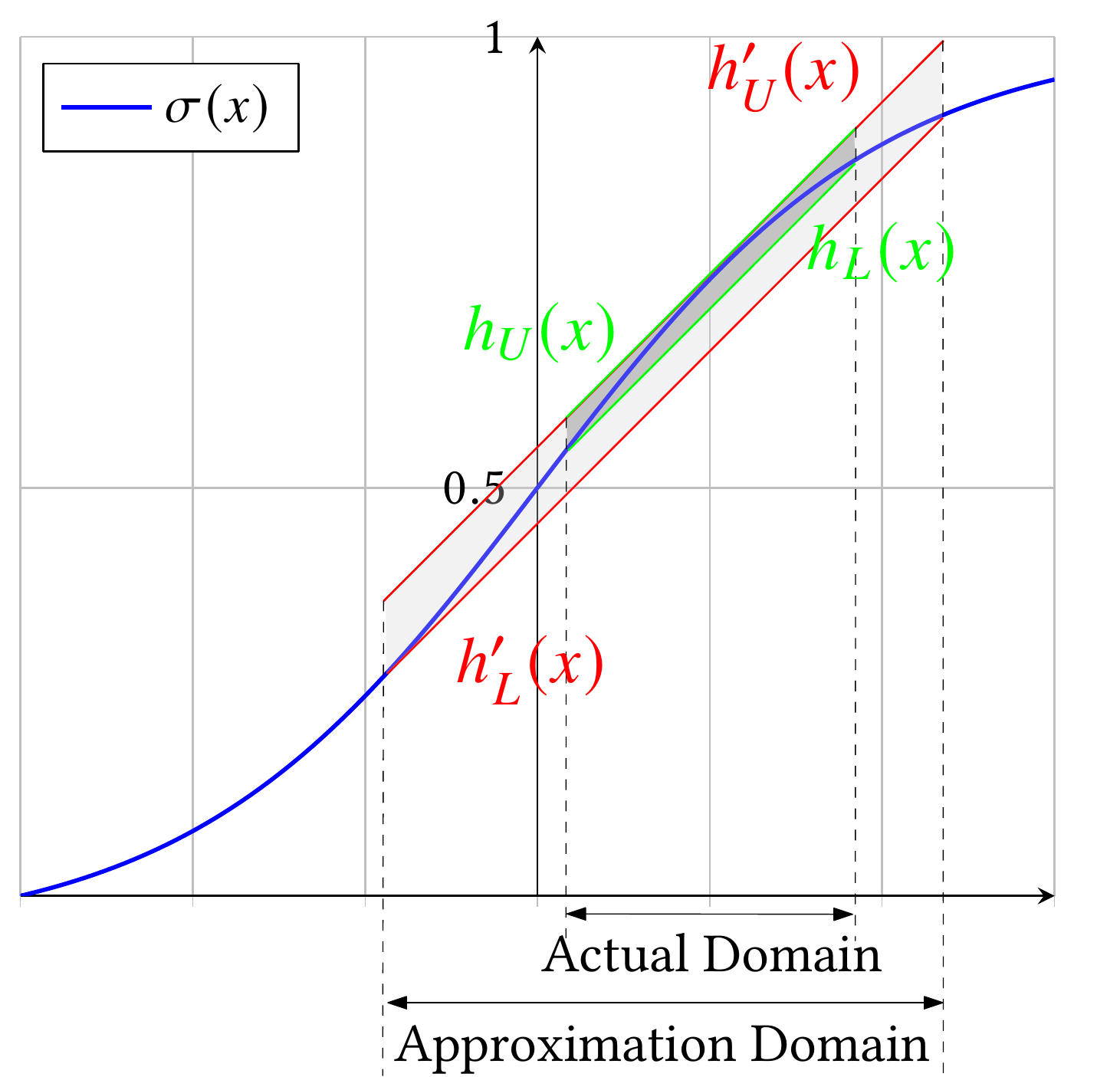}
        \caption{Parallel line~\cite{wu2021tightening,zhang2018efficient}.}
    \end{subfigure}
    \begin{subfigure}{0.26\linewidth}
        \centering
        \includegraphics[width=0.87\linewidth]{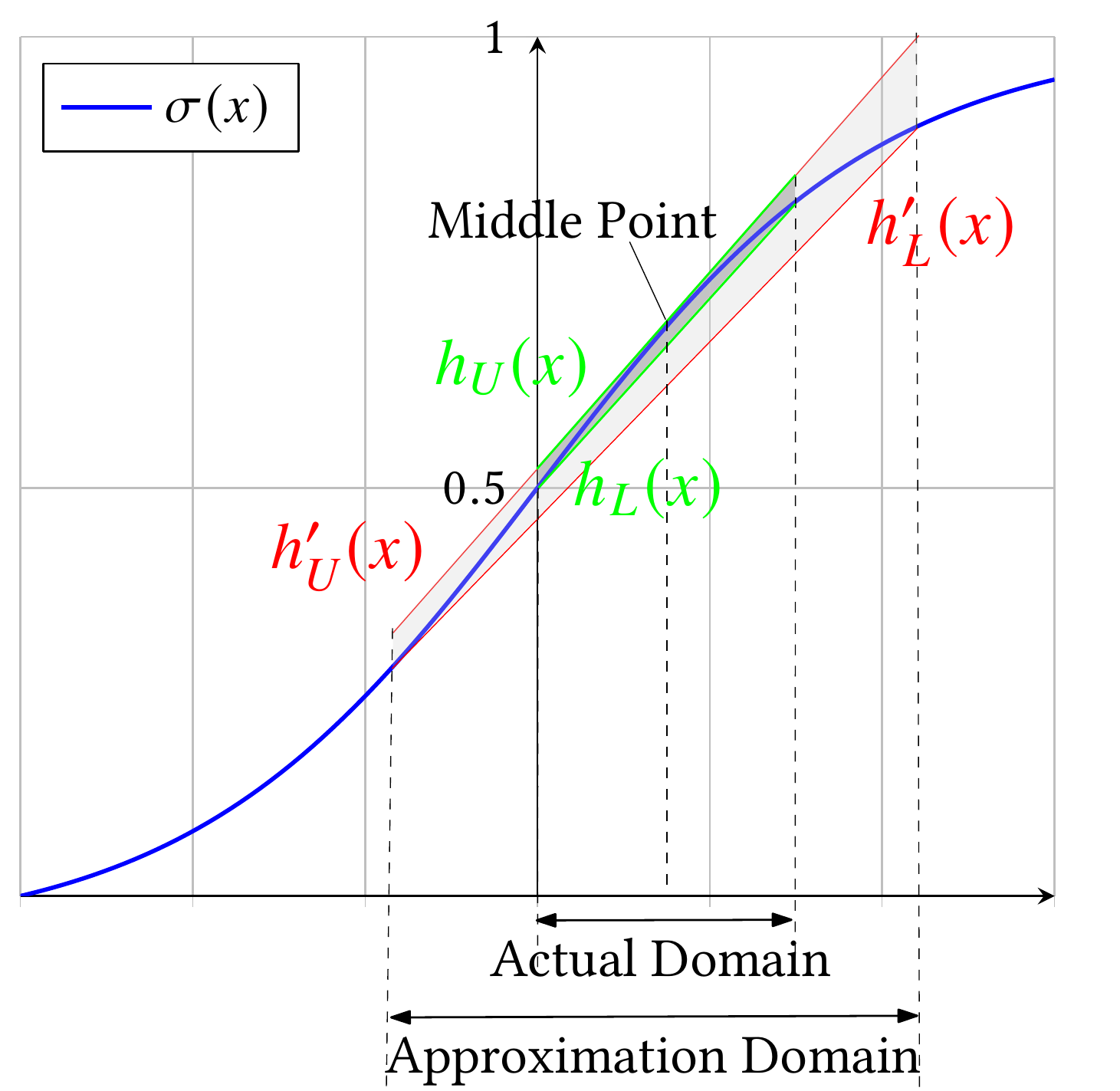}
        \caption{Tangent line at middle point \cite{HenriksenL20}.}
    \end{subfigure}
\vspace{-2mm}
    \caption{The differences between the linear over-approximations that are defined on the estimated approximation and the actual one respectively according to the four state-of-the-art approaches. 
    	The red lines refer to the upper and lower bounds defined on approximation domains, and the green lines refer to those defined on actual domains. The light and dark gray areas represent the corresponding overestimation introduced by the over-approximations, respectively.}
\vspace{-2mm}
    \label{fig:cmp_diff_method}
\end{figure*}



To justify the second reason for the accumulative propagation in Section \ref{sec:problem}, we introduce the notion of \emph{approximation domain}, to represent the domain of activation functions, on which over-approximations are defined. 

\begin{definition}[Approximation Domain]
	\label{def:approximation_domain}
	Given a neural network $F$ and an input region 
	$\mathbb{B}_\infty(x_0, \epsilon)$, the approximation domain of the $r$-th hidden neuron in the $i$-th layer is $[l_r^{(i)}, u_r^{(i)}]$, where, 
	\begin{equation*}
		\begin{split}
			l_r^{(i)} =&  \min z_r^{(i)}(x), u_r^{(i)} = \max  z_r^{(i)}(x)\\
			s.t. \quad & z^{(j)}(x) = W^{(j)}\hat{z}^{j-1}(x) + b^{(j)}, j \in {1,...,i}\\
			& h_L(z^{(j)}(x)) \leq \hat{z}^{(j)}(x) \leq h_U(z^{(j)}(x)), j \in {1,...,i-1}\\
			& x \in \mathbb{B}_\infty(x_0, \epsilon), \hat{z}^{(0)}(x)=x
		\end{split}
	\end{equation*} 
\end{definition}
 Definition \ref{def:approximation_domain} formulates the way of the existing over-approximation approaches \cite{DBLP:conf/kbse/ZhangWLLZ22,HenriksenL20,zhang2018efficient,wu2021tightening} to compute overestimated domains of activation functions for defining their over-approximations. Given two different approximation domains $[l_r,u_r]$ and $[l'_r,u'_r]$, we say $[l_r,u_r]$ is more precise than $[l'_r,u'_r]$ if $l_r\ge l'_r$ and $u_r\le u'_r$. 
Let us consider the activation functions on neurons $x_5$ and $x_6$ in Figure \ref{fig:example1} as an example. Their domains are estimated based on the approximations of $x_3$ and $x_4$ by solving the corresponding linear programming problems in Definition \ref{def:approximation_domain}. The approximation domains are  $[0.684,3.304]$ and $[-0.432,1.429]$, respectively. As shown in Figure \ref{fig:diff_case}, they are much overestimated compared with the actual ones.

\vspace{-1mm}
\subsection{The Overestimation Interdependency}
\label{two_theorem}
We show the interdependency between the two problems of defining tight over-approximations for activation functions and computing the precise approximation domains. 
The interdependency means that tighter over-approximations of activation functions result in more precise approximation domains and vice versa. Here we follow the approximation tightness definition in \cite{lyu2020fastened}, by which a lower bound $h_L(x)$ is called tighter than another $h'_L(x)$ if $h_L(x)\ge h'_L(x)$ holds for all $x$ in the approximation domain of $\sigma$. Likewise, an upper bound $h_U(x)$ is tighter than another $h'_U(x)$ if $h_U(x)\le h'_U(x)$. Apparently, a tighter approximation by definition \cite{lyu2020fastened} is still tighter by the minimal-area-based definition \cite{HenriksenL20}. 

%

\begin{theorem}
\label{thm:tight_approx}
Suppose that there are two over-approximations $h_L(z^{(j)}(x)),h_U(z^{(j)}(x))$ and ${h'}_L(z^{(j)}(x)),{h'}_U(z^{(j)}(x))$ for each $z^{(j)}(x)$ in Definition \ref{def:approximation_domain} and $h_L(z^{(j)}(x)),h_U(z^{(j)}(x))$ are tighter than ${h'}_L(z^{(j)}(x)),{h'}_U(z^{(j)}(x))$, respectively. 
The approximation domain 
$[l_r^{(i)}, u_r^{(i)}]$ computed by 
$h_L(z^{(j)}(x)),h_U(z^{(j)}(x))$ must be more precise than the one 
$[{l'}_r^{(i)}, {u'}_r^{(i)}]$  by  ${h'}_L(z^{(j)}(x)),{h'}_U(z^{(j)}(x))$. 
\end{theorem}

Intuitively, Theorem \ref{thm:tight_approx} claims that tighter approximations lead to more precise approximation domains for the activation functions of the neurons in subsequent layers of a DNN. 


\begin{theorem}
    \label{thm:precise_domain}
	Given two approximation domains $[l_r^{(i)},u_r^{(i)}]$ and $[{l'}_r^{(i)},{u'}_r^{(i)}]$ such that  ${l'}_r^{(i)} < l_r^{(i)}$ and $u_r^{(i)} < {u'}_r^{(i)}]$, for any over-approximation $({h'}_L(z^{(j)}(x)), {h'}_U(z^{(j)}(x)))$ of continuous function $\sigma(x)$ on $[{l'}_r^{(i)},{u'}_r^{(i)}]$, there exists an over-approximation $(h_L(z^{(j)}(x))$, $h_U(z^{(j)}(x)))$ on $[l_r^{(i)},u_r^{(i)}]$ such that $\forall z^{(j)}(x) \in [l_r^{(i)},u_r^{(i)}], h'_L(x) \le h_L(z^{(j)}(x)), h'_U(x) \ge h_U(z^{(j)}(x))$.
\end{theorem}

Theorem \ref{thm:precise_domain} claims that more precise approximation domains lead to tighter over-approximations of the activation function. Altogether, the two theorems preserve the tightness of an approximation through propagation in a neural network. 
\iftoggle{conf-ver}{The proofs of Theorems \ref{thm:tight_approx} and \ref{thm:precise_domain} and  are given in Appendix A of our technical paper \cite{}.}{The proofs of Theorems \ref{thm:tight_approx} and \ref{thm:precise_domain} are given in Appendix \ref{sec:add_proof}.}



The above two theorems show the overestimation interdependency of the two problems. As examples, Figure \ref{fig:cmp_diff_method} depicts the differences between the over-approximations that are defined on overestimated approximation domains and the actual domains based on the corresponding approximation approaches \cite{DBLP:conf/kbse/ZhangWLLZ22,HenriksenL20,wu2021tightening,zhang2018efficient}. Apparently, there exist much tighter over-approximations if we can reduce the overestimation of approximation domains.  These examples show the importance of more precise approximation domains  for activation functions to  define tighter over-approximations.

It is worth mentioning that it is almost impractical to define over-approximations directly on the actual domain of activation functions for non-trivial neural networks, e.g., those which have two or more hidden layers. The reason is that computing the actual domains is at least as computationally expensive as the neural network verification problem per se. If we could compute the domains for all the activation functions on hidden neurons, the robustness verification problem would then be efficiently solvable by solving the linear constraints between the last hidden layer and the output layer by using linear programming. 





%% file: 4method.tex
\section{The Dual-Approximation Approach}

\label{sec:method}

In this section we present our dual-approximation approach for defining  tight over-approximation for activation functions guided by under-approximations. Specifically, we propose two algorithms to compute underestimated approximation domains for each activation function and define different over-approximation strategies according to both overestimated and  underestimated domains.


\vspace{-1mm}
\subsection{Approach Overview}

\begin{figure}[t]
	\begin{center}
		\includegraphics[width=0.28\textwidth]{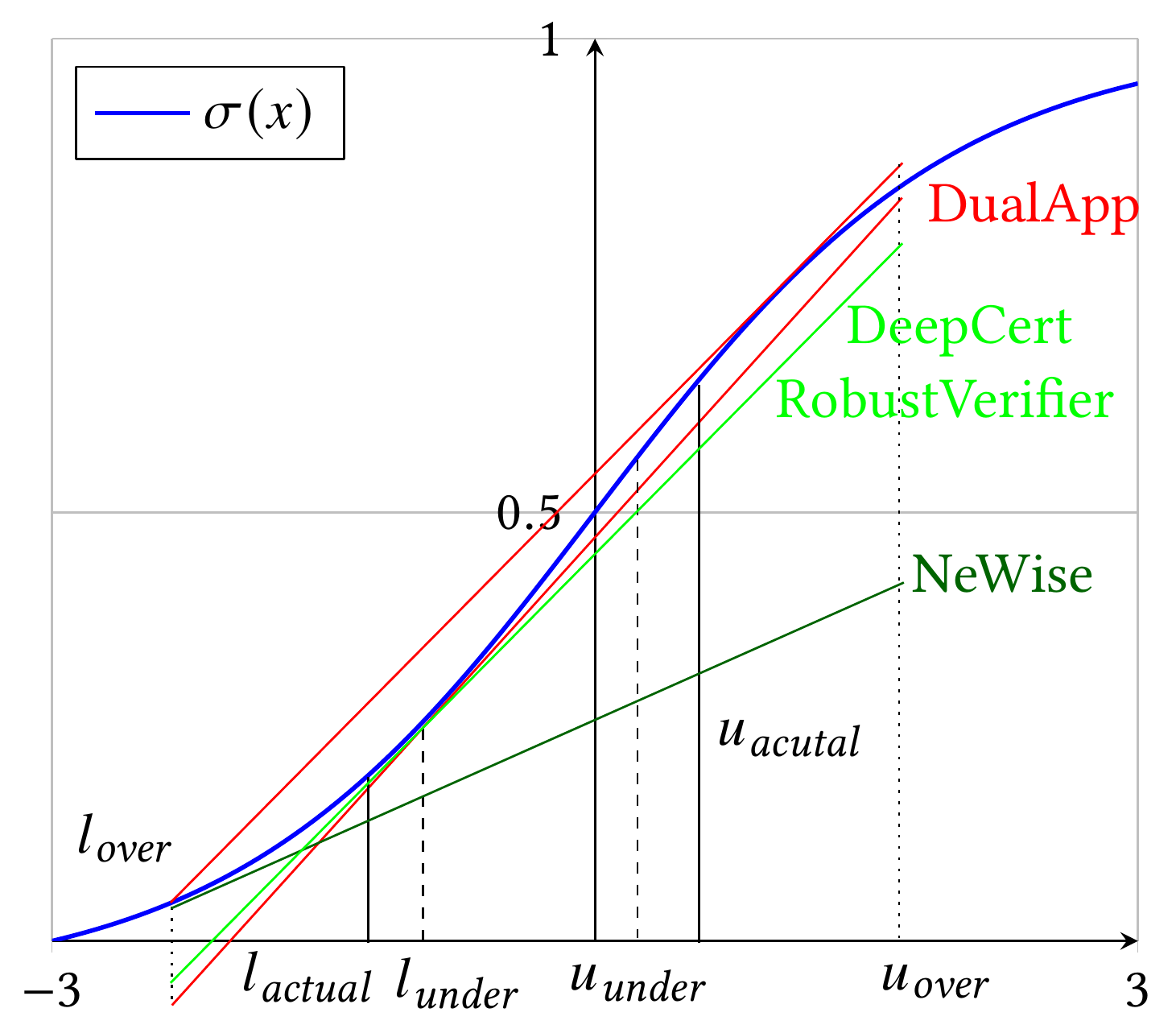}
		\caption{An illustration of our dual-approximation approach and comparison with other over-approximation approaches.}
		\label{fig:over}
		\vspace{-4mm}
	\end{center}
\end{figure}

Figure \ref{fig:over} depicts an illustration of our dual-approximation approach and the comparisons with those approaches defined only based on approximation domains. 
For each activation function, we compute an overestimated and an underestimated approximation domain, denoted by 
$[l_{\mathit{\mathit{over}}}, u_{\mathit{over}}]$ and $[l_{\mathit{under}},u_{\mathit{under}}]$, respectively. 
Underestimated domains provide useful information for defining over-approximations. Let $h(x)$ be a linear lower or upper bound of $\sigma$ on the interval $[l_{\mathit{under}},u_{\mathit{under}}]$. 
We take it as a linear over-approximation lower or upper bound for $\sigma$ on the approximation domain $[l_{\mathit{\mathit{over}}}, u_{\mathit{over}}]$ if it satisfies the condition in Definition \ref{def:linear_over_approximation}. 
According to Theorem \ref{thm:precise_domain}, we can define a tighter $h(x)$ on $[l_{\mathit{under}},u_{\mathit{under}}]$ than those defined on $[l_{\mathit{over}},u_{\mathit{over}}]$ and make sure $h(x)$ is a valid over-approximation bound on $[l_{\mathit{\mathit{over}}}, u_{\mathit{over}}]$. 
Thus, the underestimated domain is used to guarantee the over-approximation's tightness, while the approximation domain to guarantee its soundness. 
We therefore call it a \textit{dual-approximation} approach.

As shown in Figure \ref{fig:over}, we take the tangent line at $(l_\mathit{under},\sigma(l_\mathit{under}))$ as the lower bound of $\sigma$. 
Apparently, this lower bound is much tighter than the one defined by the tangent line at $(l_\mathit{over},\sigma(l_\mathit{over}))$ (the dark green line according to the approach by NeWise \cite{DBLP:conf/kbse/ZhangWLLZ22}) on the actual domain $[l_\mathit{actual},u_\mathit{actual}]$ of $\sigma$. 
It is also tighter than the green tangent line  parallel to the upper bound, according to the approaches in DeepCert \cite{wu2021tightening} and RobustVerifier \cite{lin2019robustness}.

\subsection{Under-Approximation Algorithms}

We introduce two approaches, i.e., \emph{Monte Carlo} and \emph{gradient-based}, for underestimating the actual domain of the activation functions. In other word, we propose two strategies to compute $[l_{\mathit{under}}, u_{\mathit{under}}]$. The two strategies are complementary in that the former is more efficient but computes less precise underestimated input domain, while the latter performs in the opposite way.

\setlength{\textfloatsep}{10pt}	
\begin{algorithm}[t]
	\SetKwData{Left}{left}\SetKwData{This}{this}\SetKwData{Up}{up}
	\SetKwFunction{Union}{Union}\SetKwFunction{FindCompress}{FindCompress}
	\SetKwInOut{Input}{Input}\SetKwInOut{Output}{Output}
	\caption{The Monte Carlo Approach.}
	\label{sample_algorithm}
	\Input{$F$: a network; $x_0$: an input to $F$; $\epsilon$: a $\ell_\infty$-norm radius; $n$: number of samples}
	\Output{$l_{L,r}^{(i)}, u_{L,r}^{(i)}$ for each hidden neuron $r$ on layer $i$}
	Randomly generate $n$ samples $S_n$ from $\mathbb{B}_\infty(x_0, \epsilon)$;\\
	$l_{L}\leftarrow \infty, u_{L}\leftarrow -\infty$\tcp*{Initialize all  upper and lower bounds}
	\For{each sample $x_p$ in $S_n$}{
		\For{each hidden layer $i$}{
			\For{each neuron $r$ on layer $i$}{
				$v_{p,r}^{(i)} := F_r^{(i)}(x_p)$\tcp*{Compute the output of  neuron $r$}
				$l_{L,r}^{(i)}\leftarrow \min (l_{L,r}^{(i)}, v_{p,r}^{(i)})$ \tcp*{Update $r$'s lower bound}
				$u_{L,r}^{(i)}\leftarrow \max (u_{L,r}^{(i)}, v_{p,r}^{(i)})$\tcp*{Update $r$'s upper  bound}
			}
		}
	}	
\end{algorithm}

\vspace{1mm}
\noindent 
\textbf{The Monte Carlo Approach.} 
A simple yet efficient approach for under-approximation is to randomly generate a number of valid samples and feed them into the network to track the reachable bounds of each hidden neuron's input. A sample is valid if the distance between it and the original input is less than a preset perturbation distance $\epsilon$. 

Algorithm \ref{sample_algorithm} shows the pseudo-code of the Monte Carlo approach. First, we randomly generate 
$n$ valid samples from $\mathbb{B}_\infty(x_0, \epsilon)$ (Line 1) and initialize the lower and upper bounds $l_{L,r}^{(i)}$ and $u_{L,r}^{(i)}$ of each hidden neuron (Line 2). 
Then we feed each sample into the network (Line 3), record the input value  $v^{(i)}_{p,r}$ of each activation function (Line 6), and update the corresponding lower or upper bound by $v^{(i)}_{L,r}$ (Lines 7-8). 
The time complexity of Algorithm \ref{sample_algorithm} is $O(n \sum_{i=1}^{m}k_i k_{i-1})$, where $m$ refers to the layer of the neural network, and $k_i$ refers to the number of neurons of layer $i$. 

\vspace{1mm}
\noindent
\textbf{The Gradient-Based Approach.} 
The conductivity of neural networks allows us to approximate the actual domain of each hidden neuron by gradient descent \cite{ruder2016overview}. The basic idea of gradient descent is to compute two valid samples according to the gradient of an  objective function to minimize and maximize the output value of the function, respectively. 
Using gradient descent, we can compute locally optimal lower and upper bounds as the underestimated input domains of activation functions.

Algorithm \ref{GD_algorithm} shows the pseudo-code of the gradient-based approach. 
Its inputs include a neural network $F$, an input $x_0$ of $F$, an  $\ell_\infty$-norm radius $\epsilon$, and a step length $a$ of gradient descent. It returns an underestimated input domain for each neuron on the hidden layers. 
It gets function $F^{(i)}_r$ of the neural network on neuron $r$ (Line 3), computes its gradient, and records its sign $\eta^{i}_r$ as the direction to update $x_0$ (Line 4). Then, 
the gradient descent is conducted one step forward to generate a new input $x_{\mathit{lower}}$ (Line 5).  $x_{\mathit{lower}}$ is then modified to make sure it is in the normal ball. By feeding $x_{\mathit{lower}}$ to $F^{(i)}_r$, we obtain an under-approximated lower bound $l_{L,r}^{(i)}$ (Line 7). The upper bound can be computed likewise (Lines 8-10). 

Considering the time complexity of Algorithm \ref{GD_algorithm}, we need to compute the gradient for each neuron on the $i$th hidden layer, of which time complexity is $O(\sum_{j=1}^i k_j k_{j-1})$. Thus, given an $m$-hidden-layer network, the time complexity of the gradient-based algorithm is $O(\sum_{i=1}^m k_i (\sum_{j=1}^i k_j k_{j-1}))$. This is higher than the time complexity of the Monte Carlo algorithm, while it obtains a more precise underestimated domain. We compare the efficiency and effectiveness of the two algorithms in Experiment II.


\DecMargin{1em}
\begin{algorithm}[t]
	\SetKwData{Left}{left}\SetKwData{This}{this}\SetKwData{Up}{up}
	\SetKwFunction{Union}{Union}\SetKwFunction{FindCompress}{FindCompress}
	\SetKwInOut{Input}{Input}\SetKwInOut{Output}{Output}
	\caption{The Gradient-Based Approach.}
	\label{GD_algorithm}
	\Input{$F$: a network; $x_0$: an input to $F$; $\epsilon$: a $\ell_\infty$-norm radius; $a$: the step length of gradient descent}
	\Output{$l_{L,r}^{(i)}, u_{L,r}^{(i)}$ for each  neuron $r$ in each hidden layer $i$}
	\For{each hidden layer $i={1,\ldots,m}$}{
		\For{each neuron $r$ on layer $i$}{
			Get the function $F_r^{(i)}$of neuron $r$\;
			$\eta_r^{(i)} \leftarrow sign(F_r^{(i)'}(x_0))$\tcp*{Get the sign of gradient of $r$}
			$x_{\mathit{lower}} \leftarrow x_0 - a\eta_r^{(i)} $\tcp*{One-step forward}
			Cut $x_{\mathit{lower}}$ s.t. $x_{\mathit{lower}} \in \mathbb{B}_\infty(x_0, \epsilon)$\tcp*{Make $x_{\mathit{lower}}$ valid}
			$l_{L,r}^{(i)}\leftarrow F_r^{(i)}(x_{\mathit{lower}})$\tcp*{Compute and store the lower bound}
			$x_{\mathit{upper}} \leftarrow x_0 + a\eta_r^{(i)} $\tcp*{Compute the upper case}
			Cut $x_{\mathit{upper}}$ s.t. $x_{\mathit{upper}} \in \mathbb{B}_\infty(x_0, \epsilon)$\\
			$u_{L,r}^{(i)}\leftarrow F_r^{(i)}(x_{\mathit{upper}})$ \tcp*{Compute and store the upper bound}
		}
	}
\end{algorithm}	

\vspace{-2mm}
\subsection{Over-Approximation Strategies}
\label{sec:over-approximation-strategy}

\begin{figure*}
	\centering
	\begin{subfigure}{0.24\textwidth}
		\centering 
		\includegraphics[width=0.9\textwidth]{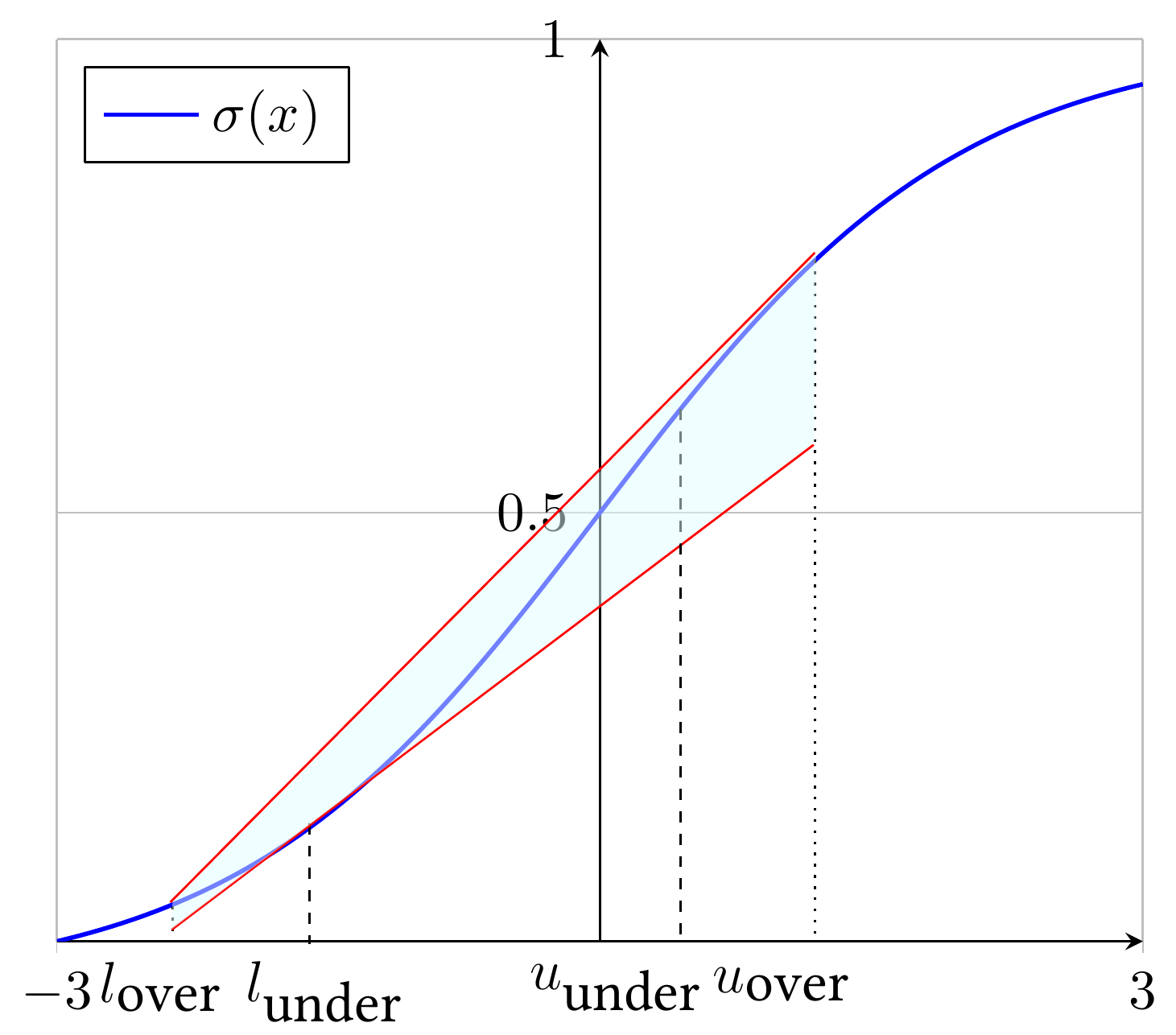}
		\caption{I: $\sigma'(l_{\mathit{over}})<k< \sigma'(u_{\mathit{over}})$.}
		\label{fig:a}
	\end{subfigure}
	\hfill
	\begin{subfigure}{0.24\textwidth}\centering
		\includegraphics[width=0.9\textwidth]{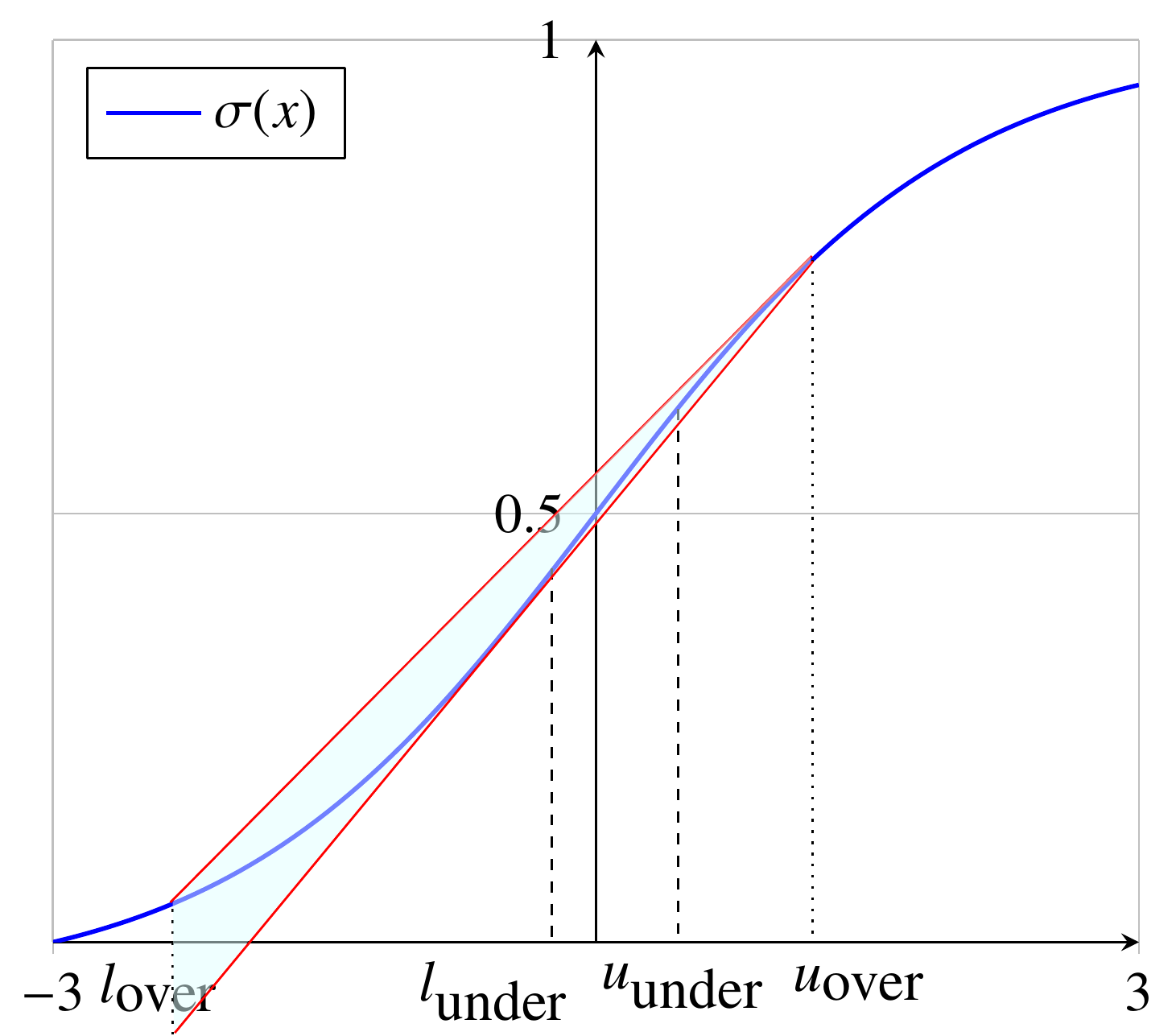}
		\caption{I: $\sigma'(l_{\mathit{over}}) < k < \sigma'(u_{\mathit{over}})$}
		\label{fig:b}
	\end{subfigure}
	\hfill
	\begin{subfigure}{0.24\textwidth}\centering
		\includegraphics[width=0.9\textwidth]{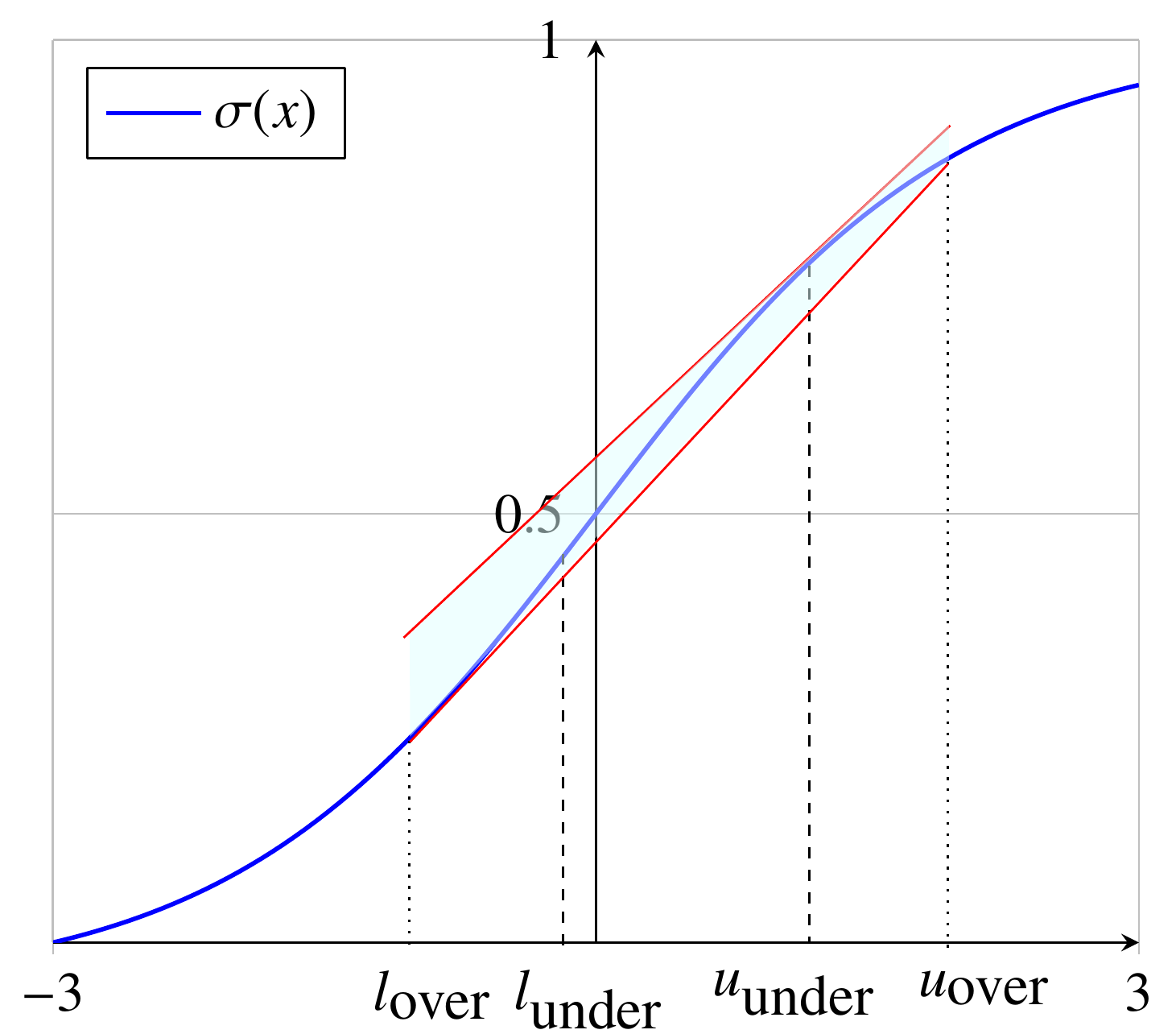}
		\caption{II: $\sigma'(l_{\mathit{over}}) > k > \sigma'(u_{\mathit{over}})$.}
		\label{fig:c}
	\end{subfigure}
	\hfill
	\begin{subfigure}{0.24\textwidth}\centering
		\includegraphics[width=0.9\textwidth]{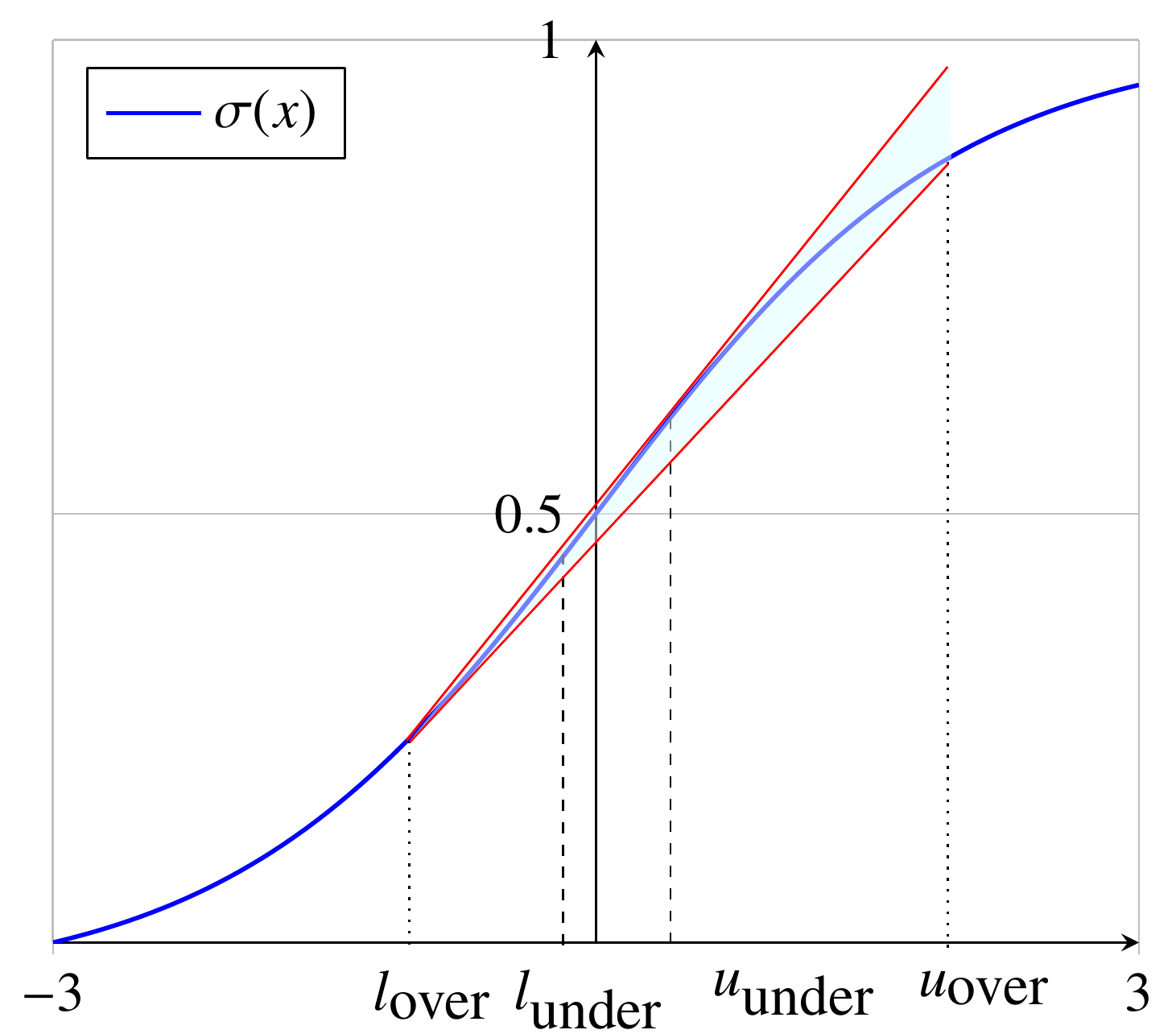}
		\caption{ II: $\sigma'(l_{\mathit{over}}) > k > \sigma'(u_{\mathit{over}})$.}
		\label{fig:d}
	\end{subfigure}
	\hfill\\
	\begin{subfigure}{0.24\textwidth}\centering
		\includegraphics[width=0.9\textwidth]{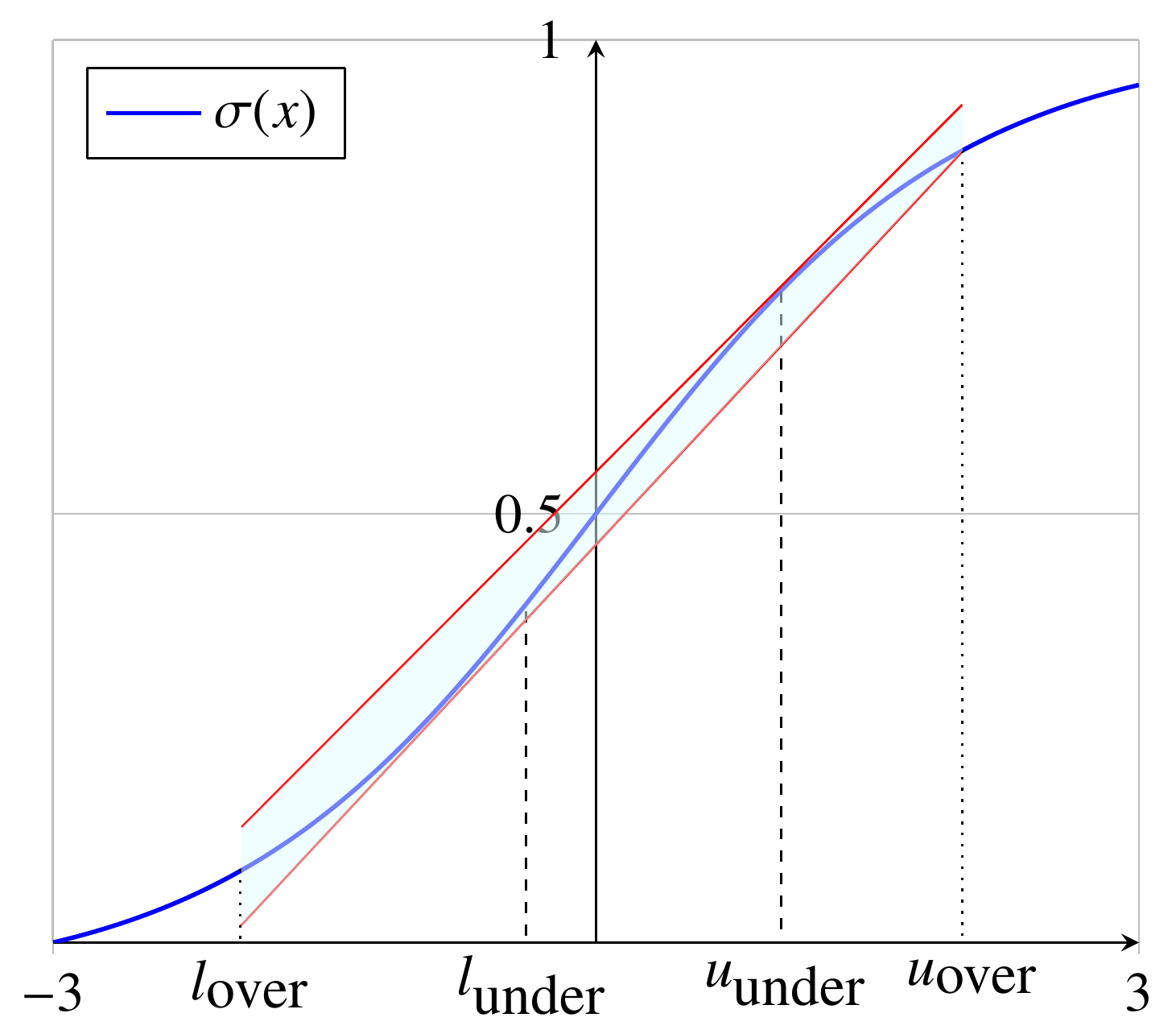}
		\caption{III: $\sigma'(l_{\mathit{over}})\!<\!k\wedge\sigma'(u_{\mathit{over}})\!<\!k$.}
		\label{fig:e}
	\end{subfigure}
	\begin{subfigure}{0.24\textwidth}\centering
		\includegraphics[width=0.9\textwidth]{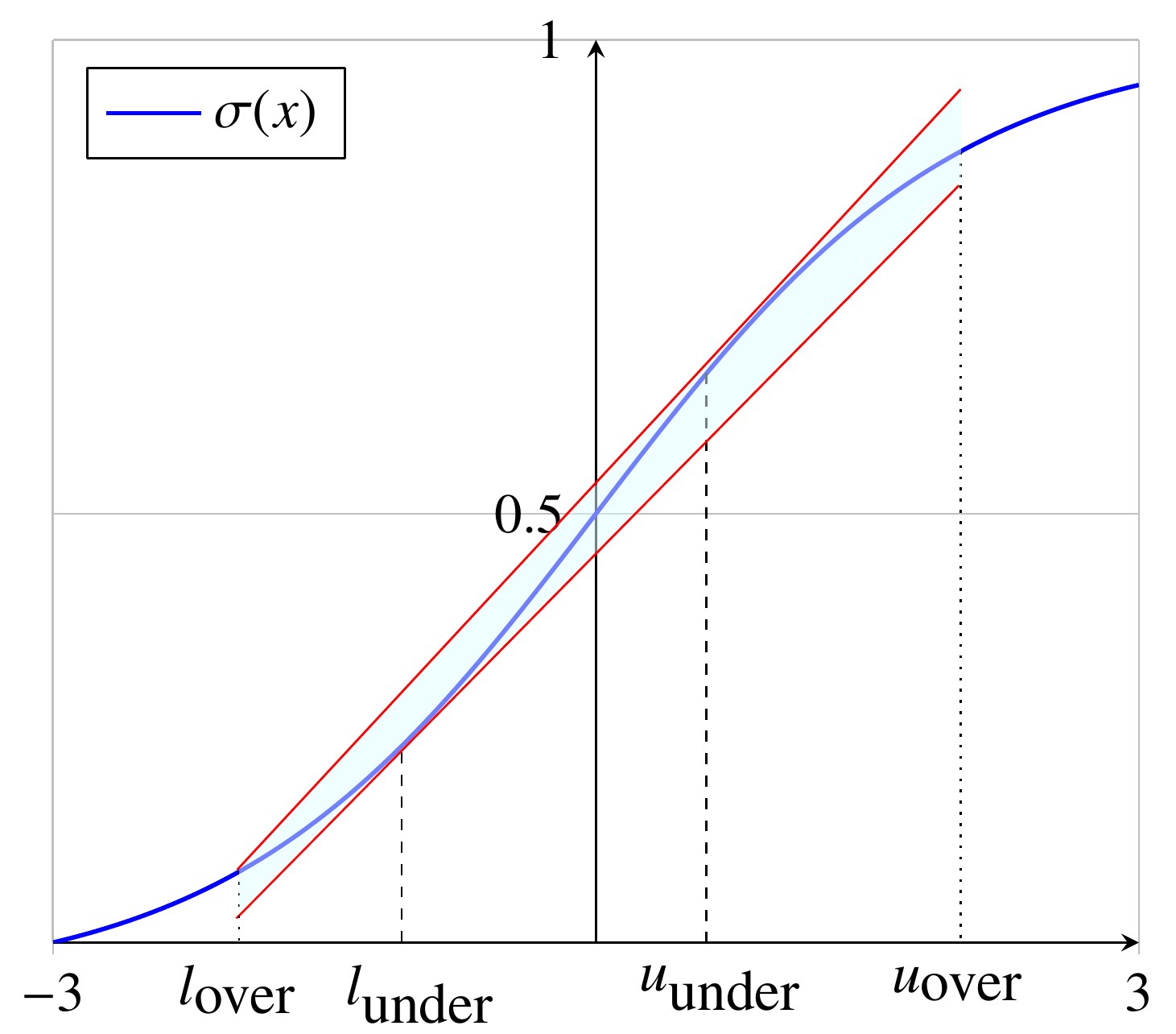}
		\caption{III: $\sigma'(l_{\mathit{over}}) \!<\! k\wedge\sigma'(u_{\mathit{over}}) \!<\! k$.}
		\label{fig:f}
	\end{subfigure}
	\hfill
	\begin{subfigure}{0.24\textwidth}\centering
		\includegraphics[width=0.9\textwidth]{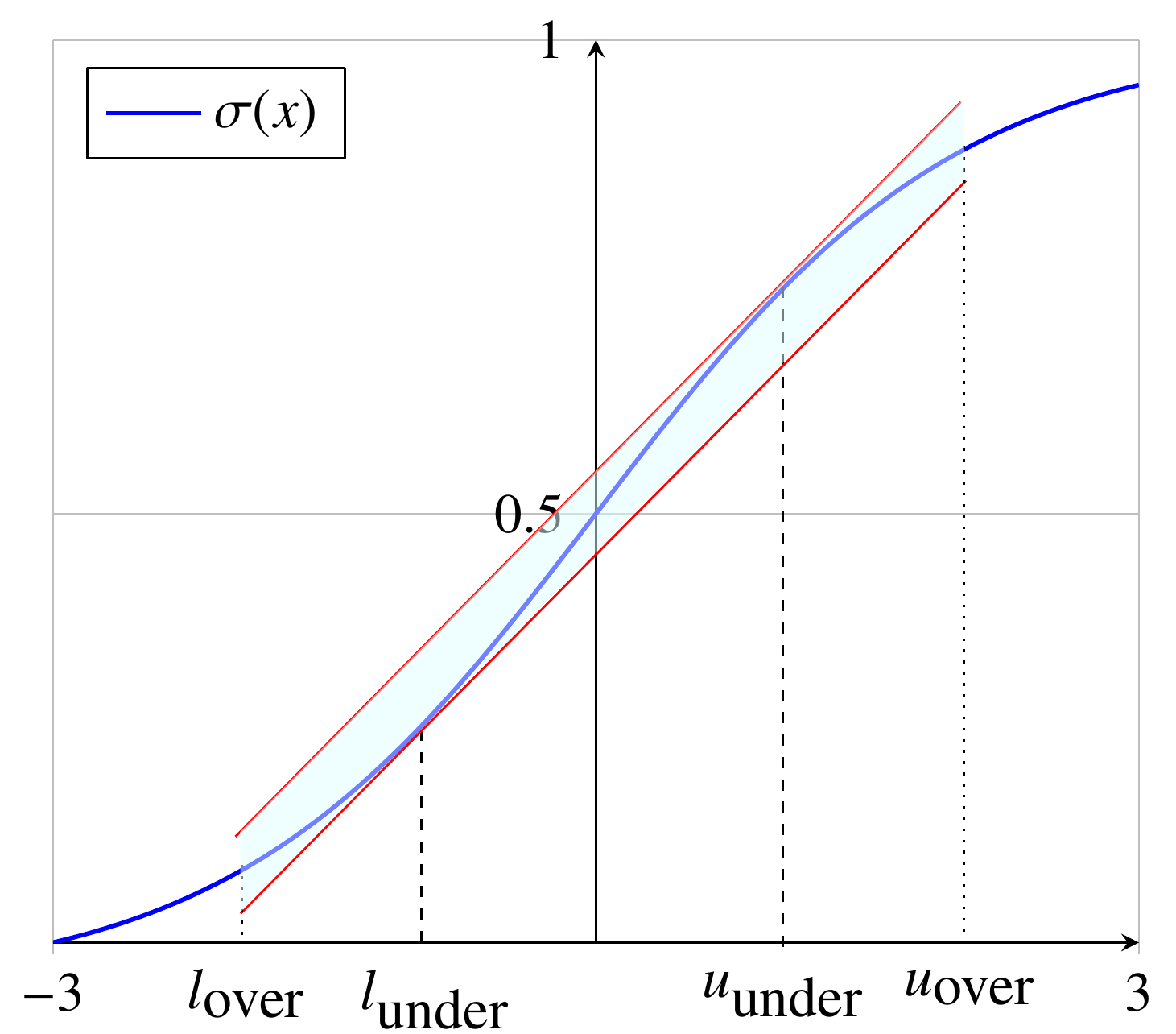}
		\caption{III: $\sigma'(l_{\mathit{over}}) \!<\! k\!\wedge\!\sigma'(u_{\mathit{over}}) \!<\! k$.}
		\label{fig:g}
	\end{subfigure}
	\hfill
	\begin{subfigure}{0.24\textwidth}\centering
		\includegraphics[width=0.9\textwidth]{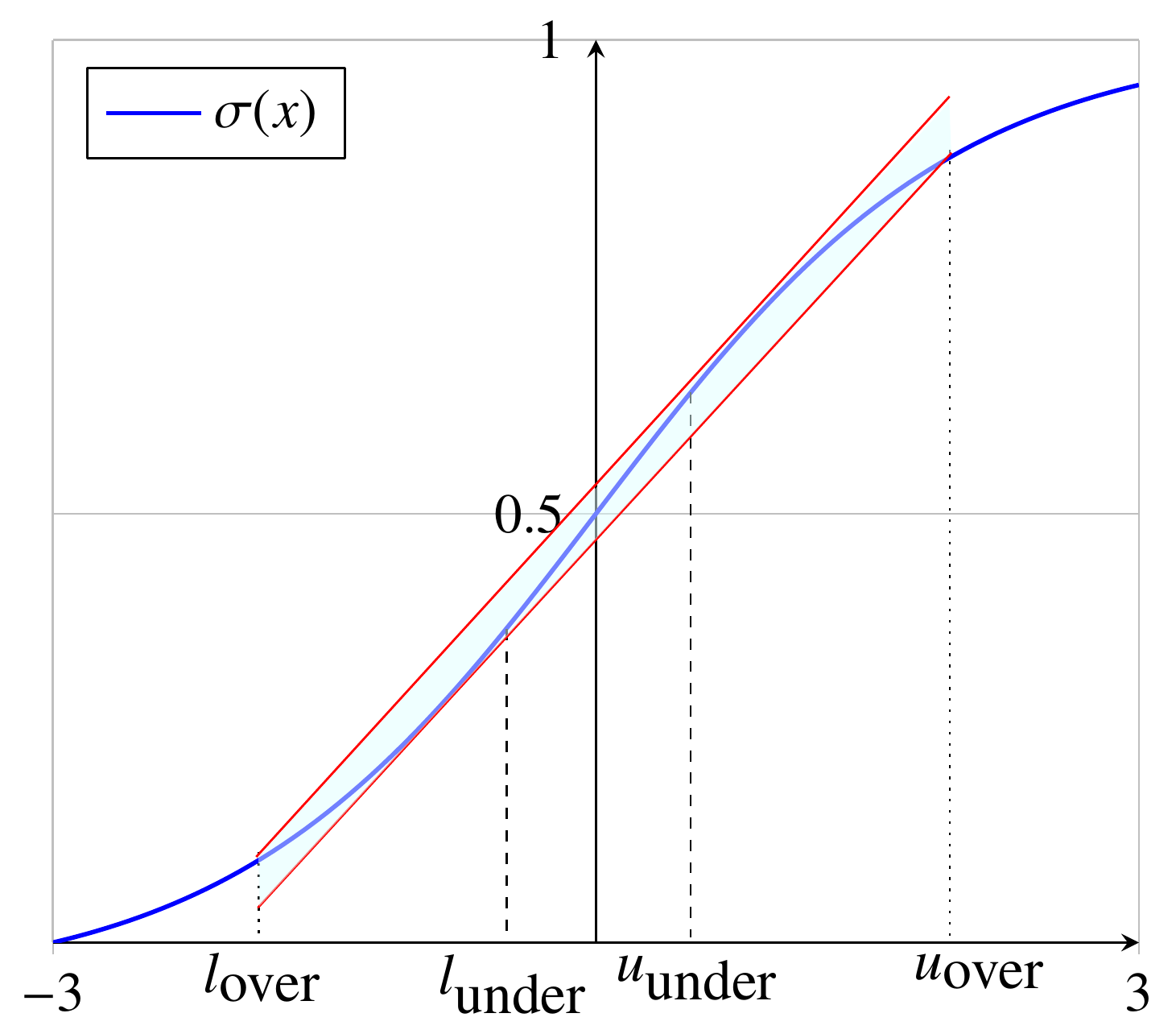}
		\caption{III: $\sigma'(l_{\mathit{over}}) \!<\! k\!\wedge\!\sigma'(u_{\mathit{over}}) \!<\! k$.}
		\label{fig:h}
	\end{subfigure}  
	\vspace{-1mm}   
	\caption{The linear over-approximation based on overestimated and underestimated approximation domains.}
		\vspace{-1mm}   
	\label{fig:figures}
\end{figure*}



We omit the superscript and subscript and consider finding the approximation method of $\sigma(x)$ with the information of upper and lower approximation domains. We assume that the lower approximation domain of input $x$ is $[l_{\mathit{under}}, u_{\mathit{under}}]$ and the upper approximation interval is $[l_{\mathit{over}}, u_{\mathit{over}}]$. As in \cite{wu2021tightening}, we consider three cases according to the relation between the slopes of $\sigma$ at the two endpoints of upper approximation interval $\sigma'(l_{\mathit{over}})$, $\sigma'(u_{\mathit{over}})$ and $k = \frac{\sigma(u_{\mathit{over}})-\sigma(l_{\mathit{over}})}{u_{\mathit{over}}-l_{\mathit{over}}}$.

\vspace{1ex}
\noindent\textbf{Case \uppercase\expandafter{\romannumeral1}.} When $\sigma'(l_{\mathit{over}}) < k < \sigma'(u_{\mathit{over}})$, the line connecting the two endpoints is the upper bound. For the lower bound, the tangent line of $\sigma$ at $l_{\mathit{under}}$ is chosen if it is sound (Figure  \ref{fig:a}), otherwise the tangent line of $\sigma$ at $d$ crossing $(u_{\mathit{over}}, \sigma(u_{\mathit{over}}))$ is chosen (Figure \ref{fig:b}). Namely, we have $h_U(x) = k(x - u_{over}) + \sigma(u_{over}) + $, and
\begin{align} 
	h_L(x) = \begin{cases}
		\sigma'(l_{\mathit{under}})(x - l_{\mathit{under}}) + \sigma(l_{\mathit{under}}), & l_{\mathit{under}} < d\\
		\sigma'(d)(x - d) + \sigma(d), & l_{\mathit{under}} \ge d.
	\end{cases}
\end{align}

\vspace{1ex}
\noindent\textbf{Case \uppercase\expandafter{\romannumeral2}.} When $\sigma'(l_{\mathit{over}}) > k > \sigma'(u_{\mathit{over}})$, it is the symmetry of Case 1. the line connecting the two endpoints can be the lower bound. For upper bound, the tangent line of $\sigma$ at $u_{\mathit{under}}$ is chosen if it is sound (Figure \ref{fig:c}), otherwise the tangent line of $\sigma$ at $d$ crossing $(l_{\mathit{under}}, \sigma(l_{\mathit{under}}))$ is chosen (Figure  \ref{fig:d}). That is, $h_L(x) = k(x - l_{over}) + \sigma(l_{over})$, and
\begin{align} 
	h_U(x) =
	\begin{cases}
		\sigma'(u_{\mathit{under}})(x - u_{\mathit{under}}) + \sigma(u_{\mathit{under}}), & u_{\mathit{under}} > d\\
		\sigma'(d)(x - d) + \sigma(d), & u_{\mathit{under}} \le d.
	\end{cases}
\end{align}

\vspace{1ex}
\noindent\textbf{Case \uppercase\expandafter{\romannumeral3}.} When $\sigma'(l_{\mathit{over}}) < k$ and $\sigma'(u_{\mathit{over}}) < k$, we first consider the upper bound. If the tangent line of $\sigma$ at $u_{\mathit{under}}$ is sound, we choose it to be the upper bound (Figure \ref{fig:e} and Fig \ref{fig:g}); otherwise we choose the tangent line of $\sigma$ at $d_1$ crossing $(l_{\mathit{under}}, \sigma(l_{\mathit{under}}))$ (Figure \ref{fig:f} and Figure \ref{fig:h}). Then we consider the lower bound. The tangent line of $\sigma$ at $l_{\mathit{under}}$ is chosen if it is sound (Figure \ref{fig:f} and Figure \ref{fig:g}), otherwise we choose the tangent line of $\sigma$ at $d_2$ crossing $(u_{\mathit{over}}, \sigma(u_{\mathit{over}}))$ (Figure \ref{fig:e} and Figure \ref{fig:h}). Namely, we have:
\begin{align} 
	h_U(x) = 
	\begin{cases}
		\sigma'(u_{\mathit{\mathit{under}}})(x - u_{\mathit{under}}) + \sigma(u_{\mathit{under}}), & u_{\mathit{under}} > d_1\\
		\sigma'(d_1)(x - d_1) + \sigma(d_1), & u_{\mathit{under}} \le d_1,
	\end{cases}\\
	h_L(x) = 
	\begin{cases}
		\sigma'(l_{\mathit{under}})(x - l_{\mathit{under}}) + \sigma(l_{\mathit{under}}), & l_{\mathit{under}} < d_2\\
		\sigma'(d_2)(x - d_2) + \sigma(d_2), & l_{\mathit{under}} \ge d_2.
	\end{cases}
\end{align}

The goal of our approximation strategy is to make the overestimated output interval as close as possible to the actual one of each hidden neuron. An over-approximation is the provably tightest neuron-wise if it preserves the same output interval as the activation function on the actual domain \cite{DBLP:conf/kbse/ZhangWLLZ22}. As described in Theorem \ref{thm:precise_domain}, a more precise range allows us to define a tighter  over-approximation. Under the premise of guaranteeing soundness, we use the guiding significance of the underestimated domain to make the over-approximation closer to the actual domain so as to obtain more precise approximation bounds. Through layer-by-layer transmission, we obtain more accurate intervals for the deeper hidden neurons (by Theorem \ref{thm:tight_approx}), on which tighter over-approximations can be defined (by Theorem \ref{thm:precise_domain}). 
In this way, the overestimation interdependency during defining over-approximations is alleviated by computing the underestimated domains. 
\iftoggle{conf-ver}{
The soundness proof for our approach is given in \cite[Appendix B]{techrep}. }
{The soundness proof for our  approach is  given in Appendix  \ref{appsec:sound}.}

\begin{figure}[t]
	\begin{center}
		\includegraphics[width=0.48\textwidth]{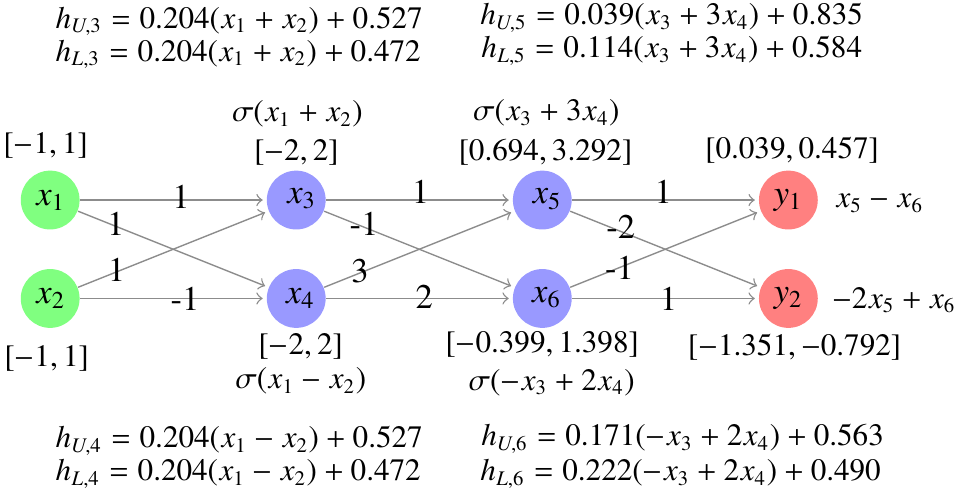}
		\caption{The approximations defined in our approach and the propagated intervals for the network in Figure \ref{fig:example1}.}
		\label{fig:approx_eg_tightest}
	\end{center}
\end{figure}

\begin{figure*}
	\centering
	\begin{subfigure}{0.17\textwidth}
		\includegraphics[width=\textwidth]{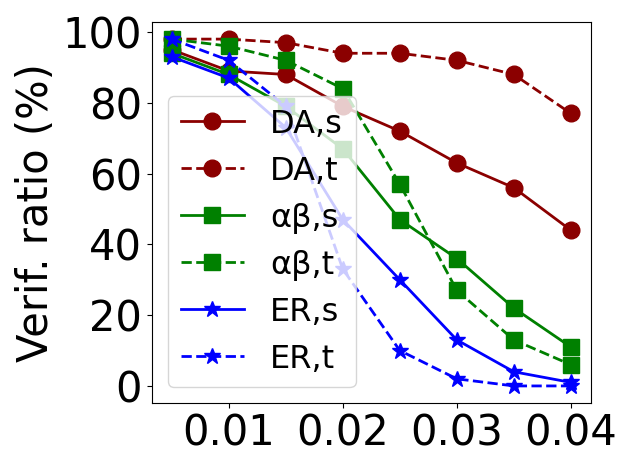}
		\caption{FC 6*500}
		\label{fig:c1}
	\end{subfigure}
	\hfill
	\begin{subfigure}{0.16\textwidth}
		\includegraphics[width=\textwidth]{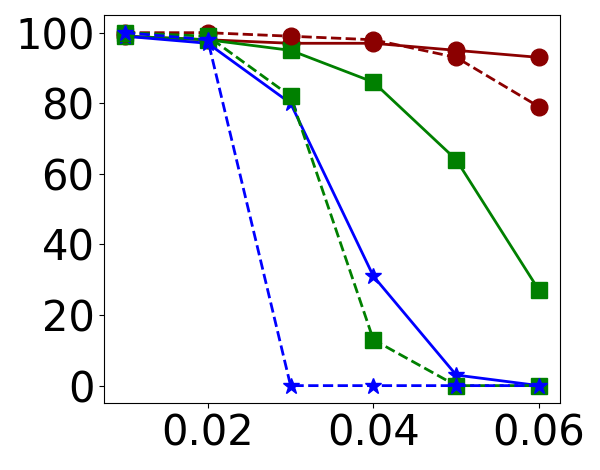}
		\caption{FC PGD 0.1}
		\label{fig:c2}
	\end{subfigure}
	\hfill
	\begin{subfigure}{0.16\textwidth}
		\includegraphics[width=\textwidth]{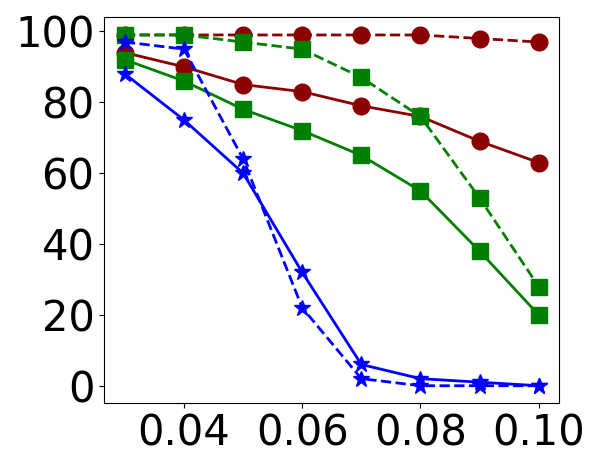}
		\caption{FC PGD 0.3}
		\label{fig:c3}
	\end{subfigure}
	\hfill
	\begin{subfigure}{0.16\textwidth}
		\includegraphics[width=\textwidth]{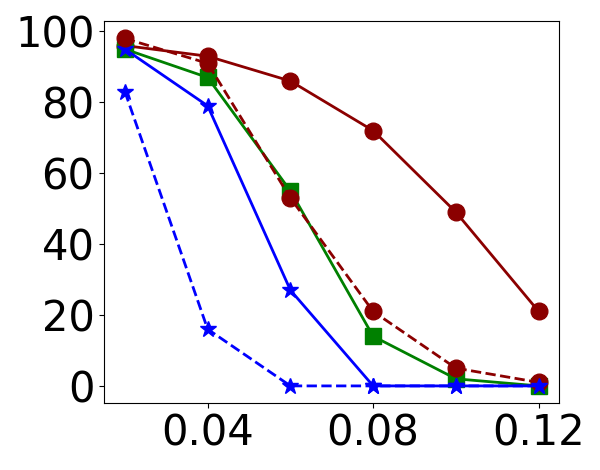}
		\caption{ConvMed}
		\label{fig:c4}
	\end{subfigure}
	\hfill
	\begin{subfigure}{0.16\textwidth}
		\includegraphics[width=\textwidth]{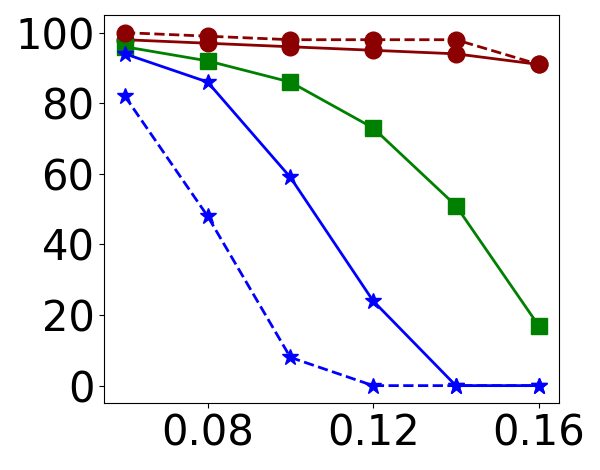}
		\caption{Conv PGD 0.1}
		\label{fig:c5}
	\end{subfigure}
	\hfill
	\begin{subfigure}{0.16\textwidth}
		\includegraphics[width=\textwidth]{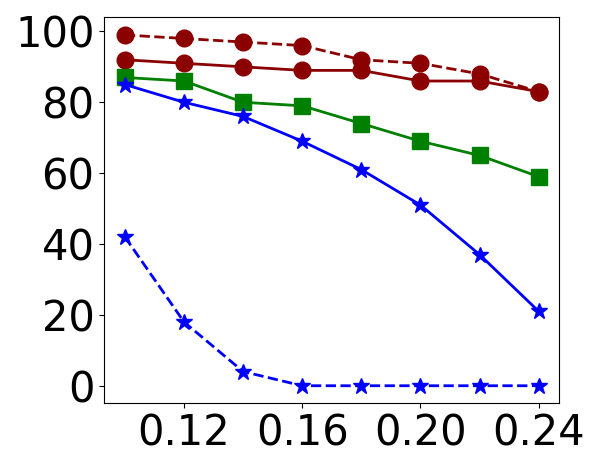}
		\caption{Conv PGD 0.3}
		\label{fig:c6}
	\end{subfigure}
	\hfill\\
	\begin{subfigure}{0.17\textwidth}
		\includegraphics[width=\textwidth]{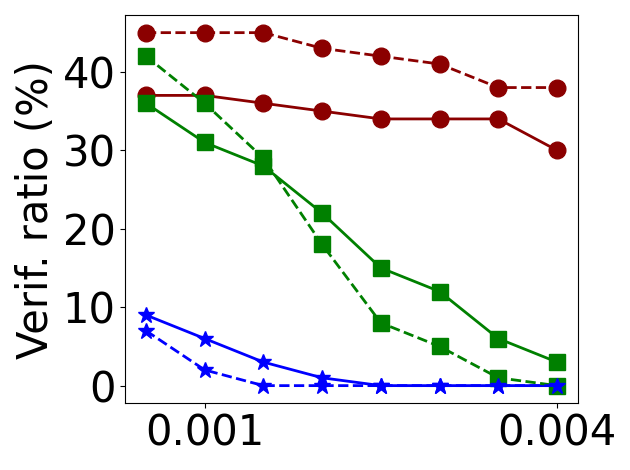}
		\caption{FC 6*500}
		\label{fig:c7}
	\end{subfigure}
	\hfill
	\begin{subfigure}{0.16\textwidth}
		\includegraphics[width=\textwidth]{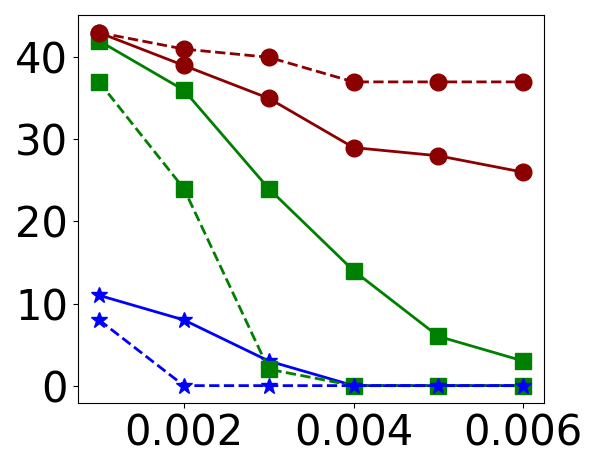}
		\caption{FC PGD 0.0078}
		\label{fig:c8}
	\end{subfigure}
	\hfill
	\begin{subfigure}{0.16\textwidth}
		\includegraphics[width=\textwidth]{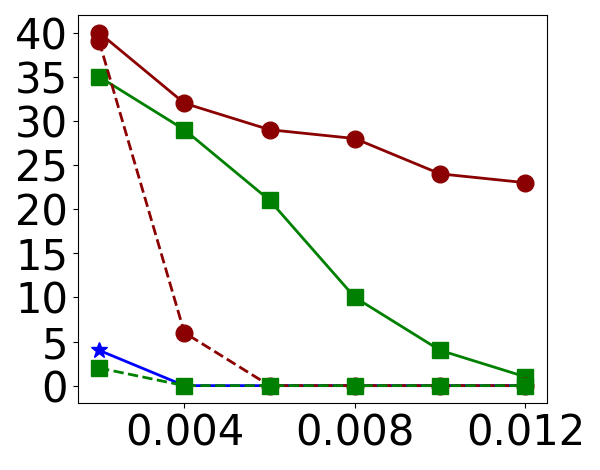}
		\caption{FC PGD 0.0313}
		\label{fig:c9}
	\end{subfigure}
	\hfill
	\begin{subfigure}{0.16\textwidth}
		\includegraphics[width=\textwidth]{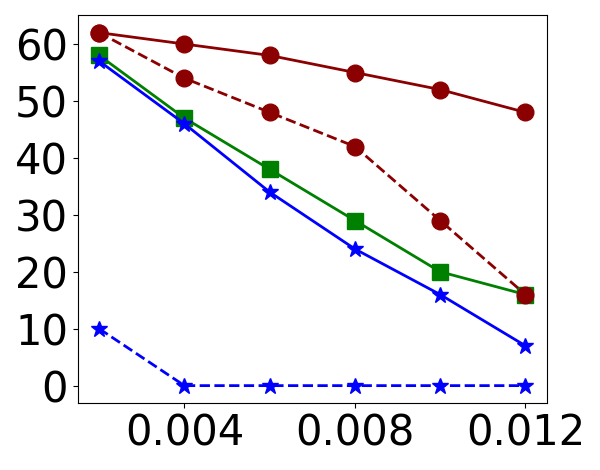}
		\caption{ConvMed}
		\label{fig:c10}
	\end{subfigure}
	\hfill
	\begin{subfigure}{0.16\textwidth}
		\includegraphics[width=\textwidth]{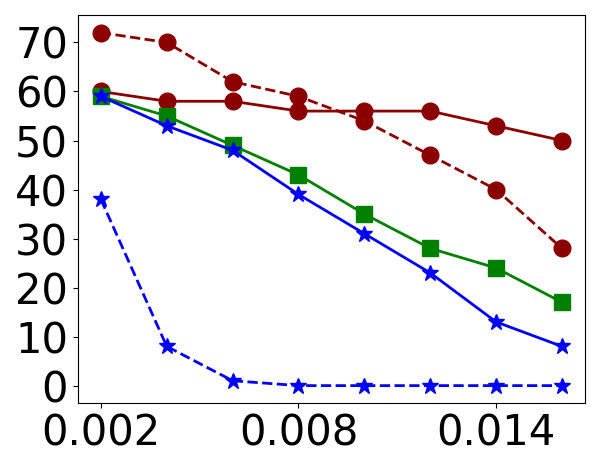}
		\caption{Conv PGD 0.0078}
		\label{fig:c11}
	\end{subfigure}
	\hfill
	\begin{subfigure}{0.16\textwidth}
		\includegraphics[width=\textwidth]{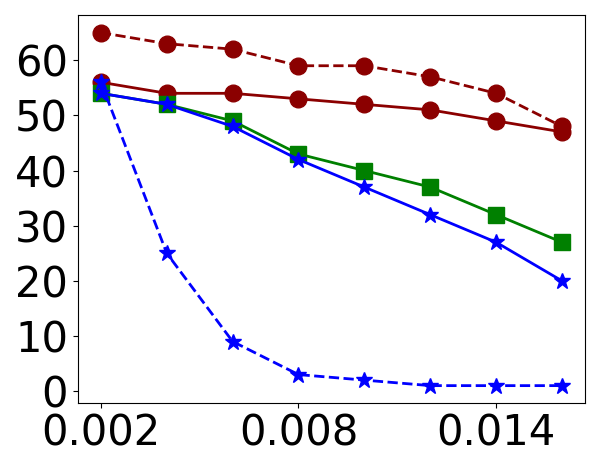}
		\caption{Conv PGD 0.0078}
		\label{fig:c12}
	\end{subfigure}
	\vspace{-1mm}
	\caption{Comparison of the verified robustness rate of DualApp, $\alpha$-$\beta$-CROWN, and ERAN on 24 neural networks and two datasets. (a-f) refer to the comparison on MNIST while (g-l) to CIFAR-10. The horizontal axis represents the perturbation $\epsilon$, and the vertical axis represents the ratio of images that are verified robustness.  Legend $DA,s$ refers to the result of DualApp on the neural network with sigmoid activation function, and $ER,t$ refers to the result of ERAN with tanh.}
	\label{fig:cmp_ab_er}
\end{figure*}

\begin{example}
    \label{exm:2}
We reconsider the network in Figure \ref{fig:example1}. Figure \ref{fig:approx_eg_tightest} shows the approximations and the propagated intervals for neurons in hidden layers and the output layer. As $x_3,x_4$ have precise input intervals, only the activation functions of $x_5,x_6$ need to be under-approximated. Thus, we only need to redefine  their approximations according to our approach. We underestimate the input domains of $x_5,x_6$ and use them to guide the over-approximations in our approach. We achieve $9.74\%$ and $0.27\%$ reductions for the overestimation of $y_1,y_2$'s output ranges.  

\end{example}

Example \ref{exm:2} demonstrates the effectiveness of our proposed dual-approximation approach even when it is applied to only one hidden layer. That is, the overestimation of propagated intervals can be further reduced through defining tighter over-approximations based on both underestimated and approximation domains. Tighter over-approximations can produce more precise verification results, and we will show this experimentally in the next section.


%% file: 5experiment.tex
\section{Implementation and Evaluation}

\label{sec:experiment}

We implement our dual-approximation approach into a tool called DualApp. 
We assess it, along with six 
state-of-the-art tools, 
with respect to the 
 DNNs with the S-curved activation functions. Our goal is threefold:

\begin{enumerate}[(1)]
	\item to demonstrate that, compared to the state of the art, DualApp is more tight on robustness verification results; 
 \item to explore the hyper-parameter 
 space  
  for our two  methods that leverage under-approximation; and 
        \item to measure
        the trade-off between these two complementary 
       under-approximation methods.

\end{enumerate}



\vspace{-2mm}
\subsection{Benchmark and Implementation}

\textbf{Competitors.}
We consider six state-of-the-art DNN robustness verification tools: 
 $\alpha$-$\beta$-CROWN~\cite{DBLP:journals/corr/abs-2208-05740}, ERAN~\cite{muller2022prima}, NeWise~\cite{DBLP:conf/kbse/ZhangWLLZ22}, DeepCert~\cite{wu2021tightening}, VeriNet~\cite{HenriksenL20}, and RobustVerifier~\cite{lin2019robustness}. They rely on the over-approximation domain
 to define and optimize linear lower and upper bounds except that ERAN is based on the abstract domain combining floating point polyhedra with intervals \cite{singh2019abstract}. 

\vspace{1ex}
\noindent
\textbf{Datasets and Neural Networks.}
In the comparison experiment with $\alpha$-$\beta$-CROWN and ERAN, we collect 24 models exposed by ERAN for MNIST~\cite{DBLP:journals/pieee/LeCunBBH98} and CIFAR-10~\cite{krizhevsky2009learning}, including CNNs and FNNs trained with normal training method and adversarial training method with PGD attack~\cite{DBLP:conf/corl/WangZPL21}. These models contain Sigmoid and Tanh activation functions. For NeWise, DeepCert, VeriNet, and RobustVerifier, we collect and train totally 84 convolutional neural networks (CNNs) and fully-connected neural networks (FNNs) on image datasets MNIST, Fashion MNIST~\cite{xiao2017/online} and CIFAR-10. Sigmoid, Tanh, and Arctan are contained in these models, respectively. As most researches are based on the ReLU activation function, there are few public neural networks in benchmarks with S-curved activate functions, and their sizes are small.

\vspace{1ex}
\noindent
\textbf{Metrics. }
We use two metrics in our comparisons: (\romannumeral1) \emph{verified robustness ratio}, which is the percentage of images that must be correctly classified under a fixed perturbation $\epsilon$, and (\romannumeral2) \emph{certified lower bound}, which is the largest perturbation $\epsilon$ within which all input images must be correctly classified. 
We consider strong baselines in that we assess DualApp on the benchmarks and metrics for which the competitors report the optimal performance. 
In particular,
we use (i) for the comparison with $\alpha$-$\beta$-CROWN and ERAN as both  report the highest
verified robustness rate as in, e.g.,
 the 2022 VNN-COMP competition~\cite{muller2022third}.

\vspace{1ex}
\noindent
\textbf{Implementation.}
For a fair comparison, we implemented the approximation strategies of NeWise, DeepCert, VeriNet, RobustVerifier, and our dual-approximation algorithm in DualApp using python with the TensorFlow framework. We apply the method used in NeWise to train neural networks and load datasets. We implement the algorithms and strategies defined in Section \ref{sec:method} to compute under-approximation domains and linear upper and lower bounds, thus obtaining the final output intervals and verification results for each image. To compute the certified lower bound, we set the initial value of $\epsilon$ to $0.05$ and update it $15$ times using the dichotomy method based on the verification results.

\vspace{1ex}
\noindent
\textbf{Experimental Setup.}
We conducted all the experiments on a workstation equipped with a 32-core AMD Ryzen Threadripper PRO 5975WX CPU, 256GB RAM, and an Nvidia RTX 3090Ti GPU running Ubuntu 22.04.

\vspace{-2mm}
\subsection{Experimental Results}

\begin{table}[t]
    \centering
    \caption{Comparison of the verification time  among DualApp, $\alpha$-$\beta$-CROWN, and ERAN on 24 neural networks and two datasets. We report the average time of verifying 100 images on 6 neural networks with the same architecture used in Figure \ref{fig:cmp_ab_er}, e.g., the time of FC models on MNIST refers to the experiments in Figure \ref{fig:c1}-\ref{fig:c3}. }
    \label{tab:time}
    \resizebox{\linewidth}{!}{
    \begin{tabular}{|c|c|c|c|r|}
    \hline
    \textbf{Dataset} & \textbf{Model} & \textbf{DualApp} & \textbf{$\alpha$-$\beta$-CROWN} & \textbf{ERAN} \\
    \hline
    \multirow{2}{*}{MNIST} & FC & 3.81s & 2.30s & 14.39s \\
    \cline{2-5}
     & CONV & 1.09s & 2.25s & 3.30s \\
     \hline
     \multirow{2}{*}{CIFAR-10} & FC & 3.12s & 4.46s & 34.16s \\
     \cline{2-5}
      & CONV & 1.81s & 0.88s & 6.46s \\
      \hline
    \end{tabular}
    }
\end{table}

\begin{table*}[t]
	\centering
	\caption{Comparing the DualApp (DA) and four state-of-the-art tools including NeWise (NW), DeepCert (DC), VeriNet (VN), and RobustVerifier (RV) on the CNNs and FNNs with the Sigmoid activation function. ${\rm CNN}_{l-k}$ denotes a CNN with $l$ layers and  $k$ filters of size $3 \times 3$ on each layer. ${\rm FNN}_{l\times k}$ denotes a FNN with $l$ layers and $k$ neurons on each layer.}
	\vspace{-1mm}
	\label{tab:sigmoid}
	\resizebox{\linewidth}{!}{
		\begin{tabular}{|c|c|r|r|r|r|r|r|r|r|r|r|r|r|}
			\hline
			\multirow{2}{*}{\textbf{Dataset}}                                                 & \multirow{2}{*}{\textbf{Model}} & \multirow{2}{*}{\textbf{Nodes}} & \textbf{DA} & \multicolumn{2}{c|}{\textbf{NW}}              & \multicolumn{2}{c|}{\textbf{DC}}            & \multicolumn{2}{c|}{\textbf{VN}}             & \multicolumn{2}{c|}{\textbf{RV}}      & \multirow{2}{*}{\begin{tabular}[c]{@{}c@{}}\textbf{DA} \\ Time (s)\end{tabular}} & \multirow{2}{*}{\begin{tabular}[c]{@{}c@{}}\textbf{Others}\\ Time (s)\end{tabular}} \\ \cline{4-14}
			&                        &                        &  {Bounds}  & {Bounds}  & Impr. (\%) & {Bounds}  & Impr. (\%)  & {Bounds}  & Impr. (\%) & {Bounds}  & Impr. (\%) &                                                                               &                                 \\ \hline
			\multirow{7}{*}{Mnist}                                                    & ${\rm CNN}_{4-5}$      & 8,690                  & 0.05819  & {0.05698} & 2.12      & {0.05394} & 7.88      & {0.05425} & 7.26      & {0.05220} & 11.48     & 14.70                                                                         & 0.98   $\pm$ 0.02  \\
			& ${\rm CNN}_{5-5}$      & 10,690                 & 0.05985  & {0.05813} & 2.96      & {0.05481} & 9.20      & {0.05503} & 8.76      & {0.05125} & 16.78     & 20.13                                                                         & 2.67   $\pm$ 0.29  \\
			& ${\rm CNN}_{6-5}$      & 12,300                 & 0.06450   & {0.06235} & 3.45      & {0.05898} & 9.36      & {0.05882} & 9.66      & {0.05409} & 19.25     & 25.09                                                                         & 4.86   $\pm$ 0.34  \\
			& ${\rm CNN}_{8-5}$     & 14,570                 & 0.11412  & {0.09559} & 19.38     & {0.08782} & 29.95     & {0.08819} & 29.40     & {0.06853} & 66.53     & 34.39                                                                         & 11.89   $\pm$ 0.21 \\
			& ${\rm FNN}_{5\times 100}$                  & 510                    & 0.00633  & {0.00575} & 10.09     & {0.00607} & 4.28      & {0.00616} & 2.76      & {0.00519} & 21.97     & 7.10                                                                           & 0.79   $\pm$ 0.05  \\
			& ${\rm FNN}_{6\times 200}$                  & 1,210                  & 0.02969  & {0.02909} & 2.06      & {0.02511} & 18.24     & {0.02829} & 4.95      & {0.01811} & 63.94     & 8.64                                                                         & 2.82   $\pm$ 0.34  \\ \hline
			\multirow{7}{*}{\begin{tabular}[c]{@{}l@{}}Fashion \\ Mnist\end{tabular}} & ${\rm CNN}_{4-5}$      & 8,690                  & 0.07703  & {0.07473} & 3.08      & {0.07204} & 6.93      & {0.07200} & 6.99      & {0.06663} & 15.61     & 15.26                                                                         & 1.06   $\pm$ 0.09  \\
			& ${\rm CNN}_{5-5}$      & 10,690                 & 0.07288  & {0.07044} & 3.46      & {0.06764} & 7.75      & {0.06764} & 7.75      & {0.06046} & 20.54     & 20.95                                                                         & 3.18   $\pm$ 0.42  \\
			& ${\rm CNN}_{6-5}$      & 12,300                 & 0.07655  & {0.07350}  & 4.15      & {0.06949} & 10.16     & {0.06910} & 10.78     & {0.06265} & 22.19     & 25.96                                                                         & 5.63   $\pm$ 0.77  \\
			& ${\rm FNN}_{1\times 50}$                   & 60                     & 0.03616  & {0.03284} & 10.11     & {0.03511} & 2.99      & {0.03560} & 1.57      & {0.02922} & 23.75     & 0.84                                                                          & 0.02   $\pm$ 0.00    \\
			& ${\rm FNN}_{5\times 100}$                  & 510                    & 0.00801  & {0.00710}  & 12.82     & {0.00776} & 3.22      & {0.00789} & 1.52      & {0.00656} & 22.10     & 2.98                                                                          & 0.65   $\pm$ 0.00     \\ \hline
			\multirow{5}{*}{Cifar-10}                                                 & ${\rm CNN}_{3-2}$    & 2,514                  & 0.03197  & {0.03138} & 1.88      & {0.03120} & 2.47      & {0.03119} & 2.50      & {0.03105} & 2.96      & 5.54                                                                          & 0.32   $\pm$ 0.02  \\
			& ${\rm CNN}_{5-5}$     & 10,690                 & 0.01973  & {0.01926} & 2.44      & {0.01921} & 2.71      & {0.01913} & 3.14      & {0.01864} & 5.85      & 31.45                                                                         & 4.86   $\pm$ 0.41  \\
			& ${\rm CNN}_{6-5}$      & 12,300                 & 0.02338  & {0.02289} & 2.14      & {0.02240} & 4.38      & {0.02234} & 4.66      & {0.02124} & 10.08     & 43.51                                                                         & 10.53   $\pm$ 0.67 \\
			& ${\rm FNN}_{5\times 100}$                 & 510                    & 0.00370  & {0.00329} & 12.46      & {0.00368} & 0.54      & {0.00368} & 0.54      & {0.00331} & 11.78      & 2.97                                                                          & 0.64   $\pm$ 0.01  \\
			& ${\rm FNN}_{3\times 700}$                    & 2,110                  & 0.00428  & {0.00348}  & 22.99      & {0.00427} & 0.23     & {0.00426} & 0.47      & {0.00397} & 7.81      & 32.68                                                                         & 10.85   $\pm$ 0.58 \\ \hline
		\end{tabular}
	}
	\label{exp_only_sample}
\end{table*}

\noindent \textbf{Experiment \uppercase\expandafter{\romannumeral1}: Comparisons with the Competitors.}
Figure \ref{fig:cmp_ab_er} shows the comparison results among our approach with the Monte Carlo algorithm, $\alpha$-$\beta$-CROWN, and ERAN on 24 networks with Sigmoid and Tanh activation functions. In this experiment, the perturbation $\epsilon$ is set to increase in a fixed step for each network. Following the strategy for choosing image samples in all the competing tools, we choose the first 100 images from the corresponding test set to verify. We take 1000 samples for each image to compute the underestimated domain in our method. The experimental results strongly show our method can reduce overestimation and compute higher verified robustness ratios.
In most cases, our method improves by \emph{dozens to hundreds of percent} compared with the other two methods. In particular, the improvement becomes significantly greater as the perturbation $\epsilon$ increases. For example, in Figure \ref{fig:c2}, the improvement of our method relative to $\alpha$-$\beta$-CROWN is $12.79\%$ when $\epsilon = 0.04$ on Sigmoid networks, and the number reaches $244.44\%$ when $\epsilon$ enlarge to $0.06$. In Figure \ref{fig:c12}, the improvement even reaches $5600\%$ compared with ERAN on Tanh networks when $\epsilon = 0.012$.  
That is because large perturbations imply large input intervals and consequently large overestimation of approximation domains. The underestimated domains become more dominant in defining tight over-approximations. These experimental results further demonstrate the importance of underestimated domains in tightening the over-approximations.

Regarding the verification efficiency, Table \ref{tab:time} shows the verification time of the experiments in Figure \ref{fig:cmp_ab_er}. We observe that DualApp is more efficient than ERAN on all the experiments. The verification time of DualApp is similar to that of 
$\alpha$-$\beta$-CROWN, with each being more efficient on half of the experiments. The extra time by DualApp is spent on computing underestimated domains.


Table \ref{tab:sigmoid} shows the comparison results between our approach with the Monte Carlo algorithm and the other four tools on 16 networks with Sigmoid activation function. We randomly choose 100 inputs from each test set and compute the average of their certified lower bounds. In our method, we take 1000 samples for each image to compute the underestimated domain. The result shows that our approach outperforms all four competitors in all cases. On average, the improvement of our approach achieves $10.64\%$ compared to others.
Regarding efficiency, our Monte Carlo approach takes a little more time than other tools because of the sampling procedure. We trade time for a more precise approximation.
For Tanh and Arctan neural networks, our approach also performs best on these models among all the tools. 
\iftoggle{conf-ver}{We defer the results to 
\cite[Appendix C.1]{techrep}.}{
We defer the results to Appendix \ref{sec:add_exp1}.}

To conclude, the experimental results demonstrate that, compared with those approaches that rely  only on approximation domains, our dual-approximation approach introduces less overestimation and consequently returns more precise verification results on both robustness rates and certified bounds. The improvement is even more significant with larger perturbations. That is because larger perturbations cause more overestimation to approximation domains, which in return make over-approximations less tight.



\begin{figure}
	\centering
	\begin{subfigure}{0.235\textwidth}
		\includegraphics[width=\textwidth]{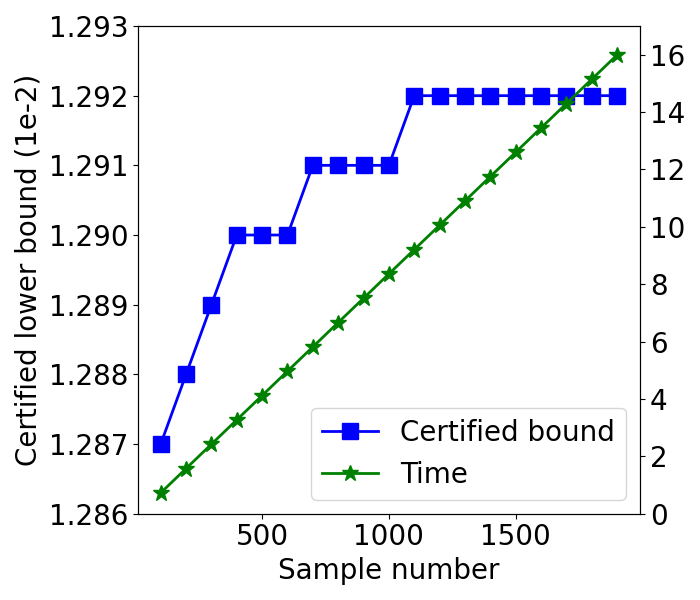}
		\caption{Monte Carlo algorithm}
		\label{fig:Sampling_fashion_mnist_fnn}
	\end{subfigure}
	\begin{subfigure}{0.235\textwidth}
		\includegraphics[width=\textwidth]{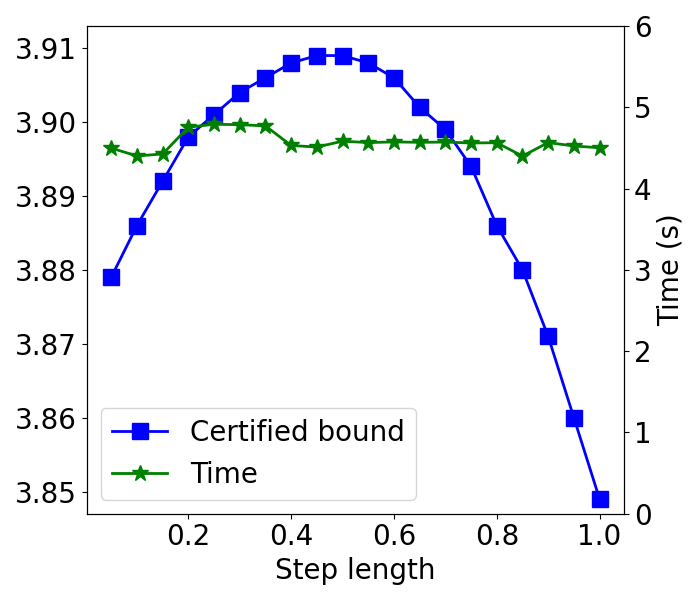}
		\caption{Gradient-based algorithm}
		\label{fig:GD_fashion_mnist_fnn}
	\end{subfigure}
		\vspace{-4mm}
	\caption{Exploring the influence of sample number and step length on certified lower bounds. 
	\vspace{-1mm}	
	}

	\label{GDSampling}
\end{figure}

\begin{figure*}
	\centering
	\begin{subfigure}{0.23\textwidth}
		\includegraphics[width=0.9\textwidth]{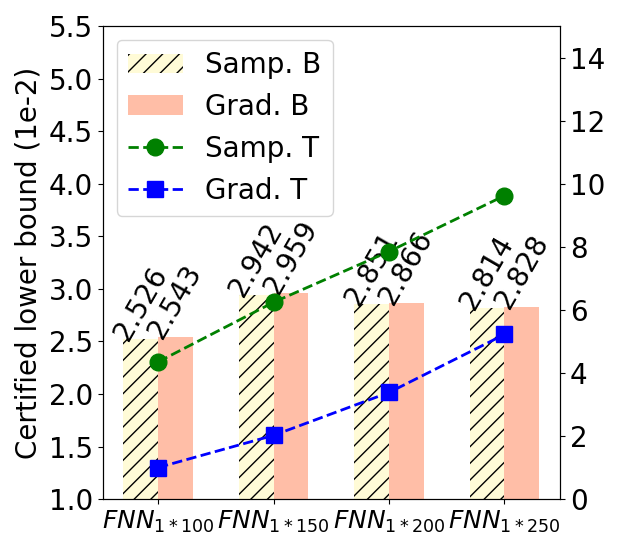}  
		\caption{FNNs on MNIST.}
		\label{fig:GDSampling_mnist_fnn}
	\end{subfigure}
	\hfill
	\begin{subfigure}{0.23\textwidth}
		\includegraphics[width=0.9\textwidth]{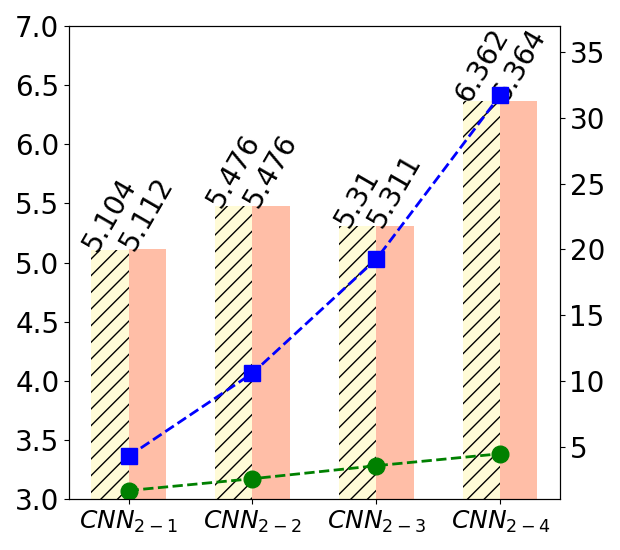}
		\caption{CNNs on MNIST.}
		\label{fig:GDSampling_mnist_cnn}
	\end{subfigure}
	\hfill
	\begin{subfigure}{0.23\textwidth}
		\includegraphics[width=0.9\textwidth]{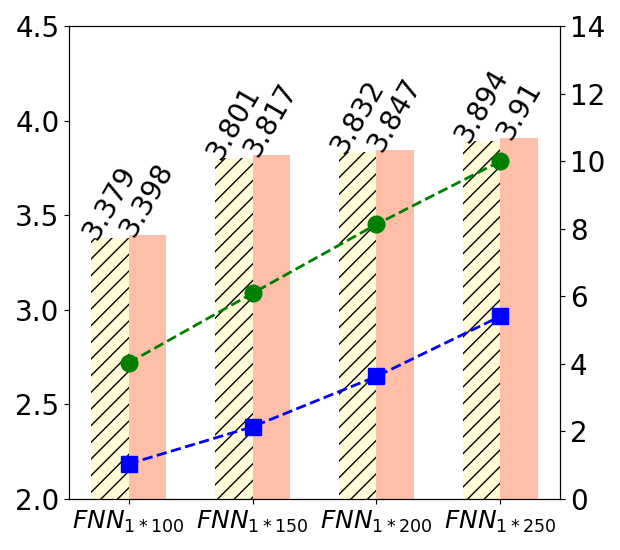}  
		\caption{FNNs on Fashion MNIST.}
		\label{fig:GDSampling_fmnist_fnn}
	\end{subfigure}
	\hfill
	\begin{subfigure}{0.23\textwidth}
		\includegraphics[width=0.9\textwidth]{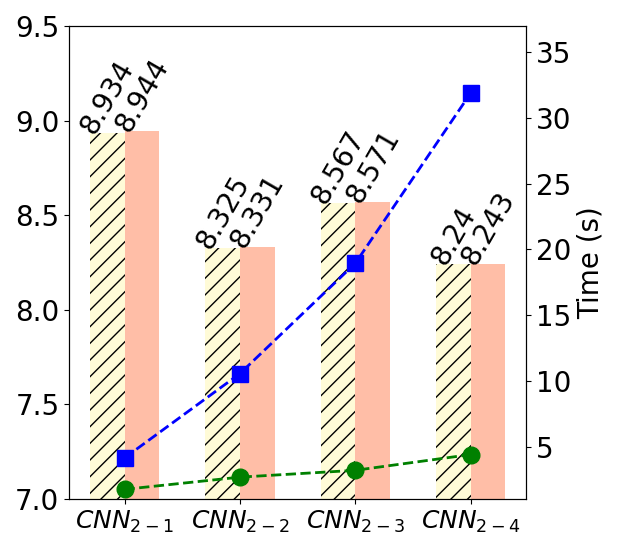}
		\caption{FNNs on Fashion MNIST.}
		\label{fig:GDSampling_fmnist_cnn}
	\end{subfigure}
	\vspace{-1mm}
	\caption{Comparison between the Monte Carlo and gradient-based algorithms for the robustness verification on 4 FNNs and 4 CNNs. The sample number is set to $1000$ while the step length to $0.45$.}
	\vspace{-1mm}
	\label{fig:exp2}
\end{figure*}

\vspace{1ex}
\noindent
\textbf{Experiment \uppercase\expandafter{\romannumeral2}: Hyper-parameters.}
As  approximation-based verification is intrinsically incomplete and the optimal values of hyper-parameters are unknowable, it is important to explore the
hyper-parameter space  for more effective and efficient verification.
Hence,  we measure the impacts of the two hyper-parameters, i.e., the sample number and the step length of gradient descent, in our Monte Carlo and gradient-based algorithms, respectively. They are quantitatively measured with respect to the certified lower bound and verification time. A larger lower bound implies a better hyper-parameter setting with respect to verification results. 


We conduct the experiments on eight neural networks trained on MNIST and Fashion MNIST, respectively. Figure \ref{fig:Sampling_fashion_mnist_fnn} shows the relation between certified lower bounds and the number of samples, resp. the time cost, for an FNN$_{1*150}$ trained on Fashion MNIST. 
The computed bound is monotonously increasing with more samples, 
and stabilizes around $1000$ samples, which indicates that the under-approximation domain cannot be improved by simply running more simulations. Figure \ref{fig:GD_fashion_mnist_fnn} shows the result of the gradient-based algorithm for the same FNN$_{1*150}$. We consider the step length from $0.05\epsilon$ to $\epsilon$ by step of $0.05\epsilon$. It indicates that when the step length is set around $0.45\epsilon$, the computed bound is maximal. 

As for the efficiency, we observe a linear relationship between the number of samples and the overhead for the Monte Carlo method. In contrast, the time cost is almost the same and independent of the step length. This conclusion is applicable to other networks and perturbation $\epsilon$.  
\iftoggle{conf-ver}{
 The complete experimental results are provided in Appendix C.2 in our technical report \cite{}. }
 {
 The complete experimental results are provided in Appendix \ref{sec:add_exp3}.
 }

\vspace{1ex}
\noindent \textbf{Experiment \uppercase\expandafter{\romannumeral3}: 
Monte Carlo vs. Gradient. }
We evaluate the performance of our two under-approximation algorithms. Figure~\ref{fig:exp2} shows the certified lower bounds and the time cost of two algorithms on eight FNNs and eight CNNs with the Sigmoid activation function, respectively. We set the sample number to $1000$ and step length to $0.45$ (the optimal hyper-parameters from Experiment \uppercase\expandafter{\romannumeral2}). We observe that, compared to the Monte Carlo algorithm,  the certified lower bounds computed by the gradient-based algorithm are always larger. The reason is that local information of neural network functions such as monotonicity can be obtained through gradients for computing more precise underestimated domains. 
This further demonstrates the usefulness of underestimated domains in defining tight over-approximations.

The experimental results also show that, the gradient-based algorithm costs less time on simple FNNs, while  more time on complex CNNs. With small-sized neural networks, the gradient-based algorithm is faster as fewer steps are required in computing the gradient. 

%% file: 6relatedwork.tex
\section{Related Work}

\label{sec:related_work}

This work is a sequel to existing efforts on approximation-based DNN robustness analysis. We classify them into two categories. 

\vspace{1ex}
\noindent \textbf{Over-approximation  verification approaches.}
Due to the intrinsic complexity in the neural network robustness verification, approximating the non-linear activation functions is the mainstreaming approach for scalability. Zhang \emph{et al.} defined three cases for over-approximating S-curved activation functions \cite{zhang2018efficient}. Wu and Zhang proposed a fine-grained approach and identified five cases for defining tighter approximations \cite{wu2021tightening}. Lyu \emph{et al.} proposed to define tight approximations by optimization at the price of sacrificing efficiency. Henriksen and Lomuscio \cite{HenriksenL20} defined tight approximations by minimizing the gap area between the bound and the curve. However, all these approaches are proved superior to others only on specific networks~\cite{DBLP:conf/kbse/ZhangWLLZ22}. The approximation approach  in~\cite{DBLP:conf/kbse/ZhangWLLZ22} is proved to be the tightest when the networks are monotonous. All these approaches only consider overestimated approximation domains, and thus they are superior to each other only on some selected network models. It is also unclear under what conditions these approaches except \cite{DBLP:conf/kbse/ZhangWLLZ22} are theoretically better than others. We believe that, the under-estimated domains could provide useful information to these approaches for defining tighter over-approximations. 

Paulsen and Wang recently proposed an interesting approach for synthesizing tight approximations guided by generated examples \cite{paulsen2022example,paulsen2022linsyn}. 
Similar to our approach, these approaches compute sound and tight over-approximations from unsound templates. However, they require global optimization techniques to guarantee soundness, which is very time-consuming, while our approach ensures the soundness of individual neurons statistically. 

\vspace{1ex}
\noindent \textbf{Under-approximation analysis  approaches.} The essence of our dual approximation approach is to underestimate activation functions' domains for guiding the definition of tight over-approximations. There are several related under-approximation approaches based on either white-box attacks \cite{chakraborty2018adversarial} or testings \cite{he2020towards}. For instance, 
the fast gradient sign method (FGSM) \cite{goodfellow2014explaining} is a well-known approach for generating adversarial examples to intrigue corner cases for classifications. Other attack approaches include C\&W \cite{carlini2017towards}, DeepFool \cite{moosavi2016deepfool}, and JSMA \cite{papernot2016limitations}. 
The white-box testing for neural networks is to generate specific test cases to intrigue target neurons under different coverage criteria. Various test case generation and selection approaches have been proposed \cite{lee2020effective,sun2019structural,yu2022white,guo2018dlfuzz,GaoFYL0X22}. We believe that our approach provides a flexible hybrid mechanism to combine these attack- and testing-based under-approximation approaches into over-approximation-based verification approaches for neural network robustness verification. 
These sophisticated attack and testing approaches can be used for underestimating the approximation domains. We would consider integrating them into our dual-approximation approach for the improvement of both tightness and efficiency.

%% file: 7conclusion.tex
\vspace{-2mm}
\section{Conclusion}

\label{sec:conclusion}

We have proposed a dual-approximation approach to define tight over-approximations for the robustness verification of DNNs. 
Underlying this approach is our finding of \emph{approximation domain} of the activation function that is crucial in defining tight over-approximations, yet overlooked by all existing approximation approaches. 
%
Accordingly,
we have devised two complementary under-approximation algorithms to compute underestimated domains, which are proved to be useful to define tight over-approximations for neural networks. 
Based on this, we proposed a novel dual-approximation approach to define tight over-approximations via the additional underestimated domain of activation functions. 
Our experimental results have
demonstrated the outperformance of our approach over the state of the art.

Our dual-approximation approach
 sheds light on a new direction at  
 integrating under-approximation approaches such as attacks and testings into over-approximation-based verification approaches for neural networks. 
Besides, it could also be integrated into other abstraction-based neural network verification approaches~\cite{gehr2018ai2,singh2019abstract,zhang2020detecting} as they require non-linear activation functions that shall be over-approximated to handle abstract domains. 
In addition to the robustness verification, we believe that our approach is also applicable to the variants of robustness verification problems,  such as fairness~\cite{bastani2019probabilistic} and $\epsilon$-weekend robustness~\cite{huang2022}. Verifying those properties can be reduced to optimization problems  that contain the nonlinear activation functions in networks. 

\vspace{-2mm}

\section*{Acknowledgments}
 The authors thank all the anonymous reviewers for their valuable feedback. This work is supported by National Key Research Program (2020AAA0107800), NSFC-ISF Joint Program (62161146001, 3420/21), Huawei, Shanghai Trusted Industry Internet Software Collaborative Innovation Center, and the ``Digital Silk Road'' Shanghai International Joint Lab of Trustworthy Intelligent Software (22510750100).
Min Zhang is the corresponding author.

%% file: 8appendix.tex
\section{The Proofs of Theorems}

\label{sec:add_proof}

In this section, we give the proof of Theorem \ref{thm:tight_approx} and Theorem \ref{thm:precise_domain} proposed in Section \ref{two_theorem}. 

\subsection{Proof of Theorem \ref{thm:tight_approx}}

\begin{theorem}[3.1]
Suppose that there are two over-approximations $h_L(z^{(j)}(x)),h_U(z^{(j)}(x))$ and ${h'}_L(z^{(j)}(x)),{h'}_U(z^{(j)}(x))$ for each $z^{(j)}(x)$ in Definition \ref{def:approximation_domain} and $h_L(z^{(j)}(x)),h_U(z^{(j)}(x))$ are tighter than ${h'}_L(z^{(j)}(x)),{h'}_U(z^{(j)}(x))$, respectively. 
The approximation domain 
$[l_r^{(i)}, u_r^{(i)}]$ computed by 
$h_L(z^{(j)}(x)),h_U(z^{(j)}(x))$ must be more precise than the one 
$[{l'}_r^{(i)}, {u'}_r^{(i)}]$  by  ${h'}_L(z^{(j)}(x)),{h'}_U(z^{(j)}(x))$. 
\end{theorem}

\begin{proof}
    Assume that there is a neuron $N_i$ with m neurons $N_1$, $N_2$, $\cdots$, $N_m$ connected to it on the last layer. There are two over-approximation methods, one of which is tighter than the other. For $N_i$, the tighter approximated output of it is $[l_i, u_i]$, and the looser is $[l'_i, u'_i]$ where $l'_i \leq l_i$ and $u_i \leq u'_i$. For $N_1$, $N_2$, $\cdots$, $N_m$, they are $[l_1, u_1]$, $[l_2, u_2]$, $\cdots$, $[l_m, u_m]$ and $[l'_1, u'_1]$, $[l'_2, u'_2]$, $\cdots$, $[l'_m, u'_m]$, respectively. Let $w_1$, $w_2$, $\cdots$, $w_m$ and $b_1$, $b_2$, $\cdots$, $b_m$ be the weights of biases of the connection between $N_i$ and $N_1$, $N_2$, $\cdots$, $N_m$. We consider three cases where the weights are positive or negative: 
    
    \textbf{Case I: } All weights are positive. If $w_1, w_2, \cdots, w_m > 0$, then the approximation domains of $N_i$ are $[l_i, u_i]$ and $[l'_i, u'_i]$ where
    \begin{align}
        l_i = & \enspace w_1  l_1 + w_2  l_2 + \cdots + w_m  l_m + b_1 + b_2 + \cdots + b_m \notag \\
        u_i = & \enspace w_1  u_1 + w_2  u_2 + \cdots + w_m  u_m + b_1 + b_2 + \cdots + b_m \notag \\
        l'_i = & \enspace w_1  l'_1 + w_2  l'_2 + \cdots + w_m  l'_m + b_1 + b_2 + \cdots + b_m \notag \\
        u'_i = & \enspace w_1  u'_1 + w_2  u'_2 + \cdots + w_m  u'_m + b_1 + b_2 + \cdots + b_m \notag
    \end{align}
    Clearly, $l_i \geq l'_i$ and $u_i \leq u'_i$. We know that $[l_i, u_i]$ is tighter than $[l'_i, u'_i]$.
    
    \textbf{Case II: } $m-1$ weights are positive. If $w_1, w_2, \cdots, w_{m-1} > 0$ and $w_m \leq 0$, then:
    \begin{align}
        l_i = & \enspace w_1  l_1 + w_2  l_2 + \cdots + w_{m-1}  l_{m-1} + w_m  u_m + \notag \\ & \enspace b_1 + b_2 + \cdots + b_m \notag \\
        u_i = & \enspace w_1  u_1 + w_2  u_2 + \cdots + w_{m-1}  u_{m-1} + w_m  l_m + \notag \\ & \enspace b_1 + b_2 + \cdots + b_m \notag \\
        l'_i = & \enspace w_1  l'_1 + w_2  l'_2 + \cdots + w_{m-1}  l'_{m-1} + w_m  u'_m + \notag \\ & \enspace b_1 + b_2 + \cdots + b_m \notag \\
        u'_i = & \enspace w_1  u'_1 + w_2  u'_2 + \cdots + w_{m-1}  u'_{m-1} + w_m  l'_m + \notag \\ & \enspace b_1 + b_2 + \cdots + b_m \notag
    \end{align}
    Since $l_1 \geq l'_1$, $l_2 \geq l'_2$, $\cdots$, $l_{m-1} \leq l'_{m-1}$ and $u_m \leq u'_m$, we have $l_i \geq l'_i$. Similarly, $u_i \leq u'_i$, and then $[l_i, u_i$ is tighter than $[l'_i, u'_i]$.
    
    \textbf{Case III:} No weights are positive. If $w_1, w_2, \cdots. w_m <= 0$, then:
    \begin{align}
        l_i = & \enspace w_1  u_1 + w_2  u_2 + \cdots + w_m  u_m + b_1 + b_2 + \cdots + b_m \notag \\
        u_i = & \enspace w_1  l_1 + w_2  l_2 + \cdots + w_m  l_m + b_1 + b_2 + \cdots + b_m \notag \\
        l'_i = & \enspace w_1  u'_1 + w_2  u'_2 + \cdots + w_m  u'_m + b_1 + b_2 + \cdots + b_m \notag \\
        u'_i = & \enspace w_1  l'_1 + w_2  l'_2 + \cdots + w_m  l'_m + b_1 + b_2 + \cdots + b_m \notag
    \end{align}
    Clearly, $l_i \geq l'_i$ and $u_i \leq u'_i$, and then $[l_i, u_i$ is tighter than $[l'_i, u'_i]$.

    From the three cases above, we can conclude that whether the weights $w$ of connection between two neurons are positive or non-positive, and regardless of the biases $b$, if the over-approximation is tighter, the approximation domain of neurons in the next layer will be more precise.
\end{proof}

\subsection{Proof of Theorem \ref{thm:precise_domain}}

\begin{theorem}[3.2]
	Given two approximation domains $[l_r^{(i)},u_r^{(i)}]$ and $[{l'}_r^{(i)},{u'}_r^{(i)}]$ such that  ${l'}_r^{(i)} < l_r^{(i)}$ and $u_r^{(i)} < {u'}_r^{(i)}]$, for any over-approximation $({h'}_L(z^{(j)}(x)), {h'}_U(z^{(j)}(x)))$ of continuous function $\sigma(x)$ on $[{l'}_r^{(i)},{u'}_r^{(i)}]$, there exists an over-approximation $(h_L(z^{(j)}(x))$, $h_U(z^{(j)}(x)))$ on $[l_r^{(i)},u_r^{(i)}]$ such that $\forall z^{(j)}(x) \in [l_r^{(i)},u_r^{(i)}], h'_L(x) \le h_L(z^{(j)}(x)), h'_U(x) \ge h_U(z^{(j)}(x))$.
\end{theorem}

\begin{figure}
	\centering
	\begin{subfigure}{0.49\linewidth}
		\centering
		\includegraphics[width=0.95\linewidth]{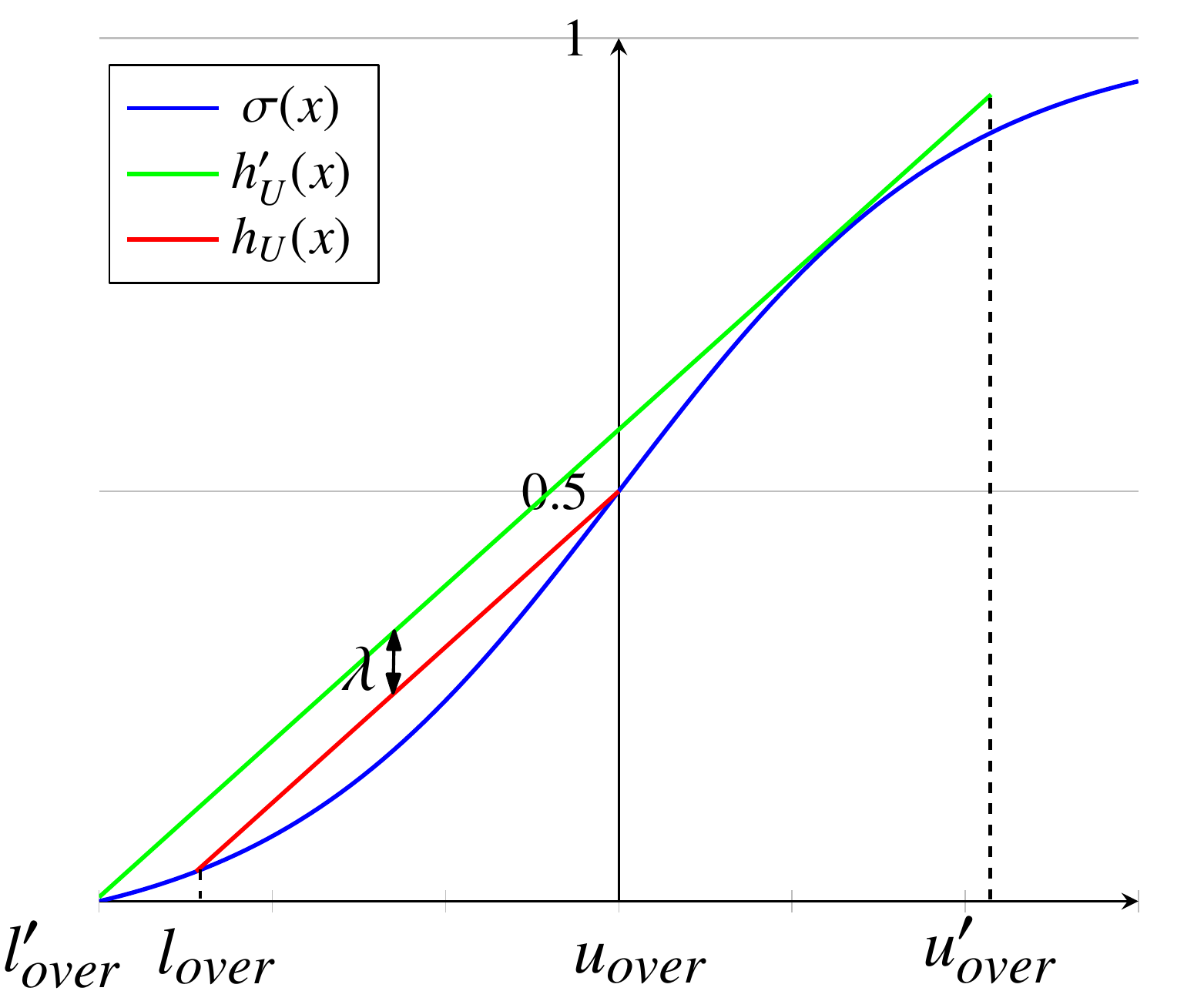}
		\caption{$h_U(x) = h'_U(x) - \lambda$}
	\end{subfigure}
	\begin{subfigure}{0.49\linewidth}
		\centering
		\includegraphics[width=0.95\linewidth]{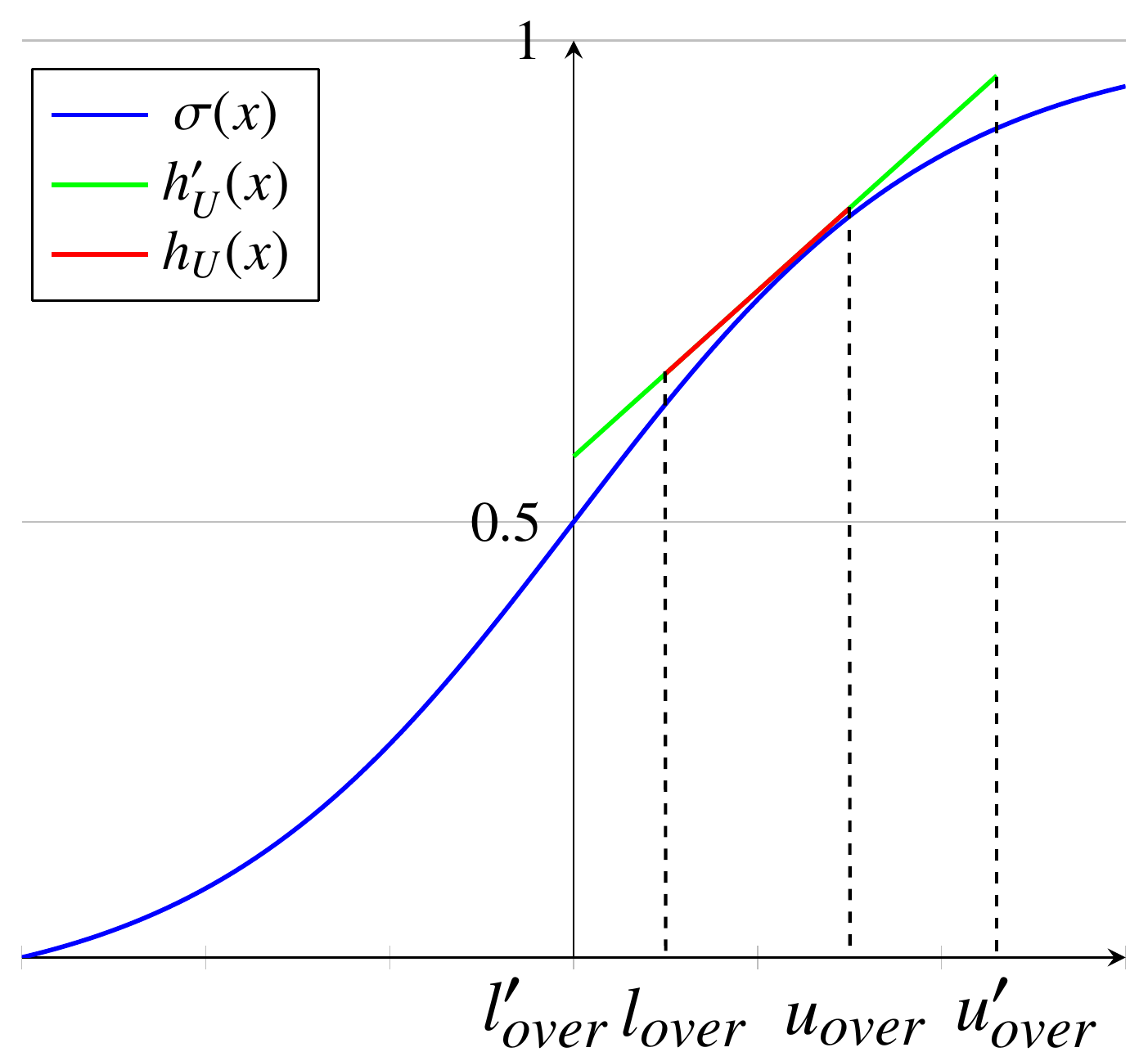}
		\caption{$h_U(x) = h'_U(x)$}
	\end{subfigure}
	\caption{Two cases of defining $h_U(x)$ based on $h'_U(x)$}
	\label{fig:proof1}
\end{figure}

\begin{proof}
    To simplify, We omit superscript and subscript and use $[l_{over}, u_{over}]$ and $[l'_{over}, u'_{over}]$ to denote the overestimated approximation domains.
    Since we have $l'_{over} < l_{over}$ and $u_{over} < u'_{over}$, which means $[l_{over}, u_{over}] \subset [l'_{\mathit{over}},u'_{\mathit{over}}]$, and $\sigma$ is a continuous function, we can conclude that the range of $\sigma$ satisfies: $\sigma([l_{\mathit{over}},u_{\mathit{over}}]) \subset \sigma([l'_{\mathit{over}},u'_{\mathit{over}}])$. Given the linear upper bound $h'_U(x)$ on $[l'_{over}, u'_{over}]$, we can define a linear upper bound $h_U(x)$ on $[l_{over}, u_{over}]$ as:
    \begin{align}
        h_U(x) = \begin{cases}
            h'_U(x) - \lambda & \mathit{if}~\forall x \in [l_{over}, u_{over}], \exists \lambda > 0, \\ 
            & \quad h'_U(x) - \sigma(x) \geq \lambda \\
            h'_U(x) & \mathit{otherwise}
            \end{cases}
    \end{align}
    The two cases above are intuitively shown in Figure \ref{fig:proof1}. Apparently, $h_U(x)$ is tighter than $h'_U(x)$ since $h_U(x) \leq h'_U(x)$.
    
    Now we show that $h_U(x)$ is a sound upper bound for $\sigma$ on $[l_{over}, u_{over}]$. That is, for all $x$ in $[l_{over}, u_{over}]$, if $h_U(x) = h'_U(x) - \lambda$, then $h_U(x) - \sigma(x) = h'_U(x) - \lambda - \sigma(x)$. Since $h'_U(x) - \sigma(x) \geq \lambda$, we have $h_U(x) - \sigma(x) \geq 0$, and in this case $h_U(x)$ is sound. Moreover, if $h_U(x) = h'_U(x)$, since $h'_U(x)$ is a linear upper bound, $h'_U(x) \geq \sigma(x)$, and $h_U(x) \geq \sigma(x)$. 
    For the linear lower bound, the proof is similar. We can conclude that more precise approximation domains lead to tighter over-approximations of activation functions. 
\end{proof}

\section{Soundness of Our Approach}
\label{appsec:sound}
We demonstrate the soundness of our dual-approximation approach. For non-linear functions $\sigma(x)$, we first underestimate its actual domain using Monte Carlo or gradient-based approaches, and then over-approximate it on both the underestimated domain and the overestimated domain. The over-approximation strategy computes $h_L(x)$ and $h_U(x)$ for each case. We prove that our approach is sound by showing $\forall x \in [l_{over}, u_{over}], h_L(x) \leq \sigma(x) \leq h_U(x)$. 

\begin{theorem}
    The dual-approximation approach is sound. 
\end{theorem}

\begin{proof}
    Consider the three cases defined in Section \ref{sec:over-approximation-strategy}. In case \uppercase\expandafter{\romannumeral1}, the linear upper bound $h_U(x) = k(x - u_{over}) + \sigma(u_{over})$. 
    Let $f(x) = h_U(x) - \sigma(x)$, we show $h_U(x)$ is sound by proving $f(x) \geq 0$. The derivative of $f(x)$ is $f'(x) = k - \sigma'(x)$. 
    
    If $l_{over}, u_{over} < 0$, as $\sigma'(l_{over}) < k < \sigma'(u_{over})$ and $\sigma(x)$ is monotonically increasing, there must exists $x_0 \in (l_{over}, u_{over})$ where $\sigma'(x_0) = k$. When $x \in [l_{over}, x_0]$, $f'(x) > 0$, and $f(x)$ is monotonically increasing; when $x \in [x_0, u_{over}]$, $f'(x) < 0$, and $f(x)$ is monotonically decreasing. The two minimum occurs at $l_{over}$ and $u_{over}$. Since $f(l_{over}) = f(u_{over}) = 0$, we can conclude that $f(x) \geq 0$.

    If $l_{over} < 0$ and $u_{over} > 0$, we divide $[l_{over}, u_{over}]$ into $[l_{over}, 0]$ and $[0, u_{over}]$, and consider them separately. 
    For $x \in [l_{over}, 0]$, the case is similar to the above. There exists some $x_0$ where $f'(x_0) = 0$. $f(x)$ is monotonically increasing in $[l_{over}, x_0]$, and monotonically decreasing in $[x_0, 0]$. 
    For $x \in [0, u_{over}]$, as $k < \sigma'(u_{over})$ and $\sigma(x)$ is monotonically decreasing, we know that $f'(x) < 0$, and $f(x)$ is monotonically decreasing in $[0, u_{over}]$. All in all, $f(x)$ is monotonically increasing in $[l_{over}, x_0]$, and monotonically decreasing in $[x_0, u_{over}]$.  The two minimum occurs at $l_{over}$ and $u_{over}$. Since $f(l_{over}) = f(u_{over}) = 0$, we know that $f(x) \geq 0$.

    The condition where $l_{over}, u_{over} > 0$ is nonexistent in this case.
    
    In the proof above, we show that $\forall x \in [l_{over},u_{over}]$, the upper bound $h_U(x) \geq \sigma(x)$, and thus it is sound.
    As for the lower bound $h_L(x)$, as well as Case \uppercase\expandafter{\romannumeral2} and \uppercase\expandafter{\romannumeral3}, the proof is similar.
\end{proof}


\section{Additional Experimental Results}

\begin{table*}[h!]
\centering
\caption{Additional Comparison result I: Comparison of Monte-Carlo-version DualApp (DA) and existing tools, NeWise (NW), DeepCert (DC), VeriNet (VN) and RobustVerifier (RV) on 16 Sigmoid networks.}
\setlength{\tabcolsep}{3.8pt}
\begin{tabular}{|l|r|r|r|r|r|r|r|r|r|r|r|r|r|}
\hline
\multirow{2}{*}{Database}                                                  & \multirow{2}{*}{Model}                                       & \multirow{2}{*}{Nodes} & DA & NW  & \multirow{2}{*}{\begin{tabular}[c]{@{}l@{}}Impr.\\ (\%)\end{tabular}} & DC & \multirow{2}{*}{\begin{tabular}[c]{@{}l@{}}Impr.\\ (\%)\end{tabular}} & VN & \multirow{2}{*}{\begin{tabular}[c]{@{}l@{}}Impr.\\ (\%)\end{tabular}} & \begin{tabular}[c]{@{}l@{}}RV\end{tabular} & \multirow{2}{*}{\begin{tabular}[c]{@{}l@{}}Impr.\\ (\%)\end{tabular}} & \multirow{2}{*}{\begin{tabular}[c]{@{}l@{}}DA \\ Time (s)\end{tabular}} & \multirow{2}{*}{Others Time(s)}   \\ \cline{4-5} \cline{7-7} \cline{9-9} \cline{11-11}
                                                                           &                                                              &                        & Bounds   & Bounds  &                                                                        & Bounds   &                                                                        & Bounds  &                                                                        & Bounds                                                    &                                                                        &                                                                               &                                   \\ \hline
\multirow{5}{*}{Mnist}                                                     & \begin{tabular}[c]{@{}l@{}}${\rm CNN}_{3-2}$\end{tabular}  & 2,514                  & 0.06119   & 0.06074 & 0.74                                                                & 0.05789 & 5.70                                                                  &  0.05803    & 5.45                                                                   & 0.05686                                                   & 7.62                                                                   & 2.81                                                                          & 0.15   \textpm 0.02    \\  
                                                                           & ${\rm CNN}_{3-4}$          & 5018   & 0.04875 & 0.04776 & 2.07  & 0.04721 & 3.26 & 0.04715 & 3.39 & 0.04639 & 5.09                                                                  & 3.44                                                                          & 0.15   \textpm 0.02    \\ 
                                                                           & ${\rm CNN}_{4-5}$          & 8690   & 0.04863 & 0.04763 & 2.10  & 0.04551 & 6.86 & 0.04548 & 6.93 & 0.04355 & 11.66                                                                  & 15.33                                                                          & 0.89   \textpm 0.03    \\ 
                                                                           & ${\rm FNN}_{3 \times 50}$  & 160    & 0.00779 & 0.00693 & 12.41 & 0.0076  & 2.50 & 0.00768 & 1.43 & 0.00649 & 20.03                                                                  & 3.28                                                                          & 0.15   \textpm 0.02    \\ 
                                                                           & ${\rm FNN}_{3 \times 100}$ & 310    & 0.00885 & 0.00777 & 13.90 & 0.00858 & 3.15 & 0.00871 & 1.61 & 0.00744 & 18.95                                                                  & 5.46                                                                         & 0.22   \textpm 0.02    \\   \hline
\multirow{7}{*}{\begin{tabular}[c]{@{}l@{}}Fashion \\ Mnist\end{tabular}} & \begin{tabular}[c]{@{}l@{}}${\rm CNN}_{3-2}$\end{tabular}  & 2,514                  & 0.03006  & 0.02839 & 5.88                                                                   & 0.02584  & 16.33                                                                  & 0.02930  & 2.59                                                                   & 0.02811                                                   & 6.94                                                                   & 2.86                                                                          & 0.15   \textpm 0.01    \\ 
                                                                           & \begin{tabular}[c]{@{}l@{}}${\rm CNN}_{3-4}$\end{tabular} & 5,018                  & 0.03430   & 0.03125 & 9.76                                                                   & 0.03084  & 11.22                                                                  & 0.03390  & 1.18                                                                   & 0.03094                                                   & 10.86                                                                  & 5.92                                                                          & 0.28   \textpm 0.03    \\  
                                                                           & ${\rm FNN}_{3\times 50}$                                                         & 160                    & 0.00574  & 0.00476 & 20.59                                                                  & 0.00447  & 28.41                                                                  & 0.00565 & 1.59                                                                   & 0.00476                                                   & 20.59                                                                  & 3.15                                                                         & 0.14   \textpm 0.01    \\ 
                                                                           & ${\rm FNN}_{3\times 100}$                                                        & 310                    & 0.00497  & 0.00391 & 27.11                                                                  & 0.00387  & 28.42                                                                  & 0.00490  & 1.43                                                                   & 0.00471                                                   & 5.52                                                                   & 5.32                                                                          & 0.42   \textpm 0.08    \\  
                                                                           & ${\rm FNN}_{3\times 400}$                                                        & 1210                   & 0.00442  & 0.00312 & 41.67                                                                  & 0.00322  & 37.27                                                                  & 0.00437 & 1.14                                                                   & 0.00355                                                   & 24.51                                                                  & 25.22                                                                         & 4.20   \textpm 0.21    \\ 
                                                                           & ${\rm FNN}_{3\times 700}$                                                        & 2,110                  & 0.00408  & 0.00284 & 43.66                                                                  & 0.00290   & 40.69                                                                  & 0.00401 & 1.75                                                                   & 0.00318                                                   & 28.30                                                                  & 50.12                                                                         & 12.14   \textpm 0.39   \\  \hline
\multirow{5}{*}{Cifar-10}                                                  & \begin{tabular}[c]{@{}l@{}}${\rm CNN}_{3-4}$\end{tabular}  & 5,018                  & 0.01136  & 0.01064 & 6.77                                                                   & 0.01067  & 6.47                                                                   & 0.01133 & 0.26                                                                   & 0.01118                                                   & 1.61                                                                   & 4.87                                                                          & 0.28   \textpm 0.03    \\  
                                                                           & ${\rm CNN}_{4-5}$          & 8690   & 0.01942 & 0.01863 & 4.24  & 0.01917 & 1.30 & 0.01912 & 1.57 & 0.01887 & 2.91                                                                  & 17.22                                                                          & 1.17   \textpm 0.08    \\ 
                                                                           & ${\rm FNN}_{3\times 50}$                                                         & 160    & 0.00455 & 0.00406 & 12.07 & 0.00454 & 0.22 & 0.00452 & 0.66 & 0.00415 & 9.64                                                                  & 6.68                                                                          & 2.07   \textpm 0.26    \\ 
                                                                           & ${\rm FNN}_{3\times 100}$                                                        & 310    & 0.00469 & 0.00401 & 16.96 & 0.00468 & 0.21 & 0.00466 & 0.64 & 0.00426 & 10.09                                                                  & 12.23                                                                         & 3.98\textpm   0.84     \\ 
                                                                           & ${\rm FNN}_{3\times 200}$                                                       & 610    & 0.00408 & 0.00344 & 18.60 & 0.00407 & 0.25 & 0.00406 & 0.49 & 0.00377 & 8.22                                                                  & 30.27                                                                         & 12.82   \textpm 2.11   \\   \hline
\end{tabular}
\label{tab:appendix_sigmoid}
\end{table*}

\begin{table*}[h!]
\centering
\caption{Additional Comparison result II: Comparison results of Monte-Carlo-version DualApp (DA) and existing tools, NeWise (NW), DeepCert (DC), VeriNet (VN), and RobustVerifier (RV) on 16 Tanh networks.}
\setlength{\tabcolsep}{3.8pt}
\begin{tabular}{|l|r|r|r|r|r|r|r|r|r|r|r|r|r|}
\hline
\multirow{2}{*}{Database}                                                 & \multirow{2}{*}{Model}                                       & \multirow{2}{*}{Nodes} & DA & NW  & \multirow{2}{*}{\begin{tabular}[c]{@{}l@{}}Impr.\\ (\%)\end{tabular}} & DC & \multirow{2}{*}{\begin{tabular}[c]{@{}l@{}}Impr.\\ (\%)\end{tabular}} & VN & \multirow{2}{*}{\begin{tabular}[c]{@{}l@{}}Impr.\\ (\%)\end{tabular}} & \begin{tabular}[c]{@{}l@{}}RV\end{tabular} & \multirow{2}{*}{\begin{tabular}[c]{@{}l@{}}Impr.\\ (\%)\end{tabular}} & \multirow{2}{*}{\begin{tabular}[c]{@{}l@{}}DA \\ Time (s)\end{tabular}} & \multirow{2}{*}{Others Time(s)}   \\ \cline{4-5} \cline{7-7} \cline{9-9} \cline{11-11}
                                                                          &                                                              &                        & Bounds   & Bounds  &                                                                        & Bounds   &                                                                        & Bounds  &                                                                        & Bounds                                                    &                                                                        &                                                                               &                                   \\ \hline
\multirow{6}{*}{Mnist}                     \setlength{\tabcolsep}{2pt}                               & \begin{tabular}[c]{@{}l@{}}${\rm CNN}_{3-2}$\end{tabular} & 2,514                  & 0.02677  & 0.02558 & 4.65                                                                   & 0.02618  & 2.25                                                                   & 0.02622 & 2.10                                                                   & 0.02565                                                   & 4.37                                                                   & 3.34                                                                          & 0.16   \textpm 0.02    \\ 
                                                                          &   ${\rm FNN}_{3\times 50}$                                                        & 160                    & 0.00545  & 0.00456 & 19.52                                                                  & 0.00529  & 3.02                                                                   & 0.00536 & 1.68                                                                   & 0.00439                                                   & 24.15                                                                  & 3.02                                                                          & 0.14   \textpm 0.00    \\ 
                                                                          & ${\rm FNN}_{3\times 100}$                                                       & 310                    & 0.00623  & 0.00499 & 24.85                                                                  & 0.00609  & 2.30                                                                   & 0.00616 & 1.14                                                                   & 0.00512                                                   & 21.68                                                                  & 5.28                                                                          & 0.37   \textpm 0.01    \\ 
                                                                          & ${\rm FNN}_{3\times 200}$                                                        & 610                    & 0.00676  & 0.00508 & 33.07                                                                  & 0.00663  & 1.96                                                                   & 0.00669 & 1.05                                                                   & 0.00557                                                   & 21.36                                                                  & 10.42                                                                          & 1.19   \textpm 0.05    \\ 
                                                                          & ${\rm FNN}_{3\times 400}$                                                        & 1,210                  & 0.00672  & 0.00479 & 40.29                                                                  & 0.00660   & 1.82                                                                   & 0.00667 & 0.75                                                                   & 0.00553                                                   & 21.52                                                                  & 24.28                                                                         & 4.63   \textpm 0.12    \\ 
                                                                          & ${\rm FNN}_{3\times 700}$                                                        & 2,110                  & 0.00665  & 0.00459 & 44.88                                                                  & 0.00650   & 2.31                                                                   & 0.00660  & 0.76                                                                   & 0.00537                                                   & 23.84                                                                  & 50.48                                                                         & 12.27   \textpm 0.43   \\  \hline
\multirow{7}{*}{\begin{tabular}[c]{@{}l@{}}Fashion \\ Mnist\end{tabular}} & \begin{tabular}[c]{@{}l@{}}${\rm CNN}_{3-2}$\end{tabular} & 2514   & 0.09247 & 0.09091 & 1.72  & 0.08805 & 5.02 & 0.08772 & 5.41 & 0.08387 & 10.25                                                                   & 3.22                                                                          & 0.15   \textpm 0.02    \\ 
                                                                          & \begin{tabular}[c]{@{}l@{}}${\rm CNN}_{3-4}$\end{tabular} & 5018   & 0.07704 & 0.07452 & 3.38  & 0.0729  & 5.68 & 0.07295 & 5.61 & 0.06848 & 12.50                                                                   & 5.68                                                                          & 0.26   \textpm 0.05    \\  
                                                                          & ${\rm FNN}_{3\times 50}$                                                         & 160    & 0.01035 & 0.00915 & 13.11 & 0.01013 & 2.17 & 0.0102  & 1.47 & 0.00858 & 20.63                                                                  & 3.87                                                                          & 0.20   \textpm 0.01    \\ 
                                                                          & ${\rm FNN}_{3\times 100}$                                                        & 310    & 0.00921 & 0.00797 & 15.56 & 0.00907 & 1.54 & 0.00911 & 1.10 & 0.00778 & 18.38                                                                  & 6.33                                                                          & 0.57   \textpm 0.12    \\  
                                                                          & ${\rm FNN}_{3\times 700}$                                                        & 2,110  & 0.00757 & 0.00634 & 19.40 & 0.00749 & 1.07 & 0.00751 & 0.80 & 0.00666 & 13.66                                                                  & 78.02                                                                         & 13.44   \textpm 0.62   \\  \hline
\multirow{4}{*}{Cifar-10}                                                 & \begin{tabular}[c]{@{}l@{}}${\rm CNN}_{3-4}$\end{tabular} & 5018   & 0.02551 & 0.02469 & 3.32  & 0.02528 & 0.91 & 0.02523 & 1.11 & 0.02492 & 2.37                                                                   & 6.02                                                                          & 0.26   \textpm 0.05    \\ 
                                                                           
                                                                          & ${\rm CNN}_{4-5}$          & 8690   & 0.01942 & 0.01863 & 4.24  & 0.01917 & 1.30 & 0.01912 & 1.57 & 0.01887 & 2.91                                                                  & 7.79                                                                          & 1.73   \textpm 0.24    \\  
                                                                           
                                                                          & ${\rm FNN}_{3\times 50}$                                                         & 160                    & 0.00287  & 0.00232 & 23.71                                                                  & 0.00285  & 0.70                                                                   & 0.00285 & 0.70                                                                   & 0.00254                                                   & 12.99                                                                  & 7.79                                                                          & 1.73   \textpm 0.24    \\ 
                                                                          & ${\rm FNN}_{3\times 100}$                                                        & 310                    & 0.00253  & 0.00194 & 30.41                                                                  & 0.00251  & 0.80                                                                   & 0.00251 & 0.80                                                                   & 0.00225                                                   & 12.44                                                                  & 19.47                                                                          & 10.45   \textpm 4.36   \\ 
                                                                          & ${\rm FNN}_{3\times 700}$                                                        & 2,110                  & 0.00229  & 0.00155 & 47.74                                                                  & 0.00226  & 1.33                                                                   & 0.00228 & 0.44                                                                   & 0.00201                                                   & 13.93                                                                  & 122.89                                                                         & 66.01   \textpm 12.74 \\  \hline
\end{tabular}
\label{tab:appendix_tanh}
\end{table*}

\begin{table*}[h!]
\centering
\caption{Additional Comparison results III: Comparison of Monte-Carlo-version DualApp (DA) and existing tools, NeWise (NW), DeepCert (DC), VeriNet (VN) and RobustVerifier (RV) on 18 Arctan networks.}
\setlength{\tabcolsep}{4.2pt}
\begin{tabular}{|l|r|r|r|r|r|r|r|r|r|r|r|r|r|}
\hline
\multirow{2}{*}{Database}                                                  & \multirow{2}{*}{Model}                                       & \multirow{2}{*}{Nodes} & DA & NW  & \multirow{2}{*}{\begin{tabular}[c]{@{}l@{}}Impr.\\ (\%)\end{tabular}} & DC & \multirow{2}{*}{\begin{tabular}[c]{@{}l@{}}Impr.\\ (\%)\end{tabular}} & VN & \multirow{2}{*}{\begin{tabular}[c]{@{}l@{}}Impr.\\ (\%)\end{tabular}} & \begin{tabular}[c]{@{}l@{}}RV\end{tabular} & \multirow{2}{*}{\begin{tabular}[c]{@{}l@{}}Impr.\\ (\%)\end{tabular}} & \multirow{2}{*}{\begin{tabular}[c]{@{}l@{}}DA\end{tabular}} & \multirow{2}{*}{Others Time(s)}   \\ \cline{4-5} \cline{7-7} \cline{9-9} \cline{11-11}
                                                                           &                                                              &                        & Bounds   & Bounds  &                                                                        & Bounds   &                                                                        & Bounds  &                                                                        & Bounds                                                    &                                                                        &                                                                               &                                   \\ \hline
\multirow{6}{*}{Mnist}                                                     & \begin{tabular}[c]{@{}l@{}}${\rm CNN}_{3-2}$\end{tabular}  & 2,514                  & 0.01920   & 0.01821 & 5.44                                                                   & 0.01836  & 4.58                                                                   & 0.01896 & 1.27                                                                   & 0.01829                                                   & 4.98                                                                   & 2.78                                                                          & 0.15   \textpm 0.02    \\  
                                                                           & ${\rm FNN}_{3\times 50}$                                                         & 160                    & 0.00568  & 0.00481 & 18.09                                                                  & 0.00437  & 29.98                                                                  & 0.00558 & 1.79                                                                   & 0.00465                                                   & 22.15                                                                  & 3.02                                                                          & 0.16   \textpm 0.02    \\ 
                                                                           & ${\rm FNN}_{3\times 100}$                                                        & 310                    & 0.00635  & 0.00518 & 22.59                                                                  & 0.00495  & 28.28                                                                  & 0.00627 & 1.28                                                                   & 0.00529                                                   & 20.04                                                                  & 5.11                                                                          & 0.37   \textpm 0.04    \\ 
                                                                           & ${\rm FNN}_{3\times 200}$                                                        & 610                    & 0.00716  & 0.00547 & 30.90                                                                  & 0.00554  & 29.24                                                                  & 0.00708 & 1.13                                                                   & 0.00587                                                   & 21.98                                                                  & 10.34                                                                          & 1.20   \textpm 0.02    \\ 
                                                                           & ${\rm FNN}_{3\times 400}$                                                        & 1,210                  & 0.00733  & 0.00528 & 38.83                                                                  & 0.00551  & 33.03                                                                  & 0.00725 & 1.10                                                                   & 0.00589                                                   & 24.45                                                                  & 25.46                                                                         & 4.65   \textpm 0.42    \\ 
                                                                           & ${\rm FNN}_{3\times 700}$                                                        & 2,110                  & 0.00721  & 0.00506 & 42.49                                                                  & 0.00532  & 35.53                                                                  & 0.00713 & 1.12                                                                   & 0.00572                                                   & 26.05                                                                  & 50.98                                                                         & 12.18   \textpm 0.23   \\  \hline
\multirow{7}{*}{\begin{tabular}[c]{@{}l@{}}Fashion \\ Mnist\end{tabular}} & \begin{tabular}[c]{@{}l@{}}${\rm CNN}_{3-2}$\end{tabular}  & 2,514                  & 0.03006  & 0.02839 & 5.88                                                                   & 0.02584  & 16.33                                                                  & 0.02930  & 2.59                                                                   & 0.02811                                                   & 6.94                                                                   & 2.86                                                                          & 0.15   \textpm 0.01    \\ 
                                                                           & \begin{tabular}[c]{@{}l@{}}${\rm CNN}_{3-4}$\end{tabular} & 5,018                  & 0.03430   & 0.03125 & 9.76                                                                   & 0.03084  & 11.22                                                                  & 0.03390  & 1.18                                                                   & 0.03094                                                   & 10.86                                                                  & 5.92                                                                          & 0.28   \textpm 0.03    \\  
                                                                           & ${\rm FNN}_{3\times 50}$                                                         & 160                    & 0.00574  & 0.00476 & 20.59                                                                  & 0.00447  & 28.41                                                                  & 0.00565 & 1.59                                                                   & 0.00476                                                   & 20.59                                                                  & 3.15                                                                         & 0.14   \textpm 0.01    \\ 
                                                                           & ${\rm FNN}_{3\times 100}$                                                        & 310                    & 0.00497  & 0.00391 & 27.11                                                                  & 0.00387  & 28.42                                                                  & 0.00490  & 1.43                                                                   & 0.00471                                                   & 5.52                                                                   & 5.32                                                                          & 0.42   \textpm 0.08    \\ 
                                                                           & ${\rm FNN}_{3\times 200}$                                                        & 610                    & 0.00471  & 0.00349 & 34.96                                                                  & 0.00358  & 31.56                                                                  & 0.00466 & 1.07                                                                   & 0.00390                                                    & 20.77                                                                  & 10.56                                                                          & 1.20   \textpm 0.06    \\ 
                                                                           & ${\rm FNN}_{3\times 400}$                                                        & 1210                   & 0.00442  & 0.00312 & 41.67                                                                  & 0.00322  & 37.27                                                                  & 0.00437 & 1.14                                                                   & 0.00355                                                   & 24.51                                                                  & 25.22                                                                         & 4.20   \textpm 0.21    \\ 
                                                                           & ${\rm FNN}_{3\times 700}$                                                        & 2,110                  & 0.00408  & 0.00284 & 43.66                                                                  & 0.00290   & 40.69                                                                  & 0.00401 & 1.75                                                                   & 0.00318                                                   & 28.30                                                                  & 50.12                                                                         & 12.14   \textpm 0.39   \\  \hline
\multirow{5}{*}{Cifar-10}                                                  & \begin{tabular}[c]{@{}l@{}}${\rm CNN}_{3-4}$\end{tabular}  & 5,018                  & 0.01136  & 0.01064 & 6.77                                                                   & 0.01067  & 6.47                                                                   & 0.01133 & 0.26                                                                   & 0.01118                                                   & 1.61                                                                   & 4.87                                                                          & 0.28   \textpm 0.03    \\  
                                                                           & ${\rm FNN}_{3\times 50}$                                                         & 160                    & 0.00323  & 0.00262 & 23.28                                                                  & 0.00266  & 21.43                                                                  & 0.00321 & 0.62                                                                   & 0.00283                                                   & 14.13                                                                  & 7.26                                                                          & 1.67   \textpm 0.16    \\ 
                                                                           & ${\rm FNN}_{3\times 100}$                                                        & 310                    & 0.00277  & 0.00216 & 28.24                                                                  & 0.00220   & 25.91                                                                  & 0.00275 & 0.73                                                                   & 0.00248                                                   & 11.69                                                                  & 13.88                                                                         & 4.33\textpm   0.88     \\ 
                                                                           & ${\rm FNN}_{3\times 200}$                                                       & 610                    & 0.00255  & 0.00189 & 34.92                                                                  & 0.00196  & 30.10                                                                  & 0.00253 & 0.79                                                                   & 0.00228                                                   & 11.84                                                                  & 27.94                                                                         & 10.48   \textpm 1.56   \\ 
                                                                           & ${\rm FNN}_{3\times 700}$                                                        & 2,110                  & 0.00246  & 0.00167 & 47.31                                                                  & 0.00179  & 37.43                                                                  & 0.00245 & 0.41                                                                   & 0.00212                                                   & 16.04                                                                  & 79.10                                                                         & 36.44   \textpm 3.85   \\  \hline
\end{tabular}
\label{tab:appendix_arctan}
\end{table*}

\subsection{Additional Results for Experiment \uppercase\expandafter{\romannumeral1}}

\label{sec:add_exp1}

This section presents the effectiveness comparison of DualApp with NeWise, DeepCert, VeriNet, and RobustVerifier on additional 16 Sigmoid, 16 Tanh, and 18 Arctan neural networks. The results are shown in Table \ref{tab:appendix_sigmoid}, \ref{tab:appendix_tanh}, and \ref{tab:appendix_arctan} respectively. All the experiments clearly show that our method outperforms our competitors on all datasets and models.

\begin{figure*}
	\centering
	\begin{subfigure}{0.32\textwidth}
		\includegraphics[width=\textwidth]{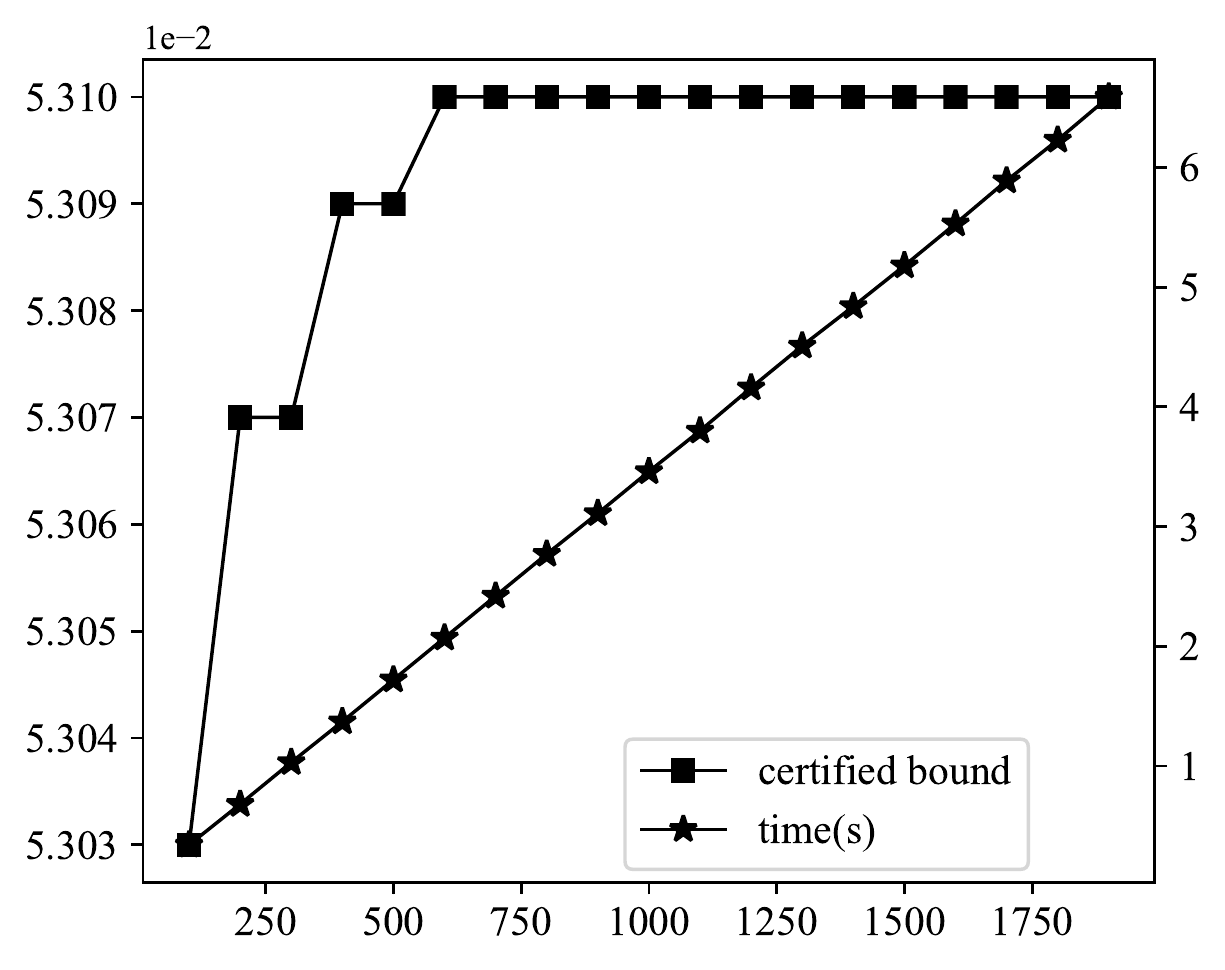}
		\caption{Sampling-based results on $\rm{CNN}_{2-3}$ trained on Mnist.}
		\label{fig:Sampling_mnist_cnn}
	\end{subfigure}
	\hfill
	\begin{subfigure}{0.32\textwidth}
		\includegraphics[width=\textwidth]{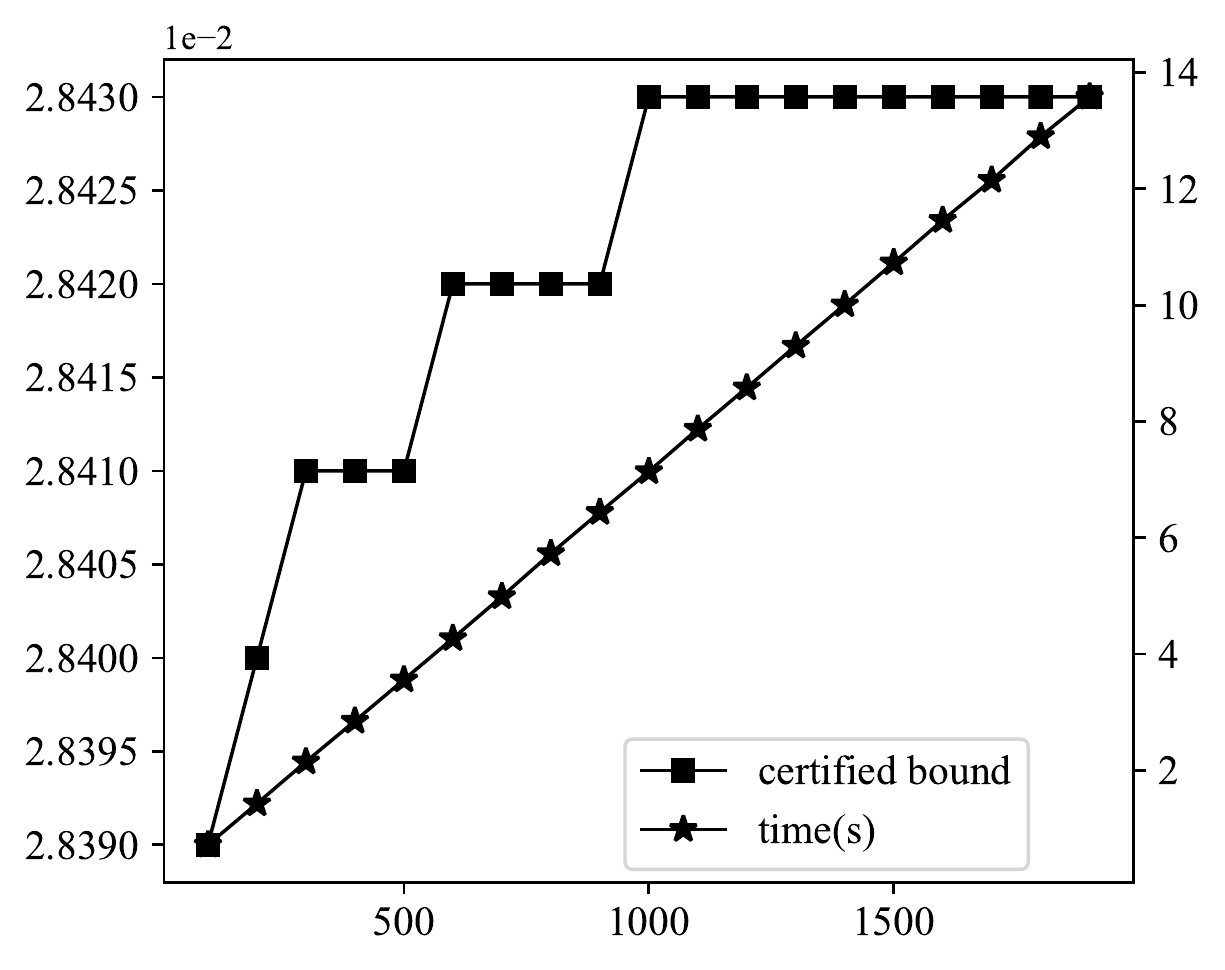}
		\caption{Sampling-based results on $\rm{FNN}_{1\times 200}$ trained on Mnist.}
		\label{fig:Sampling_mnist_fnn}
	\end{subfigure}
	\hfill
	\begin{subfigure}{0.32\textwidth}
		\includegraphics[width=\textwidth]{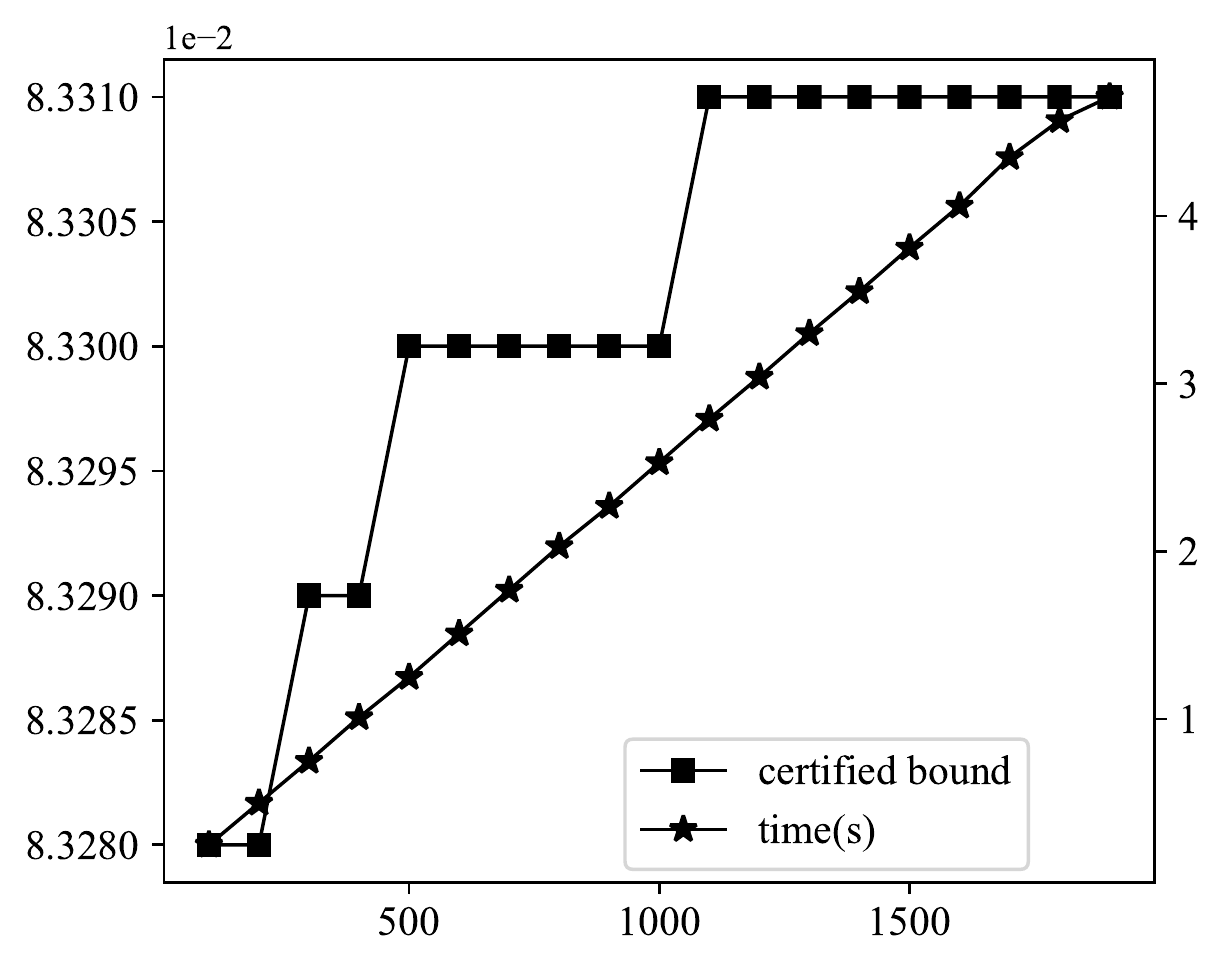}
		\caption{Sampling-based results on $\rm{CNN}_{2-2}$  trained on Fashion Mnist.}
		\label{fig:Sampling_fashion_mnist_cnn}
	\end{subfigure}
	\hfill\\
    
	\begin{subfigure}{0.32\textwidth}
		\includegraphics[width=\textwidth]{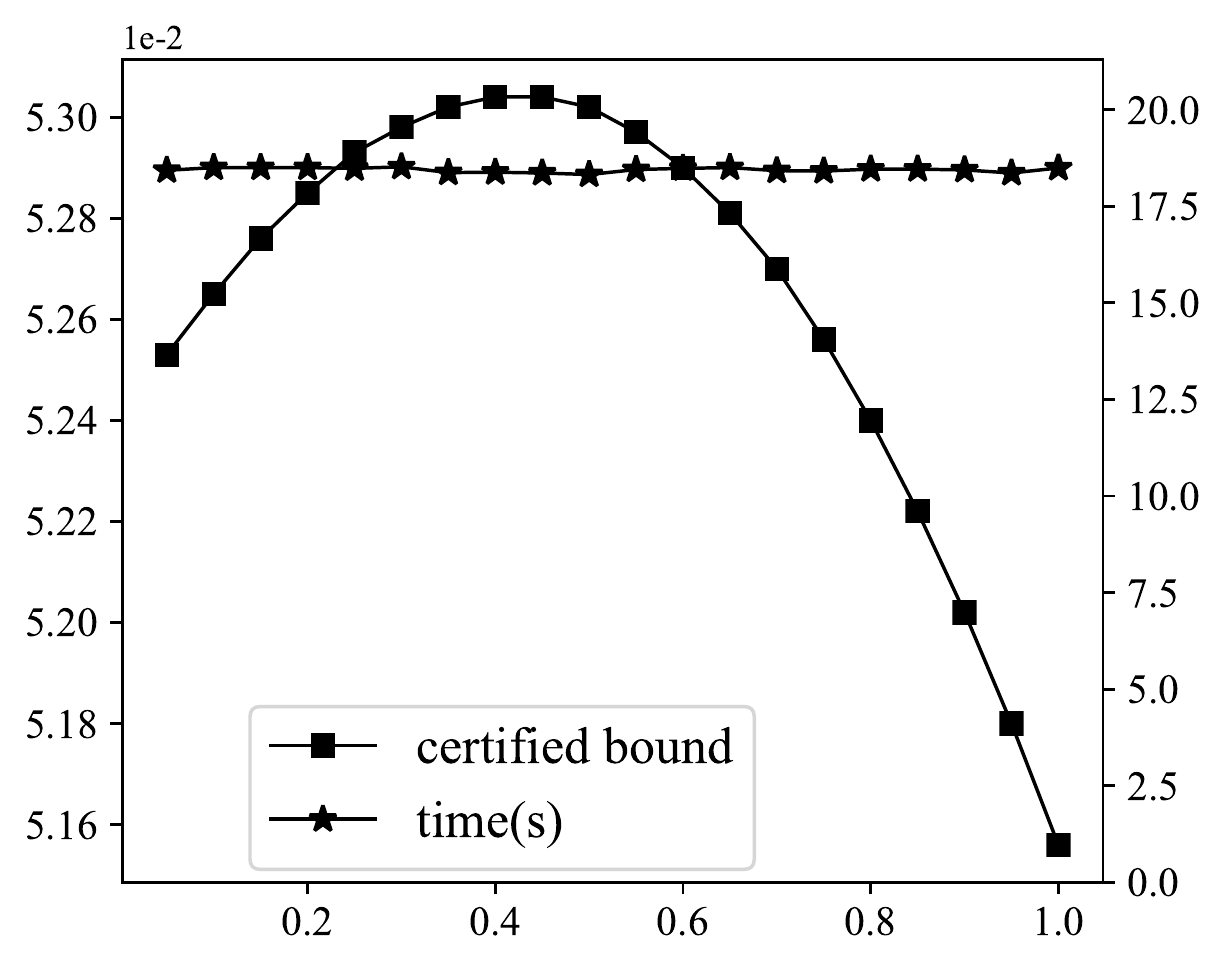}
		\caption{Gradient-based results on $\rm{CNN}_{2-3}$ trained on Mnist.}
		\label{fig:GD_mnist_cnn}
	\end{subfigure}
	\hfill
	\begin{subfigure}{0.32\textwidth}
		\includegraphics[width=\textwidth]{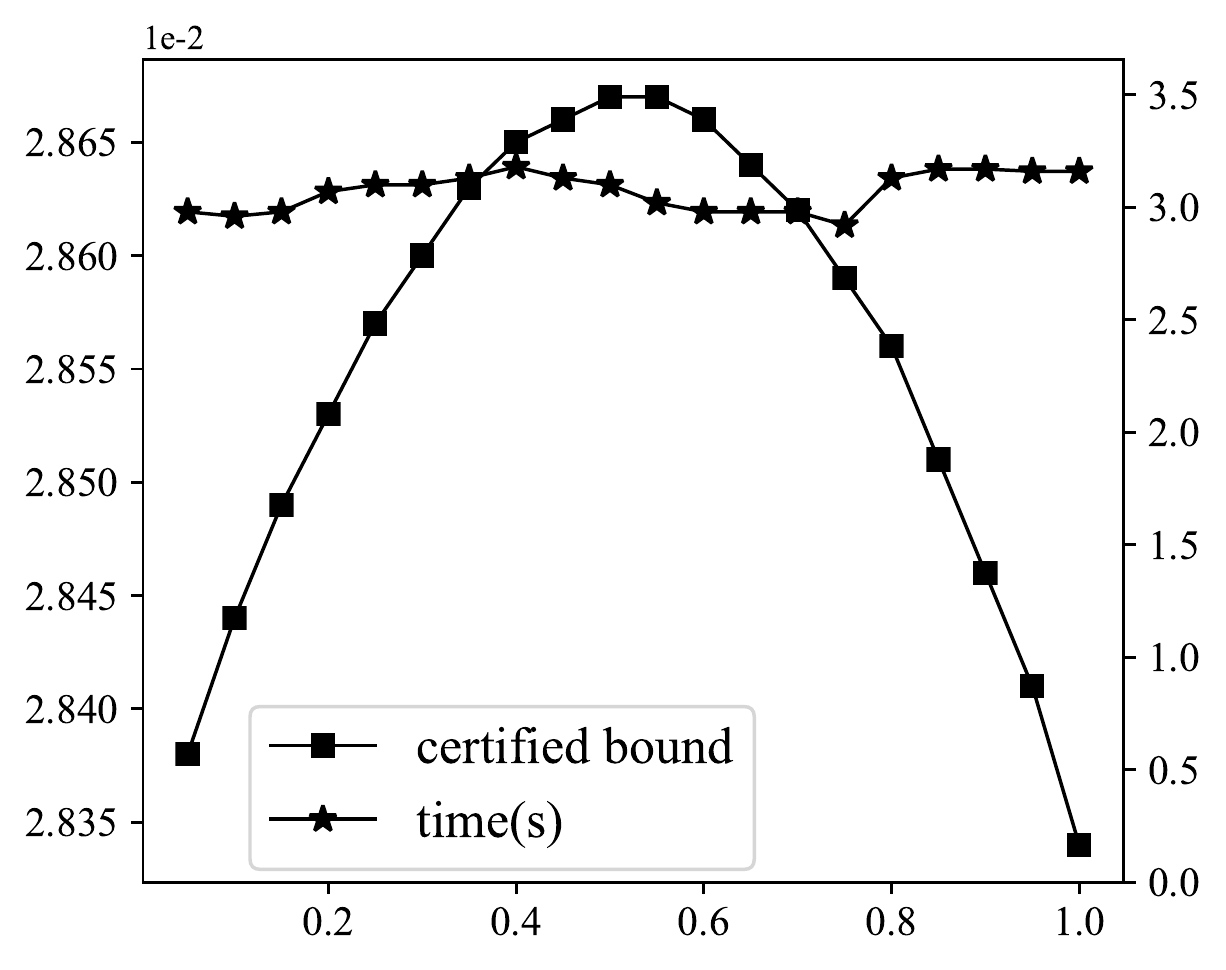}
		\caption{Gradient-based results on $\rm{FNN}_{1\times 200}$ trained on Mnist.}
		\label{fig:GD_mnist_fnn}
	\end{subfigure}
	\hfill
	\begin{subfigure}{0.32\textwidth}
		\includegraphics[width=\textwidth]{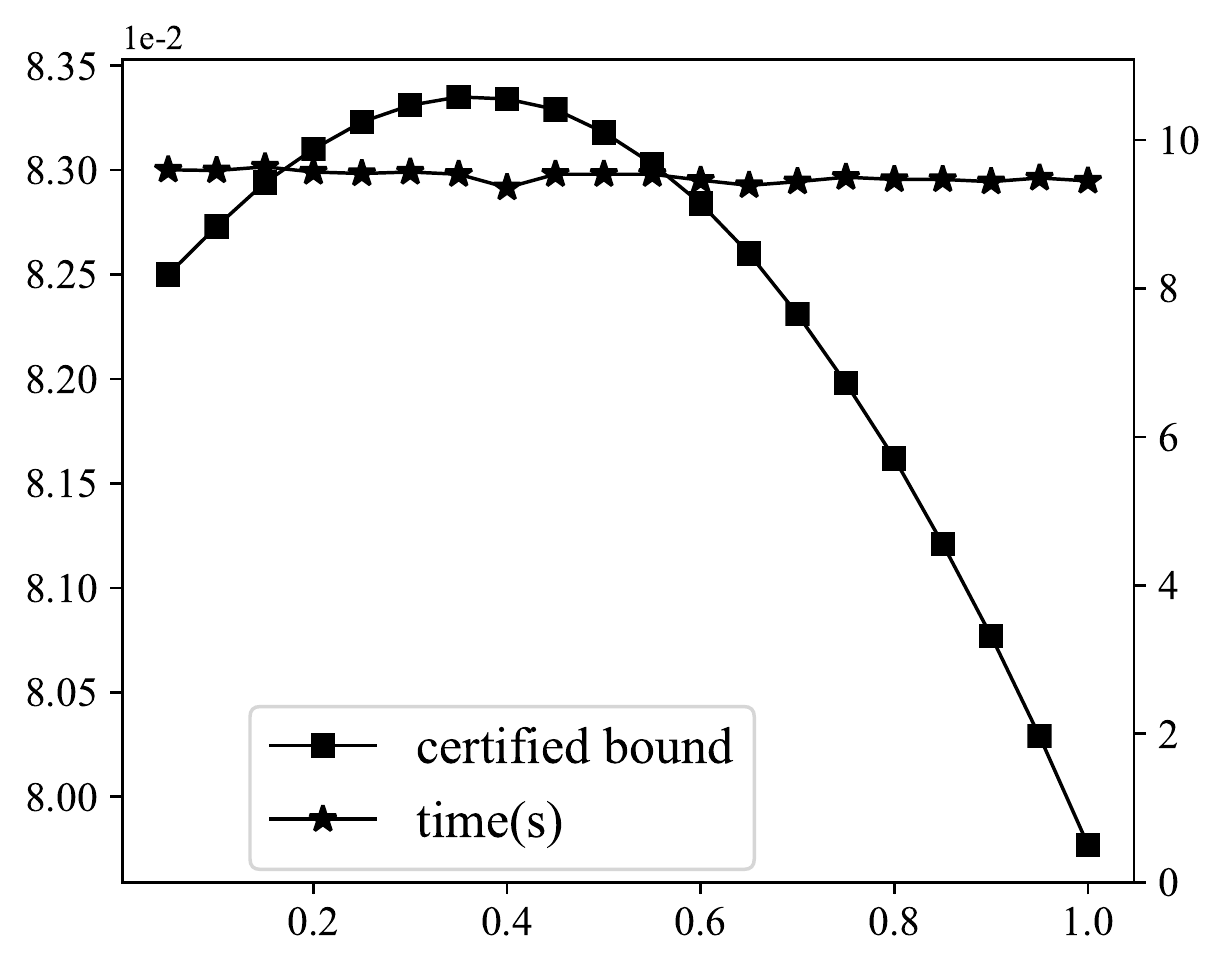}
		\caption{Gradient-based results on $\rm{CNN}_{2-2}$ trained on Fashion Mnist.}
		\label{fig:GD_fashion_mnist_cnn}
	\end{subfigure}
	\hfill
	\caption{Additional Experimental results: The effect of hyper-parameters in the Monte Carlo and gradient-based algorithms. The main coordinate denotes certified bounds and the second coordinate denotes the time consumed.}
	\label{fig:hyper_para_fig}
\end{figure*}

\subsection{Additional Results for Experiment \uppercase\expandafter{\romannumeral3}}

\label{sec:add_exp3}

This section presents additional results on the exploration of hyper-parameters in the Monte Carlo algorithm and gradient-based algorithm. The results are shown in Figure \ref{fig:hyper_para_fig}. In the Monte Carlo algorithm, the computed certified lower bounds are maximal when the sample number is around 1000. In the gradient-based algorithm, they are maximal when the step length is around $0.45\epsilon$. All these experiments are consistent with our conclusions in the text.